\documentclass{article} 
\usepackage[preprint]{icml2026}

\usepackage{hyperref}
\usepackage{url}
\usepackage[scaled]{inconsolata}
\usepackage{booktabs}
\usepackage{enumitem}
\usepackage{bm}
\usepackage{pifont}
\usepackage{xcolor}
\usepackage{tabularx}
\usepackage{array}
\usepackage{placeins}
\usepackage{microtype}
\usepackage{minitoc}
\usepackage{multirow} 
\usepackage{subcaption}
\usepackage[dvipsnames]{xcolor}
\usepackage[most]{tcolorbox}
\usepackage[font=normal,labelfont=bf]{caption}

\usepackage{xcolor}

\newtcolorbox{hypoanswer}[1][]{
  enhanced, breakable, sharp corners,
  boxrule=0pt,
  colback=Gray!5!white, colframe=Gray!5!white,
  borderline west={2pt}{0pt}{Gray!75!black},
  width=\linewidth,          
  enlarge left by=0mm,       
  enlarge right by=0mm,
  title={#1}
}
\newtcolorbox{promptbox}{
  colback=gray!10,    
  colframe=gray!40,   
  arc=3mm,            
  boxrule=0.4pt,      
  left=5pt, right=5pt, top=5pt, bottom=5pt 
}

\newtcbox{\concepthighlight}[1][]{on line,
  colback=gray!10, colframe=gray!40,
  boxrule=0pt, arc=3pt, boxsep=0pt,
  left=2pt, right=2pt, top=2pt, bottom=2pt,
  #1}

\newcommand{\conc}[1]{\concepthighlight{\texttt{#1}}}

\newcommand{\cmark}{\textcolor{green!60!black}{\ding{51}}}
\newcommand{\xmark}{\textcolor{red!70!black}{\ding{55}}}
\newcommand{\approxmark}{\textcolor{orange!80!black}{\raisebox{0.25ex}{\texttildelow}}}

\newtcolorbox{workflowbox}[1]{
  colback=blue!5!white,      
  colframe=blue!60!black,    
  fonttitle=\bfseries,       
  title=#1,                  
  boxsep=5pt,                
  left=10pt,                 
  right=10pt,                
  top=4pt,                   
  bottom=4pt,                
  breakable                  
}

\usepackage[capitalize,noabbrev]{cleveref}

\usepackage{amsmath}
\usepackage{amssymb}
\usepackage{mathtools}
\usepackage{amsthm}
\usepackage{booktabs}
\usepackage{caption}
\usepackage{graphicx}
\usepackage[most]{tcolorbox} 

\theoremstyle{plain}

\theoremstyle{definition}

\theoremstyle{remark}

\usepackage[textsize=tiny]{todonotes} 

\icmltitlerunning{Uncovering Competency Gaps}

\begin{document}

\twocolumn[
  \icmltitle{Uncovering Competency Gaps in \\ Large Language Models and Their Benchmarks}




  \begin{icmlauthorlist} 
    \icmlauthor{Maty Bohacek}{stanford,deepmind}
    \icmlauthor{Nino Scherrer}{pi}
    \icmlauthor{Nicholas Dufour}{deepmind}
    \icmlauthor{Thomas Leung}{deepmind}
    \icmlauthor{Chris Bregler}{deepmind}
    \icmlauthor{Stephanie C.Y. Chan}{deepmind}
  \end{icmlauthorlist} 

  \icmlaffiliation{stanford}{Stanford University}
  \icmlaffiliation{deepmind}{Google DeepMind} 
  \icmlaffiliation{pi}{Google, Paradigms of Intelligence Team} 

  \icmlcorrespondingauthor{Maty Bohacek}{maty@stanford.edu}

    \icmlkeywords{Large Language Models, Evaluation, Benchmarks, Sparse Autoencoders, Interpretability}
  \vskip 0.3in

    
    

]



\printAffiliationsAndNotice{Work carried out at Google DeepMind with the exception of replication studies on Llama models, which were performed exclusively using time and resources of Maty
Bohacek at Stanford University.}  


\doparttoc 
\faketableofcontents

\begin{abstract}
    The evaluation of large language models relies heavily on standardized benchmarks. These benchmarks provide useful aggregated metrics, but can obscure (i) particular sub-areas where the models are weak (``model gaps'') and (ii) imbalanced coverage in the benchmarks themselves (``benchmark gaps''). To automatically uncover both types of gaps, we propose a simple new method using concept activations from sparse autoencoders, to identify fine-grained gaps on a per-concept basis. 
    The method also benefits from grounding evaluation in the model's internal representations, as well as easy comparison across benchmarks. We applied the method to five popular open-source models and over a dozen benchmarks, as illustrative examples. 
    As validation of the approach, we found that our automatic, unsupervised method was able to recover model gaps that have been previously documented in the literature (e.g., sycophancy), in addition to identifying novel model gaps. We were also able to automatically uncover benchmark gaps: core concepts that should fall within the scope of a given benchmark. 
    Our ``competency gaps'' method can complement existing benchmarks by providing a concept-level decomposition of model behavior, and can also help benchmark developers improve and iterate upon benchmark design. 
    Code is available at \href{https://competency-gaps.github.io}{\texttt{competency-gaps.github.io}}.
\end{abstract}


\vspace{-2mm}
\section{Introduction}
\vspace{-2mm}
\label{sec:intro}
\looseness-1 Evaluating large language models (LLMs) relies heavily on benchmarks that report \emph{aggregated} scores (e.g., accuracy or pass@k). Over the last decade, hundreds of benchmarks have been introduced across diverse domains \citep{guo2023evaluating, chang2024survey}. While these benchmarks have fueled progress, uniform aggregation can obscure important sub-trends and mask specific model weaknesses \citep{hardt2025emerging, burnell2023rethink}. For instance, \citet{didolkar2024metacognitive} disaggregated performance on MATH~\citep{hendrycks2021math} and found topic-wise scores ranging from 27\% to 74\%, despite an overall score of 54\%.

\begin{figure*}[t]
    \centering 
    \includegraphics[width=0.8\linewidth]{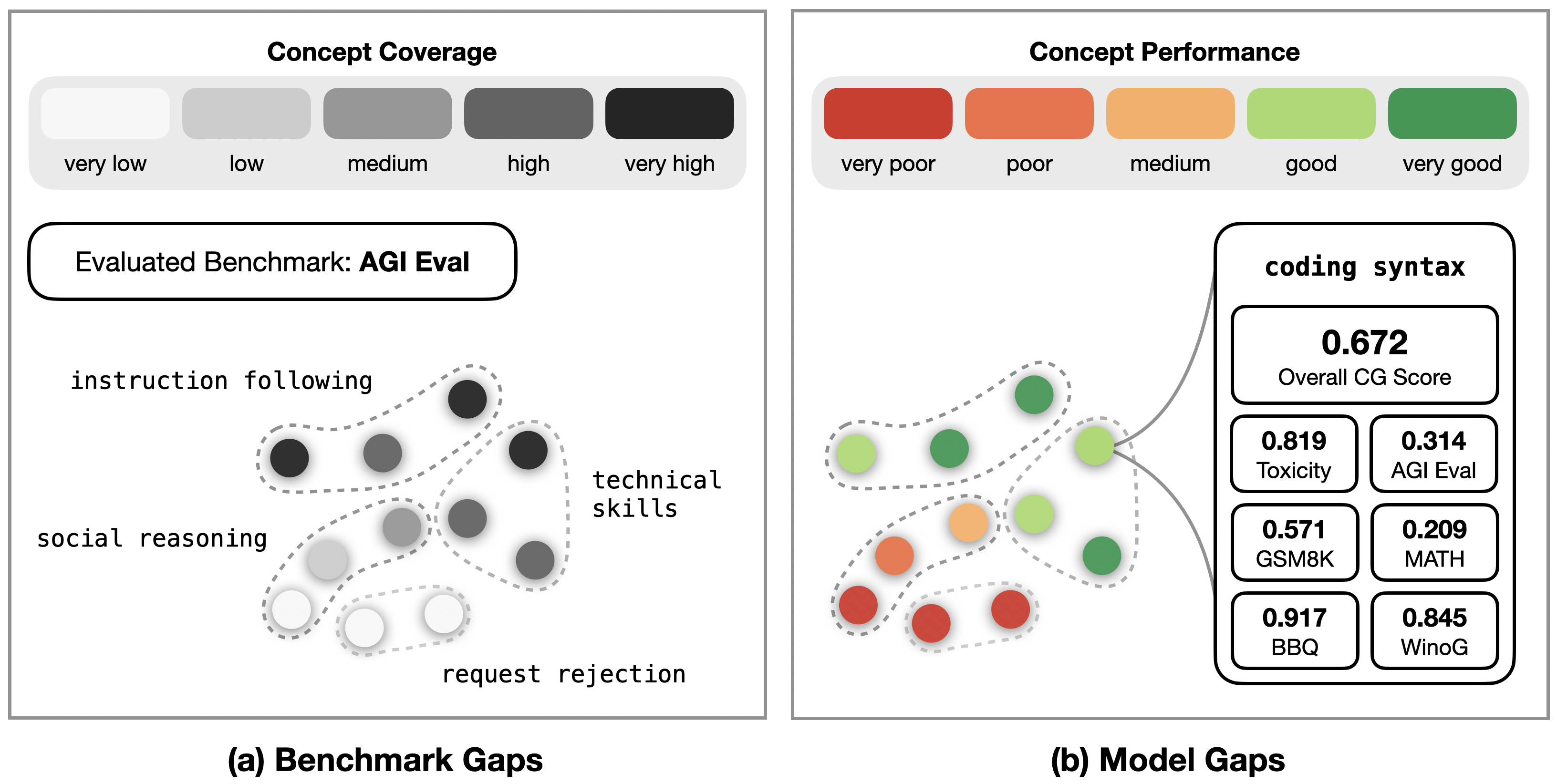}
    \caption{\textbf{Method Overview.} The method decomposes LLM evaluation into fine-grained, interpretable ``competency gaps'', using the concept dictionary learned by a sparse autoencoder. The figure provides a simplified illustration using a small number of concepts (each circle is one concept). \textbf{\textit{(a)~Benchmark Gaps}}  surface underrepresented concepts for a given benchmark (e.g. AGI Eval). We can also group the concepts into larger concept domains, e.g. ``social reasoning''. \textbf{\textit{(b)~Model~Gaps}} surface specific concept areas where a model may be under-performing. Each concept (e.g. ``coding syntax'') can be evaluated for an individual benchmark (e.g. GSM8K) or for a suite of benchmarks (``Overall CG Score'').}
    \label{fig:teaser}
    \vspace{-10pt}
\end{figure*}

\looseness-1 To counteract these aggregation issues, some benchmarks provide ``semantic’’ topic annotations (e.g.,\ hand-curated topics in MATH \citep{hendrycks2021math} or GPQA \citep{rein2024gpqa}, or embedding-based clusters~\citep{perez2023discovering}). These high-level labels help characterize benchmark distributions and disaggregate performance, but they generally provide only coarse-grained insights into model strengths and weaknesses. We currently lack a view of how finer-grained concepts, contexts, and reasoning patterns extend beyond coarse topic labels, and how they relate to real-world model usage and capabilities~\citep{miller2025evaluating,mizrahi2024state}. Furthermore, many of these semantic annotations are manually curated and difficult to scale.  Without a scalable, fine-grained understanding of benchmark distributions, we risk overlooking benchmark gaps, while systematically overtesting certain concepts. 

In this work, we are interested in the automated identification of two types of gaps: (i) model gaps, i.e., concept domains where models systematically underperform, and (ii) benchmark gaps, i.e.,\ concept domains that are inadequately represented in an evaluation dataset (see Figure~\ref{fig:teaser}). To this end, we introduce a simple new method called Competency Gaps (CG), which is both automatic and easy to scale. {The method leverages sparse autoencoders (SAEs), which decompose the internal representations of an LLM into sparse, interpretable feature vectors called SAE concept activations~\citep{bricken2023towards,cunningham2023sparse}. Each dimension of these vectors is trained to capture a distinct concept, assuming a sufficiently diverse and representative training distribution. Autointerpretability methods~\citep{mcgrath2024understanding} can be applied to assign each dimension (or ``concept'') a textual label. This fixed set of concepts is often referred to as a concept dictionary, with the dictionary size often numbering thousands or millions of concepts.} In our method, we analyze the benchmark distribution over the SAE concept dictionary to identify benchmark gaps, and also map the model's performance onto the concept space to identify model gaps\footnote{\looseness-1 The proposed approach can only detect benchmark and model gaps for concepts that are represented in the SAE space. Concept coverage of CG hence improves with more representative SAEs.}.

\looseness-1 As a demonstration of the introduced method, we evaluate Competency Gaps (CG) on five popular open-source models (\texttt{Gemma2-2B-Instruct}~\citep{gemma2024}, \texttt{Llama3.1-8B-Instruct}~\citep{grattafiori2024llama}, \texttt{Mistral-7B-Instruct-v0.1}~\citep{jiang2023mistral}, \texttt{Qwen3-4B}~\citep{yang2025qwen3}, and \texttt{DeepSeek-R1-Distill-Llama-8B}~\citep{guo2025deepseek}) across a broad range of benchmarks --- ten in our main suite, complemented by an arena-style benchmark (LMSYS Chatbot Arena) and additional capability benchmarks (SWE-Bench, Terminal-Bench, MMLU, GPQA, HLE).
\vspace{-3mm}

\begin{figure*}
    \centering
    \includegraphics[width=0.7\linewidth]{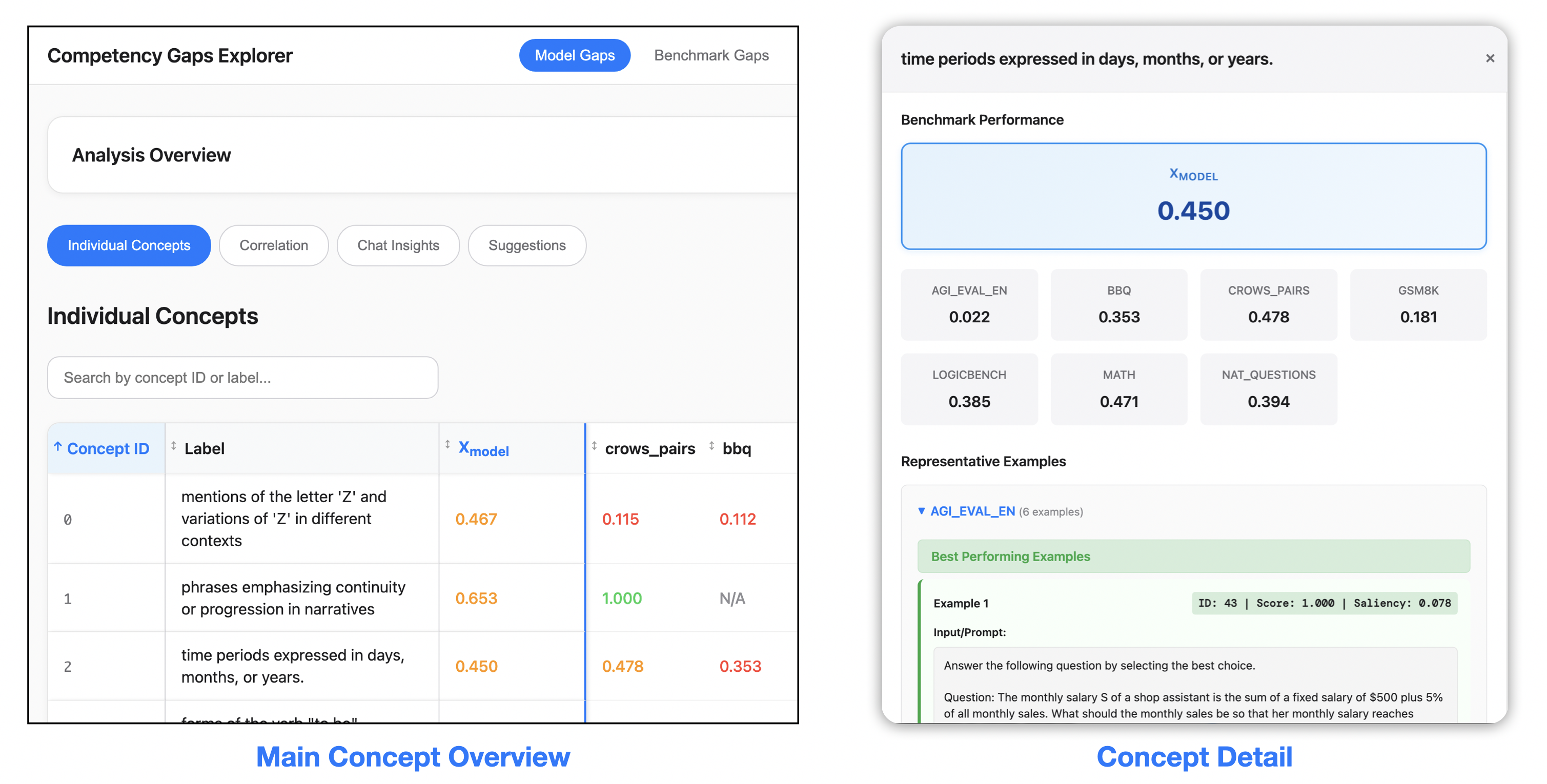}
    \vspace{-5pt}
    \caption{\textbf{Exploratory Web Application.} The CG method can output a large number of results, given the large size of most SAE concept dictionaries. This web interface enables easy, interactive exploration of CG results, including a searchable and filterable list of all concepts in the \textbf{\textit{Main Concept Overview}}, with an expandable \textbf{\textit{Concept Detail Modal}} that provides additional per-benchmark information, including specific example generations. The application also includes dedicated sections for cross-benchmark correlation, benchmark coverage inspection, and supplementary analyses. This web application is open-sourced. Additional screenshots in Appendix~\ref{app:screenshots}.
    }
    \label{fig:web_app}
    \vspace{-10pt}
\end{figure*}

\begin{itemize}[topsep=5pt, leftmargin=15pt, itemsep=-1pt]
    \item \looseness-1 \textbf{Competency Gaps (CG) Method.} We introduce a systematic method for the automated identification of benchmark coverage and model performance gaps, using an SAE-based approach. The method can be applied to any LLM and benchmark of interest.\footnote{{Using an SAE trained specifically for the model of interest allows the analysis to be grounded in that model's own representations. However, we demonstrate that even LLMs without a dedicated SAE can be evaluated using CG by leveraging the SAE from another model (Section~\ref{subsec:robustness}).}} The method decomposes aggregated benchmark scores into a large number of interpretable axes derived from the model's own representations, surfaces patterns of over- and under-tested concepts, and identifies actionable improvements for both models and benchmarks, as described in Figure~\ref{fig:workflows}. The implementation is fully open sourced.
    \item \textbf{Demonstrations of the Method on Popular LLMs and Benchmarks.} We applied CG to five popular open-source models (\texttt{Gemma2-2B-Instruct}, \texttt{Llama3.1-8B-Instruct}, \texttt{Mistral-7B-Instruct-v0.1}, \texttt{Qwen3-4B}, and \texttt{DeepSeek-R1-Distill-Llama-8B}) over more than a dozen diverse benchmarks, illustrating the kinds of insights the method enables.
    \item \textbf{Interactive Web Tool.} We also release an open-source web application (Figure~\ref{fig:web_app}) that enables users to explore model capabilities and benchmark balance, in a transparent and interpretable way.
\end{itemize}


\begin{figure*}[t!]
    \centering
    \small
    \begin{tcbitemize}[
        raster columns=2,
        raster equal height=rows,
        raster column skip=6pt,
        enhanced, sharp corners,
        boxrule=0.5pt,
        colback=Gray!5!white, colframe=Gray!75!black,
        borderline west={0pt}{0pt}{Gray!75!black},
        coltitle=black, fonttitle=\bfseries,
        attach title to upper={\par\vspace{2pt}}
    ]

    \tcbitem[title={\textbf{Recommended Workflow: Identifying Model Gaps}}]
        \begin{enumerate}[label=\arabic*., topsep=0pt, itemsep=2pt, parsep=2pt, leftmargin=*]
           
            \item { \textbf{Select Initial Benchmarks.} Gather a broad set of candidate benchmarks that are potentially relevant to the capabilities you are interested in evaluating.}
            \item { \textbf{Compute CG Scores.} Execute the Competency Gaps analysis on the initial suite: calculate the cross-benchmark coverage score $\chi_{\text{bench}}^{(c)}$ and cross-benchmark model score $\chi_{\text{model}}^{(c)}$ for all concepts.}
            \item { \textbf{Identify and Analyze Gaps.} Identify concepts with low CG scores. Once identified, you may use an LLM of choice to cluster the concepts, and/or to find conceptual gaps that are particularly relevant to your target profile.}
        \end{enumerate}

    \tcbitem[title={\textbf{Recommended Workflow: Identifying Benchmark Gaps}}]
        \begin{enumerate}[label=\arabic*., topsep=0pt, itemsep=2pt, parsep=2pt, leftmargin=*]
            \item { \textbf{Compute CG Scores.} Implement a Hugging Face data loader for your benchmark and run the Competency Gaps analysis, to compute per-concept CG scores for all concepts.}
            \item { \textbf{Identify and Analyze Gaps.} Use an LLM of choice to find low-coverage concepts that are relevant for your benchmark, and optionally ask it to cluster into themes. These insights can be used to guide improvement of the coverage of your benchmark.}
        \end{enumerate}
        \vspace{3pt}
        \textit{A similar workflow can be used to identify coverage gaps in a suite of benchmarks.}

        
    \end{tcbitemize}
    \vspace{-2mm}
    \caption{{Recommended workflows for applying the Competency Gaps (CG) method.}}
    \label{fig:workflows}
\end{figure*}

\section{Related Work}
\label{sec:rel_work}

\looseness-1 \textbf{Weakness Identification in LLMs.} Identification of LLM weaknesses has evolved from anecdotal analysis to more systematic frameworks that break down performance into specific components~\citep{jones2022capturing}. Among the first to do this were HarmBench~\citep{mazeika2024harmbench} and garak~\citep{derczynski2024garak}, which established standardized evaluations of harmful behaviors. Various methods have introduced autoraters for this task: AutoDetect~\citep{cheng2024autodetect}, for example, used three LLM-powered agents to achieve a 50\%+ weakness identification success rate. More recently, EvalTree~\citep{zeng2025evaltree} builds a hierarchical capability tree from benchmark instances and flags subtree nodes where the model performs below a statistical threshold.
\citet{gan2024reasoning} demonstrated that reasoning capabilities in LLMs can be systematically broken down and analyzed to identify specific weaknesses in logical inference chains. Other systematic evaluation methods have been proposed~\citep{kim2023prometheus}; some work in this domain has also taken an adversarial learning approach~\citep{yang2024assessing}.


\looseness-1 \textbf{Cross-Model Comparisons.} Going beyond simple benchmark scoreboards, recent work has explored approaches of characterizing the behavioral differences and internal representations across models and architectures in more detail~\citep{zheng2025model,kim2025correlated,chang2024language}. Some works, including BehaviorBox~\citep{tjuatja2025behaviorbox} and the LLM Comparator~\citep{kahng2024llm}, have emphasized side-by-side comparisons and actionable insights. Some mechanistic interpretability methods have emerged as well, e.g. by patching activations between model locations and decoding representations into interpretable text~\citep{hussein2024can}. Recent work has also scrutinized the ``universality hypothesis'', i.e.,\ universal features and circuits should appear for similar tasks across architectures~\citep{shu2025survey,yin2024comparative}.


\looseness-1 \textbf{Analysis of Benchmark Quality.} Automated quality detection has advanced through frameworks like CLEAR~\citep{chen2024automated} and SMART Filtering~\citep{gupta2024improving}, which automatically detect and filter problematic training data. A significant part of the work in this area has focused on bias evaluation~\citep{yin2026fairness,manerba2023social,koo2023benchmarking}. Some preliminary work has also considered monitoring benchmark performance across time~\citep{zhong2025watch} or using meta-learning~\citep{calian2025datarater}.


\textbf{Comparison to Related Work.} In Appendix~\ref{app:comparison_other_methods}, we compare our method to some of the above-mentioned methods that relate to the discovery of benchmark and model gaps. See \ref{app:comparison_methodology} for details of our comparison methodology, and \ref{app:baseline_comp_benchmark_gaps_overview} and \ref{app:baseline_comp_model_gaps_overview} for a high-level comparison of approaches and features of each method. Appendices~\ref{app:baseline_comp_garak}, \ref{app:baseline_comp_autodetect}, \ref{app:baseline_comp_arena_hard_auto}, and \ref{app:baseline_comp_evaltree} contain a detailed comparison of results from each method.\





\vspace{-2mm}
\section{Competency Gaps (CG) Method}
\vspace{-2mm}
\label{sec:method}
In this section, we introduce \textbf{Competency Gaps (CG)}, an automated, SAE-based method that can be used to systematically evaluate and identify:
\begin{itemize}[topsep=0pt, leftmargin=10pt, itemsep=-1.5pt]
    \item \textbf{Benchmark gaps} given a set of evaluation benchmarks $\mathcal{B}$. How good is the coverage of various concepts in a specific benchmark or set of benchmarks? Do the benchmarks have any concept coverage gaps?
    \item \textbf{Model gaps} given a language model $M$ with an associated sparse autoencoder $SAE$. How well does the model perform across various concepts? What are its strengths and weaknesses? 
\end{itemize} 

Each benchmark $b \in \mathcal{B}$ is the triple $b = (X_b, Y_b, f_b)$ of inputs, references, and a scoring function; the associated dataset $D_b$ is the concrete set of input--output pairs. Each concept $c \in C_{SAE}$ represents a distinct direction in the SAE space, and can be mapped to an autointerpretability label $l_c$ (see Appendix~\ref{app:sae_primer} for a brief SAE primer). The SAE activation score $s_{c,i}$ quantifies how much the model activates concept $c$ within a token sequence $x_i$.

\begin{hypoanswer}[]
\textbf{SAE Concept Activation Score}. A data point $i$ from a benchmark consists of an input token sequence $x_i$. The concept $c$'s activation on token $x_{i,j} \in x_i$ is computed via $SAE(x_{i,j} \ ,c)$. We sum $SAE(x_{i,j} \ ,c)$ over all tokens in $x_i$ and divide by $|x_i|$ to obtain a per-token average, the \textbf{SAE activation score $\tilde{s}_{c,i}$}:
\begin{equation}
\tilde{s}_{c,i} = \frac{1}{|x_i|}\sum_{j} SAE( x_{i,j} \ ,c)
\end{equation}
The per-token average prevents longer data points from dominating downstream statistics simply by being verbose.
\end{hypoanswer}

We describe our metrics in detail below. They satisfy the following: (i) all included benchmarks have an equal weight for computing the concept's cross-benchmark CG score, regardless of their size; (ii) all data points have an equal weight for computing per-benchmark metrics, regardless of their token length. The SAE dictionary size $K = |C_{SAE}|$ is not tuned by us; it is fixed by the pre-trained SAE's expansion factor (see Appendix~\ref{app:sae_primer}). Running the SAE adds the computational cost of approximately one extra projection at a single layer, which comes out to be a small fraction of the LLM's overall computational demands for a single forward pass. CG is hence inexpensive relative to running the benchmarks themselves.

\vspace{-2mm}
\subsection{Benchmark Gaps}
We now introduce a measure to quantify concept coverage within and across benchmarks, based on the normalized SAE concept activation score $\tilde{s}_{c,i}$ introduced above.

\begin{hypoanswer}[]
\paragraph{Per-Benchmark Coverage.} To evaluate the coverage of a concept $c$ in benchmark $b$, we define:
\begin{equation}
\chi_{\text{bench}}^{(b,c)} = \frac{\sum_{i \in D_b} \tilde{s}_{c,i}}{ \frac{1}{|C_{SAE}|} \sum_{c' \in C_{SAE}} \sum_{i \in D_b} \tilde{s}_{c',i}}
\end{equation}
which relates the activation of SAE concept $c$ in dataset $b$ to the average concept activation in $b$. Normalizing by the mean concept activation makes $\chi_{\text{bench}}^{(b,c)}$ scale-invariant across benchmarks, so values are directly comparable: $\chi_{\text{bench}}^{(b,c)} = 1$ corresponds to coverage at the benchmark's own average.
\end{hypoanswer}

\begin{hypoanswer}[]
\paragraph{Cross-Benchmark Coverage.} The overall coverage for concept $c$ in a benchmark suite $\mathcal{B}$ is the mean per-benchmark coverage taken over \emph{all} benchmarks in $\mathcal{B}$ (benchmarks where $c$ does not activate contribute $0$ and pull the score down):
\begin{equation}
\bm{\mathit{X}}_{\text{bench}}^{(c)} = \frac{1}{|\mathcal{B}|} \sum_{b \in \mathcal{B}_c} \chi_{\text{bench}}^{(b,c)}
\end{equation}
\end{hypoanswer}


\paragraph{Coverage Classification.} We label a concept as \textit{missing} from the benchmark suite if $\bm{\mathit{X}}_{\text{bench}}^{(c)} < \epsilon$ (we used $e^{-5}$). Among the remaining concepts, we consider the bottom decile ($\leq$ 10th percentile) to be \textit{underrepresented}, and the top decile ($\geq$ 90th percentile) to be \textit{overrepresented}. We can define underrepresented and overrepresented concepts for each individual benchmark similarly by using the per-benchmark concept coverage {\small $\chi_{\text{bench}}^{(b,c)}$}.

\subsection{Model Gaps}
We now turn to the quantification of model gaps, and introduce a novel measure for model performance grounded in the SAE concept space. 

\begin{hypoanswer}[]
\paragraph{Per-Benchmark Model Performance.} To evaluate model performance on a concept $c$ for a benchmark $b$, we define $\chi_{\text{model}}^{(b,c)}$:
\begin{equation}
\chi_{\text{model}}^{(b,c)} = \frac{\sum_{i \in D_b} m_b(i) \cdot \tilde{s}_{c,i}}{\sum_{i \in D_b} \tilde{s}_{c,i}}
\end{equation}
where $m_b(i)$ is the performance of the model on datapoint $i$ for benchmark $b$ (normalized to [0,1]; higher is better). The scoring policy for $m_b(i)$ is defined by the benchmark. 
If no data points activate $c$ in $b$, then $\chi_{\text{model}}^{(b,c)}$ is undefined.
\end{hypoanswer}

\begin{hypoanswer}[]
\paragraph{Cross-Benchmark Model Performance.} The overall performance for concept $c$ in a benchmark suite $\mathcal{B}$ is the mean $\chi_{\text{model}}^{(b,c)}$ across all benchmarks:
\begin{equation}
\bm{\mathit{X}}_{\text{model}}^{(c)} = \frac{1}{|\mathcal{B}_c|} \sum_{b \in \mathcal{B}_c} \chi_{\text{model}}^{(b,c)}
\end{equation}
We only consider benchmarks where $\chi_{\text{model}}^{(b,c)}$ is defined. 
If no data points in $\mathcal{B}$ activate concept $c$, then $\bm{\mathit{X}}_{\text{model}}^{(c)}$ is undefined.
\end{hypoanswer}

\begin{hypoanswer}[]
\paragraph{Model Gap.} We label a concept a \textit{model gap} if $\bm{\mathit{X}}_{\text{model}}^{(c)} < \epsilon$, for the same small $\epsilon$. These are concepts on which the model performs poorly. 
\end{hypoanswer}

\section{Demonstrations of the Method}
\label{sec:demo}

To demonstrate the kinds of insights enabled by our method, we report results of applying CG to five LLMs across several benchmark datasets. However, this analysis can be applied to any language model and is intended as part of an active, iterative process in which a benchmark or model can be continuously refined (see Figure~\ref{fig:workflows}).

\subsection{Experimental Setup}
\label{sec:experiments}





\looseness-1 \textbf{Benchmarks.} The presented method can be applied to any text-based benchmark. We demonstrate our method on ten static benchmark datasets that are regularly used for performance and safety evaluations, complemented by an arena-style benchmark and an additional set of conventional capability benchmarks (the latter reported in Appendix~\ref{app:add_results_conventional}).

\begin{itemize}[topsep=0pt,leftmargin=15pt, itemsep=-1.5pt]
    \item \emph{Factuality benchmarks:} Vectara~\citep{hughes2023vectara}; Natural Questions \citep{kwiatkowski2019natural}.
    \item \emph{Math benchmarks:} GSM8K~\citep{cobbe2021gsm8k}; MATH~\citep{hendrycks2021math}.
    \item \emph{Reasoning benchmarks:} AGI Eval~\citep{zhong2023agieval}; LogicBench~\citep{parmar2024towards}; Social IQA~\citep{sap2019socialiqa}; WinoGrande~\citep{sakaguchi2021winogrande}.
    \item \emph{Ethics \& bias benchmarks:} BBQ~\citep{parrish2021bbq}; CROWS Pairs~\citep{nangia2020crows}.
    \item {\emph{Arena-style benchmark:} LMSYS Chatbot Arena \cite{zheng2023judging}.}
    \item \emph{General ability benchmarks:} SWE-Bench~\citep{jimenez2023swebench}; Terminal-Bench~\citep{merrill2026terminal}; MMLU~\citep{hendrycks2020measuring}; GPQA~\citep{rein2024gpqa}; Humanity's Last Exam (HLE)~\citep{phan2025humanity}.
\end{itemize}


\looseness-1 \textbf{Models.} We analyzed five popular open-source LLMs with available SAEs (or transcoders) and autointerpretability labels. However, we emphasize that the method is not bound to these particular models; it can be applied to any LLM with an SAE.

\begin{itemize}[topsep=0pt, leftmargin=15pt, itemsep=-1.5pt]

    \item \texttt{Llama3.1-8B-Instruct} + Goodfire SAE attached at layer 19 \citep{grattafiori2024llama, balsam2025announcing}.

    \item \texttt{Gemma2-2B-Instruct} + Gemma Scope SAE attached at layer 20 (residual stream) \citep{gemma2024, lieberum_gemma_2024};\footnote{For the Goodfire Llama SAE, the choice of layer for SAE attachment was made by the creators of the SAE. We chose the Gemma Scope layer to be at a comparable depth. Generally, deeper layers have a tendency to represent higher-level, sentence- or discourse-level abstractions~\citep{balcells2024evolution,shi2025route}.}

    \item \texttt{Mistral-7B-Instruct-v0.1} + SAE attached at layer 16 (MLP output) \citep{jiang2023mistral, cosgrove2024mistralsae}.

    \item \texttt{Qwen3-4B} + transcoders on the residual stream \citep{yang2025qwen3, hanna2024qwen3transcoders}.

    \item \texttt{DeepSeek-R1-Distill-Llama-8B} + Llama-Scope-R1-Distill SAE attached at layer 20 (residual stream) \citep{guo2025deepseek, he2024llamascope}.

\end{itemize}

\looseness-1 The main paper reports results for \texttt{Llama3.1-8B-Instruct} on the ten static benchmarks (Section~\ref{sec:results}); results for the four additional models on the same suite are reported in Appendices~\ref{app:add_results_gemma_2_2b_it}, \ref{app:add_results_deepseek}, \ref{app:add_results_mistral}, and \ref{app:add_results_qwen3}. Results for LMSYS Chatbot Arena (\texttt{Llama3.1-8B-Instruct}) and the general ability benchmarks (\texttt{Qwen3-4B}) are reported in Appendices~\ref{app:add_results_lmsys} and~\ref{app:add_results_conventional}, respectively.

\begin{figure*}[t]
  \centering
  \begin{minipage}[t]{0.48\textwidth}
    \centering
    \includegraphics[width=\textwidth]{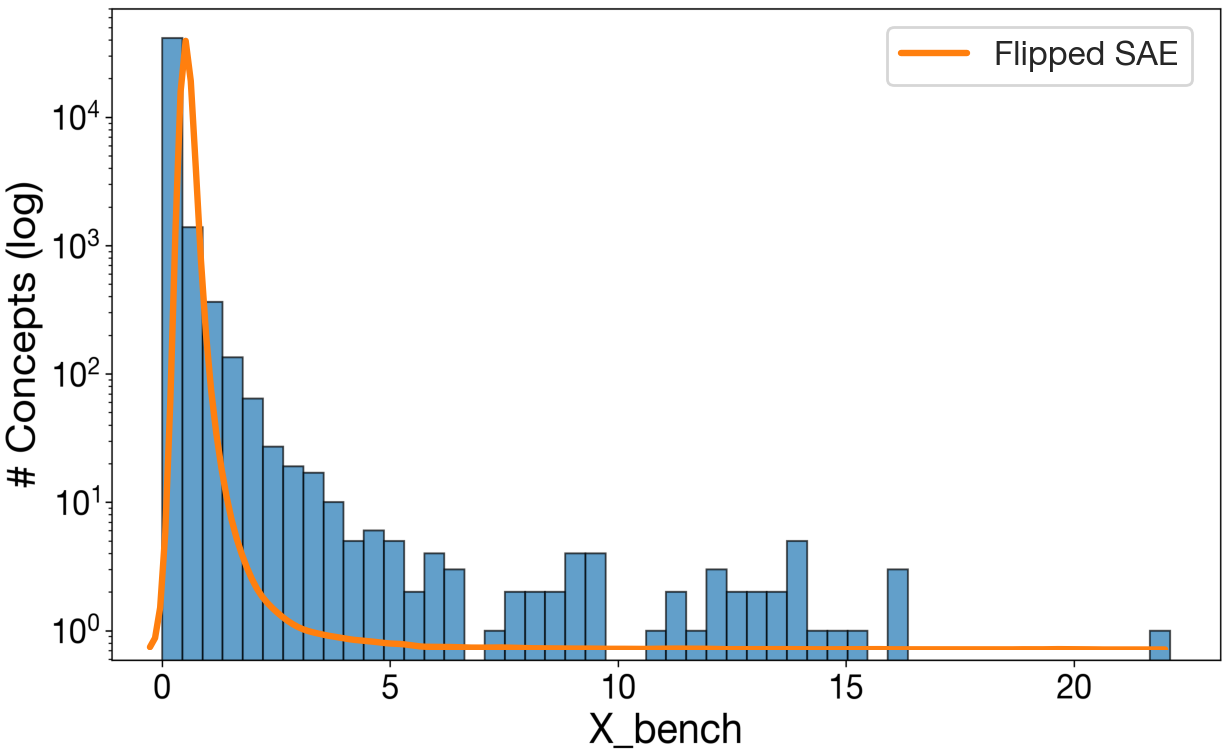}
    \caption{\textbf{Cross-Benchmark Coverage.} The distribution of {\small $\bm{\mathit{X}}_{\text{bench}}^{(c)}$} scores, indicating per-concept coverage across the $10$ evaluated benchmarks, using the SAE of Llama 3.1 8B. This distribution exhibits strong right skew (most concepts have low coverage). Thus, average performance is dominated by a few concepts with high coverage (high {\small $\bm{\mathit{X}}_{\text{bench}}^{(c)}$}).
    {The orange curve shows a similar analysis using an SAE trained for a different model (Gemma 2 2B), resulting in qualitatively similar distributions (see Section~\ref{subsec:robustness}).}}
    \label{fig:overall_coverage}
  \end{minipage}
  \hfill
  \begin{minipage}[t]{0.48\textwidth}
    \centering
    \includegraphics[width=\textwidth]{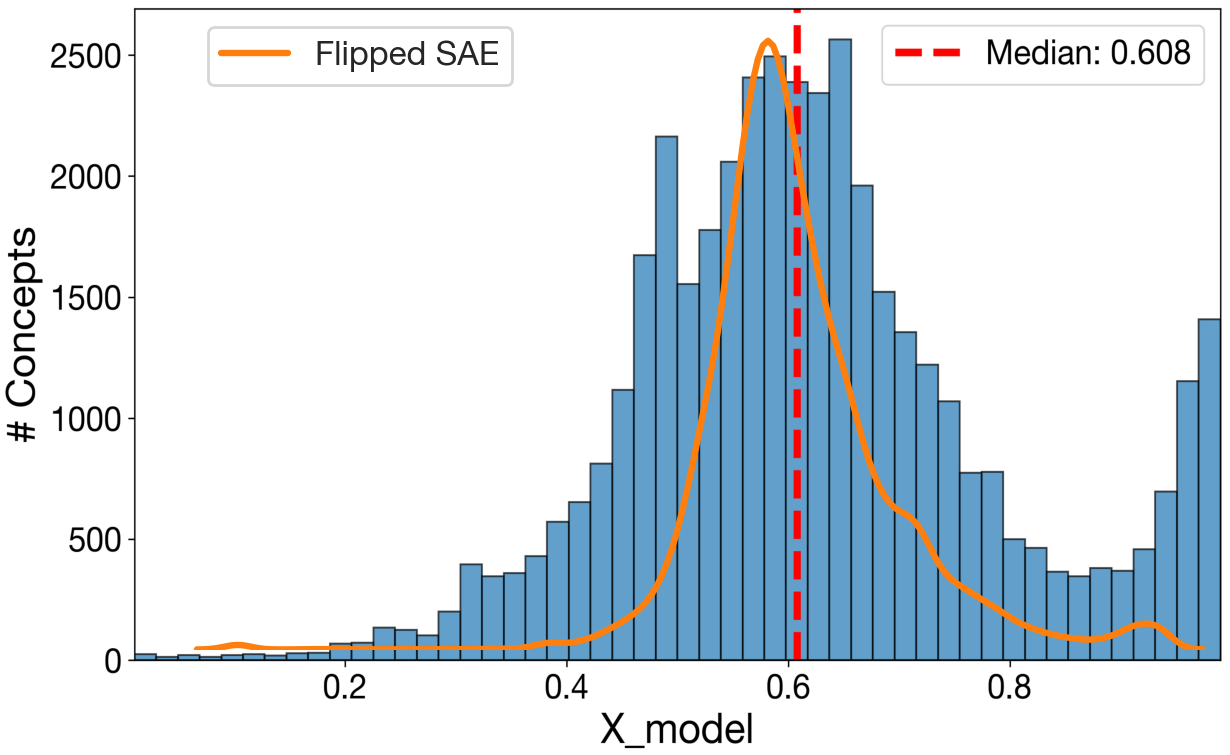}
    \caption{\textbf{Cross-Benchmark Model Performance.} The distribution of {\small $\bm{\mathit{X}}_{\text{model}}^{(c)}$} scores, indicating per-concept model performance for Llama 3.1 8B, evaluated across $10$ benchmarks. The model exhibited high variance in performance across concepts. We observe particularly high performance for some concepts related to coding, data handling, instruction following, and expressing positive sentiment toward the user. {The orange curve shows a similar analysis using an SAE trained for a different model (Gemma 2 2B), resulting in qualitatively similar distributions (see Section~\ref{subsec:robustness}).}}
    \label{fig:overall_performance}
    \vspace{0.94cm}
  \end{minipage}
  \vspace{-40pt}
\end{figure*}

\subsection{Results}
\label{sec:results}

In this section, we report results for \texttt{Llama3.1-8B-Instruct} for the ten static benchmarks described above. {We report analogous results for four additional models --- \texttt{Gemma2-2B-Instruct}~\citep{gemma2024}, \texttt{DeepSeek-R1-Distill-Llama-8B}~\citep{guo2025deepseek}, \texttt{Mistral-7B-Instruct-v0.1}~\citep{jiang2023mistral}, and \texttt{Qwen3-4B}~\citep{yang2025qwen3} --- in Appendices~\ref{app:add_results_gemma_2_2b_it}, \ref{app:add_results_deepseek}, \ref{app:add_results_mistral}, and \ref{app:add_results_qwen3}, demonstrating that CG generalizes across model families and SAE variants. We also demonstrate how CG can be used with arena-style benchmarks like LMSYS Chatbot Arena in Appendix~\ref{app:add_results_lmsys}.} To efficiently sift through the large number of results that can arise from our method (given the scale of the SAE concept dictionaries), we sometimes use another LLM (Gemini 2.5 Flash, in this case) to filter, group, and summarize sets of concept descriptions. 

\paragraph{SAE Concept Relevance.} As a complementary, more systematic lens, we also run an LLM-as-judge step that labels each concept as semantically relevant or irrelevant to each benchmark in the suite (see Appendix~\ref{app:relevance} for the prompt, per-benchmark percentages, and method). The relevant fraction varies widely across benchmarks: from $8.2\%$ (WinoGrande) to $53.1\%$ (Natural Questions) for the Gemma SAE dictionary (16K+ concepts), and from $4.9\%$ to $34.5\%$ for the Llama SAE dictionary (45K+ concepts). We release these per-concept relevance labels as a CSV alongside our code, and CG can be re-run on the relevance-filtered subset as an alternative analysis option.

\subsubsection{Benchmark Gaps}

\paragraph{Benchmarks Exhibit Skewed Representation Across Concepts.}
\label{subsubsec:existing}

\looseness-1 The cross-benchmark coverage distribution across all ten benchmarks is shown in Figure~\ref{fig:overall_coverage}. The distribution exhibits a strong right skew, resulting in both over- and under-representation of concepts in our (somewhat typical) suite of benchmarks. This skew results in any standard mean-based summary statistics being dominated by the outliers on the right part of the tail -- a small number of ``top concepts'' with high representation (high {\small $\bm{\mathit{X}}_{\text{bench}}^{(c)}$)}. Conversely, when a concept has consistently \emph{low} coverage across all included benchmarks, it risks being systematically under-tested. {Another undesirable pattern is substantial overlap across benchmarks in a single evaluation suite (Figure~\ref{fig:llama_bench_overlap}, Appendix~\ref{app:add_results_llama_3_1_8b_it})}.

\looseness-1 The top concepts mainly relate to starting new conversations and sports news, with a focus on football. Other prominent themes include syntax and attributes of articles. For example, the top 10 concepts by benchmark coverage included: \\
{\small \conc{\textbf{(56130)}} ``\texttt{English Premier League football discussions, especially about Manchester teams}''; }\\
{\small \conc{\textbf{(41290)}} ``\texttt{New conversation or topic segment boundary marker}''.}

\looseness-1 Among the concepts with the \emph{lowest} coverage score, one finds many concepts related to meta-cognition about the AI itself, e.g. its instructions, roleplaying boundaries, and how it discusses user inputs. The bottom 10 concepts by coverage include:\\
{\small \conc{\textbf{(53553)}} ``\texttt{The assistant should maintain professional boundaries when asked to roleplay}''} \\
{\small \conc{\textbf{(25352)}} ``\texttt{References to user messages or inputs in meta-discussion}''}.

We identified 314 concepts (1\%) as entirely missing from this particular suite of benchmarks. These again include concepts related to the AI's meta-cognition, as well as legal concepts. Examples include ``\texttt{The assistant explaining why it needs more information}'', ``\texttt{The assistant needs to explain its limitations or capabilities}'', and ``\texttt{Regulatory classification and compliance requirements}''. Depending on the use case, these are plausibly important concepts to include in a benchmark suite, and these results demonstrate how CG can be used to identify such gaps.






\paragraph{Individual Benchmarks Miss Relevant Concepts.}

Individual benchmarks do show a similar skew in their representation (Appendix Figure~\ref{fig:bench_lvl_coverage}), and every benchmark except Vectara misses at least 30\% of all concepts (Appendix Figure~\ref{fig:missing_concept_ratio}). However, benchmarks are usually designed to evaluate a specific subset of capabilities, and so of course it may not be desirable for individual benchmarks to have complete coverage of all concepts.

More importantly, we would like benchmarks to have coverage of \emph{relevant} concepts. To identify such missing relevant concepts, we first used the CG method to identify all missing concepts for a given benchmark. Then, we used an LLM (Gemini 2.5 Flash, in this case) to identify missing concepts that one might expect to be in scope for the benchmark (see Appendix~\ref{appsub:prompt_missing_concepts_bench_specific} for the prompt). We also used the open-sourced web app to explore and verify these examples. This process exemplifies how one might use the CG method for unsupervised discovery of benchmark gaps.

Table~\ref{tab:missing_concepts} presents illustrative examples of concepts that are missed by benchmarks. These are concepts that seem central to the desired goals of the benchmarks, e.g. \emph{AGI Eval} was missing ``\texttt{The need for thorough and objective assessment of evidence}'', and \emph{Social IQA} was missing ``\texttt{Instructions about how someone should behave or what qualities to embody}''.


\begin{table*}[htb!]
\centering 
\caption{\textbf{Examples of Missing Relevant Concepts from Three Benchmarks.}}
\label{tab:missing_concepts}
\small
\resizebox{0.95\textwidth}{!}{ 
\begin{tabular}{lcl}
\toprule
\textbf{Benchmark} & \textbf{Concept ID} & \textbf{Concept Description} \\
\toprule
\addlinespace[0.6ex]
\textit{AGI Eval} & \conc{\textbf{(33456)}} & The need for thorough and objective assessment of evidence \\
\cmidrule{2-3}
& \conc{\textbf{(59559)}} & Careful qualification and nuanced explanation of complex topics \\
\addlinespace[0.6ex]
\toprule
 \textit{LogicBench} & \conc{\textbf{(56997)}} & The model is explaining how different elements or factors relate to each other \\
\cmidrule{2-3}
& \conc{\textbf{(11957)}} & Mathematical and logical concepts across multiple languages \\
\addlinespace[0.6ex]
\toprule
\textit{Social IQA} & \conc{\textbf{(35877)}} & Speaker defending or explaining their planned actions against expectations \\
\cmidrule{2-3}
& \conc{\textbf{(1897)}} & Instructions about how someone should behave or what qualities to embody \\
\bottomrule
\end{tabular}
}

\end{table*}


\subsubsection{Model Gaps}

In addition to benchmark gaps, we also analyze model performance gaps -- on which SAE concepts do the models perform particularly well or particularly poor?


\paragraph{Model's Best-Performing Concepts Include Coding and Commitment to Help.}

Because the SAE concepts are fixed for a given model, our method enables us to easily compare and create composite results across benchmarks. Figure~\ref{fig:overall_performance} shows the distribution across concepts for cross-benchmark model performance, indicating that the model performs well on a number of concepts, with high performance across all benchmarks. Concepts with the highest cross-benchmark performance tended to concern STEM tasks (e.g.,\ coding or data handling) or helpful behaviors (e.g., positive sentiments towards the user). The top 10 concepts included: \vspace{1mm} \\
{\small \conc{\textbf{(20022)}} ``\textls[-40]{\texttt{Iteration or traversal through sequences in programming}''}}, \vspace{1mm} \\
{\small \conc{\textbf{(24074)}} ``\textls[-40]{\texttt{The assistant is about to provide an illustrative example}''}},\vspace{1mm} \\
{\small \conc{\textbf{(2461)}} ``\textls[-40]{\texttt{Assistant expressing commitment to help or do its best}''}}.\vspace{1mm} \\


\vspace{-10pt}
\paragraph{Model's Worst-Performing Concepts Include Polite Rejection, Time, and Setting Boundaries.} Perhaps more critical, from the standpoint of model evaluation, are the concepts which attained the poorest overall performance across benchmarks. An interesting recurring theme is that the worst-performing concepts include opposites to the helpful/sycophantic concepts discussed in the previous section (which came at the top in performance). Examples of worst-performing concepts include: \vspace{1mm} \\
{\small \conc{\textbf{(26535)}} ``\textls[-40]{\texttt{The assistant needs to politely reject or redirect inappropriate requests}''}},\vspace{1mm} \\ 
{\small\conc{\textbf{(56928)}} ``\textls[-40]{\texttt{The assistant maintaining professional boundaries while offering appropriate help}''}}.

Furthermore, Competency Gaps identified other groups of competencies that have been anecdotally identified as LLM weaknesses in prior literature, validating our automated approach as a scalable and systematic method for identifying such model weaknesses. These include:
\begin{itemize}[topsep=-1pt,itemsep=0pt, leftmargin=15pt]
    \item \textbf{Representations of Time:} \vspace{1mm}\\
    {\small \conc{\textbf{(29324)}} ``\textls[-40]{\texttt{Historical date and time period formatting}}''\\
    \conc{\textbf{(12644)}} ``\textls[-40]{\texttt{Cooking time durations in recipe instructions}''}})
    \item \textbf{Image Manipulations:}\vspace{1mm}\\
    {\small\conc{\textbf{(30206)}} ``\textls[-40]{\texttt{Image contrast adjustments in photo editing and computer vision}''}} \footnote{\texttt{Llama3.1-8B-Instruct} is a text-only model, and we found that such concepts were activated by metadata indicating image editing, Photoshop scripts, and tutorials on the topic. This type of review of specific data points—both in the benchmarks and in the SAE training dataset—is enabled by our web app, and proves helpful for understanding the context and nuance of various concepts.}
    \item \textbf{Palindromes / Reasoning over Letters:} \vspace{1mm}\\
    {\small\conc{\textbf{(56613)}} ``\textls[-40]{\texttt{Code examples and explanations of palindrome checking algorithms}''}}
    \item \textbf{Mathematical Operations:}\vspace{1mm}\\
    {\small\conc{\textbf{(64527)}} ``\textls[-40]{\texttt{Mathematical addition operator in calculations}''}}.
\end{itemize}

Additionally, the proposed method surfaces LLM weaknesses that have \emph{not} been previously studied in the literature. One such category is ``appeals to intuition in reasoning or decision making '': \vspace{1mm} \\
{\small  \conc{\textbf{(64413)}} ``\textls[-40]{\texttt{The concept of grokking and deep intuitive understanding}''}}\vspace{1mm} \\
{\small  \conc{\textbf{(64540)}} ``\textls[-40]{\texttt{Intuitive understanding and natural ease of use}''}}

Moreover, the exploratory web application allows users to directly examine example data points and better understand how such competencies manifest in practice (see Figure~\ref{fig:model_gap_examples}).


\subsubsection{{Robustness}}
\label{subsec:robustness}

{ Prior work has provided mixed evidence on the stability of SAE concepts ~\citep{paulo2025sparse,menon2025analyzing}. We therefore set out to evaluate the stability and generalizability of CG findings. }

{ \paragraph{A Model-Specific SAE Is Not Necessary, and Different SAEs Can Yield Similar Results.} We compared CG insights derived from Llama 3.1 8B using its own model-specific SAE, vs those obtained through Gemma 2 2B activations and SAE. As shown in Figures~\ref{fig:overall_coverage} and~\ref{fig:overall_performance}, the overall shape and medians of the score distributions were similar, especially considering the difference in dictionary sizes (Llama's SAE contains 2.8x more concepts than Gemma's). When comparing the best- and worst-performing concepts, we observed clear correspondences and similar interpretive insights, as shown in Table~\ref{tab:llama_gemma_concepts}. This suggests that CG can yield meaningful insights even for LLMs \emph{without} their own pre-trained SAE and demonstrates the overall stability of the method. Nonetheless, we expect that a model-specific SAE with a larger dictionary still offers the most precise and grounded results. }

{ We emphasize that this cross-SAE robustness is \emph{qualitative}, at the level of themes, rather than exact concept-by-concept identity. The two SAEs slice the same semantic regions at different granularities (e.g., Llama's \conc{\textbf{(2872)}} ``\texttt{Explaining time requirements and duration}'' aligns with Gemma's broader \conc{\textbf{(9936)}} ``\texttt{Dates and numeric sequences}''), so Table~\ref{tab:llama_gemma_concepts} should be read as showing thematic --- not bijective --- correspondences. This is the level of consistency required for CG's intended use as a tool for unsupervised gap discovery. }

\paragraph{{CG Scores Are Consistent Across Perturbations}} {To assess the robustness of CG, we re-ran the full analysis 100 times, each time randomly dropping 20\% of the examples per benchmark. The resulting standard deviations were low: 0.014 for $\bm{\mathit{X}}_{\text{model}}$ and 0.025 for $\bm{\mathit{X}}_{\text{bench}}$ on Llama. This suggests CG yields stable scores under random subsampling. }

\paragraph{{CG Scores Respond to Adversarial Perturbations.}} {We conducted an adversarial ablation in which we identified the top 100 best- and worst-performing concepts, then removed the most salient 100 datapoints associated with them across all benchmarks. As expected, removing rows aligned with high-performing concepts lowered median $\bm{\mathit{X}}_{\text{model}}$ on average by 0.6\%, while removing those aligned with low-performing concepts increased it on average by 1.3\% (repeated across 10 repetitions). Despite removing less than 1\% of the testing data, we were able to make predictable and consistent changes to the overall performance, suggesting that CG surfaces meaningful concept-level information. }








\section{Discussion}
\label{sec:discussion}

We have introduced a simple new method, Competency Gaps, which helps automatically identify gaps in benchmark coverage as well as gaps in model performance. We demonstrated the utility of this method, by performing example analyses on five models and a sampling of popular benchmarks. We also show that the method is robust to changes in the underlying sparse autoencoder and perturbations of the data.

Our benchmark gap analysis revealed a structural imbalance in a sample of popular benchmarks. For example, the benchmarks strongly emphasized concepts related to authority, control, and instruction-following, while neglecting complementary concepts related to polite refusals, meta-cognition, and meta-discussion about the AI itself. When evaluating across a benchmark suite, such skewed representation within the benchmarks may skew our perception of model capabilities. Our method also identified potential coverage gaps in specific benchmarks, pinpointing missing concepts that seemed relevant to each benchmark's scope.

A similar bias toward sycophancy and instruction-following emerged in the model gap analysis. Here, positive or sycophantic concepts score highest, while opposing concepts (e.g. those linked to rejecting requests or setting boundaries) score lowest. While this is likely due in part to instruction-based post-training, we note that model gaps and benchmark gaps are heavily intertwined, and that benchmark gaps may lead to model gaps --- model developers may unknowingly overlook (or be disincentivized to address) model weaknesses that are poorly covered by existing benchmarks. This entanglement is also visible in widely-used benchmarks: MMLU~\citep{hendrycks2020measuring} privileges factual recall over capabilities like calibrated refusal or meta-cognition, and SWE-Bench~\citep{jimenez2023swebench} encodes a narrow slice of software engineering (Python patches against unit tests). Models tuned for these benchmarks then look strongest along the concepts those benchmarks over-represent. Concretely, CG identifies \emph{compilation} and \emph{computer security} as among the most underrepresented concepts in SWE-Bench --- consistent with the absence of any compiler or security repository among its twelve constituent projects --- and surfaces missing historical coverage in MMLU (e.g., the Arab Spring and the Cuban Missile Crisis); we report these findings in Appendices~\ref{app:add_results_conventional_swe_bench} and~\ref{app:add_results_conventional_mmlu}.

\subsection{Design Choices}

\textbf{Model-Internal SAE Concepts vs. Unified Text Embeddings.} A natural alternative would be to use a model-independent text embedding to characterize benchmark items. We avoid this for two reasons: (1) external embeddings decouple the analysis from the model under evaluation, so surfaced ``gaps'' no longer reflect anything $M$ itself represents; (2) such embeddings are typically dense and entangled, making the resulting axes harder to interpret than sparse SAE concepts. Using a model-internal SAE grounds the analysis in $M$, at the cost of being SAE-specific. For more on this trade-off, see Section~\ref{subsec:robustness}.

\textbf{Full-Dictionary vs. Filtered Reporting.} Throughout the main results, we report statistics over the \emph{full} SAE dictionary to avoid importing selection bias from a filtering criterion. The trade-off is that the full dictionary also includes low-quality or near-duplicate concepts that can make raw distributions noisier. Filtered analyses (e.g., LLM-selected critical concepts, as described in Appendix~\ref{appsub:prompt_missing_concepts_bench_specific}) are useful as a post-hoc lens for human inspection, but should not be used for aggregate statistics.


\subsection{Limitations}
\looseness-1 \textbf{Concept Coverage is Limited to SAE Concept Space.} The usage of SAE concepts means that this method can only detect competency gaps for concepts that have \emph{some} representation in the model; concepts that are absent from the model are by construction invisible to CG. Thus, we should take care in how we interpret our results: (1) \emph{benchmark gap} analyses may reveal where models possess internal representations for concepts that evaluations fail to adequately test, and (2) \emph{model gap} analyses may identify cases where models have flawed or partial internal representations that they cannot apply effectively to downstream tasks.

\looseness-1 \textbf{Concept Activation Is Not Capability.} CG does not assume that activation of a concept implies competence in that concept. Capability is measured solely through the benchmark scoring function $m_b(i)$; concept activations are used only for attribution, weighting performance into the SAE space. Gaps are diagnosed jointly from activation and performance, not from either alone.

\looseness-1 \textbf{Uncontrolled Concept Difficulty.} A concept's CG score can be confounded by the difficulty of the items in which it activates: a low $\bm{\mathit{X}}_{\text{model}}^{(c)}$ may reflect a genuine model gap or simply the fact that $c$ activates on the hardest examples (the ``hard-question confound''). Calibrating for item-level difficulty is an important direction for future work.

\subsection{Downstream Applications and Future Work}


\paragraph{Benchmark Search and Selection.} Our method could be integrated into a database of available benchmarks (e.g., Hugging Face Datasets). Users seeking benchmarks to evaluate their models could use CG to inform their selection to achieve a desired coverage. 
\vspace{-0.5em}
\paragraph{Targeted Creation of Novel Benchmark Data.} Beyond characterizing existing benchmarks, a list of underrepresented concepts within the suite could guide the generation of novel benchmark data. Such data could be created either by prompting LLMs with autointerpretability labels or by directly applying SAE-based steering during generation.
\vspace{-0.5em}

\looseness=-1 \paragraph{Method Improvements.} We hope that this work may serve as a starting point for further development on the method. For example, in future we may wish to incorporate an automated sensitivity analysis for the choice of SAE layer, or per-token activations for more fine-grained analyses.

Current model evaluations may risk a narrower view of ``competence," potentially leaving critical gaps untested. We hope that our method enables benchmark developers and model testers to identify both benchmark gaps and model gaps, uncovering and addressing weaknesses in areas that may be essential for real-world, human-facing use-cases.

\section*{Impact Statement}

This paper presents work whose goal is to advance the field of machine learning, and especially model evaluation. There are many potential societal consequences of our work, none of which we feel must be specifically highlighted here.

\section*{Acknowledgements}

The authors would hereby like to thank the following colleagues, listed in alphabetical order, for helpful discussions: Tom Lieberum, Neel Nanda, Senthooran Rajamanoharan, and Jasper Snoek.

\section*{LLM Usage}

LLMs were used in parts of the implementation and during the writing of the paper (e.g., paragraph shortening, transition refinement, etc.). AI-powered search engines were also used to help identify some references. LLM clustering was used to sift through large amounts of data produced by our method.




\bibliography{refs}

@article{zhong2025watch,
  title={Watch the Weights: Unsupervised monitoring and control of fine-tuned {LLMs}},
  author={Zhong, Ziqian and Raghunathan, Aditi},
  journal={arXiv preprint arXiv:2508.00161},
  year={2025}
}

@article{mazeika2024harmbench,
  title={{HarmBench}: A standardized evaluation framework for automated red teaming and robust refusal},
  author={Mazeika, Mantas and Phan, Long and Yin, Xuwang and Zou, Andy and Wang, Zifan and Mu, Norman and Sakhaee, Elham and Li, Nathaniel and Basart, Steven and Li, Bo and others},
  journal={arXiv preprint arXiv:2402.04249},
  year={2024}
}

@article{gupta2024improving,
  title={Improving model evaluation using smart filtering of benchmark datasets},
  author={Gupta, Vipul and Ross, Candace and Pantoja, David and Passonneau, Rebecca J and Ung, Megan and Williams, Adina},
  journal={arXiv preprint arXiv:2410.20245},
  year={2024}
}

@article{zheng2025model,
  title={Model Directions, Not Words: Mechanistic Topic Models Using Sparse Autoencoders},
  author={Zheng, Carolina and Beltran-Velez, Nicolas and Karlekar, Sweta and Shi, Claudia and Nazaret, Achille and Mallik, Asif and Feder, Amir and Blei, David M},
  journal={arXiv preprint arXiv:2507.23220},
  year={2025}
}

@article{kim2025correlated,
  title={Correlated Errors in Large Language Models},
  author={Kim, Elliot and Garg, Avi and Peng, Kenny and Garg, Nikhil},
  journal={arXiv preprint arXiv:2506.07962},
  year={2025}
}

@article{tjuatja2025behaviorbox,
  title={{BehaviorBox}: {A}utomated Discovery of Fine-Grained Performance Differences Between Language Models},
  author={Tjuatja, Lindia and Neubig, Graham},
  journal={arXiv preprint arXiv:2506.02204},
  year={2025}
}

@article{gan2024reasoning,
  title={Reasoning robustness of {LLMs} to adversarial typographical errors},
  author={Gan, Esther and Zhao, Yiran and Cheng, Liying and Mao, Yancan and Goyal, Anirudh and Kawaguchi, Kenji and Kan, Min-Yen and Shieh, Michael},
  journal={arXiv preprint arXiv:2411.05345},
  year={2024}
}

@article{derczynski2024garak,
  title={garak: A framework for security probing large language models},
  author={Derczynski, Leon and Galinkin, Erick and Martin, Jeffrey and Majumdar, Subho and Inie, Nanna},
  journal={arXiv preprint arXiv:2406.11036},
  year={2024}
}

@misc{zhong2023agieval,
      title={{AGIEval}: A Human-Centric Benchmark for Evaluating Foundation Models}, 
      author={Wanjun Zhong and Ruixiang Cui and Yiduo Guo and Yaobo Liang and Shuai Lu and Yanlin Wang and Amin Saied and Weizhu Chen and Nan Duan},
      year={2023},
      eprint={2304.06364},
      archivePrefix={arXiv},
      primaryClass={cs.CL}
}

@article{parmar2024towards,
  title={Towards Systematic Evaluation of Logical Reasoning Ability of Large Language Models},
  author={Parmar, Mihir and Patel, Nisarg and Varshney, Neeraj and Nakamura, Mutsumi and Luo, Man and Mashetty, Santosh and Mitra, Arindam and Baral, Chitta},
  journal={arXiv preprint arXiv:2404.15522},
  year={2024}
}

@article{sap2019socialiqa,
  title={{SocialIQA}: Commonsense reasoning about social interactions},
  author={Sap, Maarten and Rashkin, Hannah and Chen, Derek and LeBras, Ronan and Choi, Yejin},
  journal={arXiv preprint arXiv:1904.09728},
  year={2019}
}

@article{sakaguchi2021winogrande,
  title={Winogrande: An adversarial winograd schema challenge at scale},
  author={Sakaguchi, Keisuke and Bras, Ronan Le and Bhagavatula, Chandra and Choi, Yejin},
  journal={Communications of the ACM},
  volume={64},
  number={9},
  pages={99--106},
  year={2021},
  publisher={ACM New York, NY, USA}
}

@article{merrill2026terminal,
  title={{Terminal-Bench}: Benchmarking agents on hard, realistic tasks in command line interfaces},
  author={Merrill, Mike A and Shaw, Alexander G and Carlini, Nicholas and Li, Boxuan and Raj, Harsh and Bercovich, Ivan and Shi, Lin and Shin, Jeong Yeon and Walshe, Thomas and Buchanan, E Kelly and others},
  journal={arXiv preprint arXiv:2601.11868},
  year={2026}
}

@article{phan2025humanity,
  title={Humanity's last exam},
  author={Phan, Long and Gatti, Alice and Han, Ziwen and Li, Nathaniel and Hu, Josephina and Zhang, Hugh and Zhang, Chen Bo Calvin and Shaaban, Mohamed and Ling, John and Shi, Sean and others},
  journal={arXiv preprint arXiv:2501.14249},
  year={2025}
}

@article{kwiatkowski2019natural,
  title={Natural questions: a benchmark for question answering research},
  author={Kwiatkowski, Tom and Palomaki, Jennimaria and Redfield, Olivia and Collins, Michael and Parikh, Ankur and Alberti, Chris and Epstein, Danielle and Polosukhin, Illia and Devlin, Jacob and Lee, Kenton and others},
  journal={Transactions of the Association for Computational Linguistics},
  volume={7},
  pages={453--466},
  year={2019},
  publisher={MIT Press One Rogers Street, Cambridge, MA 02142-1209, USA journals-info~…}
}

@misc{hughes2023vectara,
  title        = {Vectara Hallucination Leaderboard},
  author       = {Hughes, Simon and Bae, Minseok and Li, Miaoran},
  year         = {2023},
  howpublished = {\url{https://github.com/vectara/hallucination-leaderboard}}
}

@article{cobbe2021gsm8k,
  title={Training Verifiers to Solve Math Word Problems},
  author={Cobbe, Karl and Kosaraju, Vineet and Bavarian, Mohammad and Chen, Mark and Jun, Heewoo and Kaiser, Lukasz and Plappert, Matthias and Tworek, Jerry and Hilton, Jacob and Nakano, Reiichiro and Hesse, Christopher and Schulman, John},
  journal={arXiv preprint arXiv:2110.14168},
  year={2021}
}

@article{hendrycks2021math,
  title={Measuring Mathematical Problem Solving With the {MATH} Dataset},
  author={Dan Hendrycks and Collin Burns and Saurav Kadavath and Akul Arora and Steven Basart and Eric Tang and Dawn Song and Jacob Steinhardt},
  journal={NeurIPS},
  year={2021}
}

@article{parrish2021bbq,
  title={{BBQ}: A hand-built bias benchmark for question answering},
  author={Parrish, Alicia and Chen, Angelica and Nangia, Nikita and Padmakumar, Vishakh and Phang, Jason and Thompson, Jana and Htut, Phu Mon and Bowman, Samuel R},
  journal={arXiv preprint arXiv:2110.08193},
  year={2021}
}

@article{nangia2020crows,
  title={{CrowS}-{Pairs}: A challenge dataset for measuring social biases in masked language models},
  author={Nangia, Nikita and Vania, Clara and Bhalerao, Rasika and Bowman, Samuel R},
  journal={arXiv preprint arXiv:2010.00133},
  year={2020}
}

@article{bricken2023towards,
  title={Towards monosemanticity: Decomposing language models with dictionary learning},
  author={Bricken, Trenton and Templeton, Adly and Batson, Joshua and Chen, Brian and Jermyn, Adam and Conerly, Tom and Turner, Nick and Anil, Cem and Denison, Carson and Askell, Amanda and others},
  journal={Transformer Circuits Thread},
  volume={2},
  year={2023}
}

@article{cunningham2023sparse,
  title={Sparse autoencoders find highly interpretable features in language models},
  author={Cunningham, Hoagy and Ewart, Aidan and Riggs, Logan and Huben, Robert and Sharkey, Lee},
  journal={arXiv preprint arXiv:2309.08600},
  year={2023}
}

@article{shu2025survey,
  title={A survey on sparse autoencoders: Interpreting the internal mechanisms of large language models},
  author={Shu, Dong and Wu, Xuansheng and Zhao, Haiyan and Rai, Daking and Yao, Ziyu and Liu, Ninghao and Du, Mengnan},
  journal={arXiv preprint arXiv:2503.05613},
  year={2025}
}

@book{templeton2024scaling,
  title={Scaling monosemanticity: Extracting interpretable features from {Claude 3} sonnet},
  author={Templeton, Adly},
  year={2024},
  publisher={Anthropic}
}

@article{gao2024scaling,
  title={Scaling and evaluating sparse autoencoders},
  author={Gao, Leo and la Tour, Tom Dupr{\'e} and Tillman, Henk and Goh, Gabriel and Troll, Rajan and Radford, Alec and Sutskever, Ilya and Leike, Jan and Wu, Jeffrey},
  journal={arXiv preprint arXiv:2406.04093},
  year={2024}
}

@article{cheng2024autodetect,
  title={{AutoDetect}: Towards a unified framework for automated weakness detection in large language models},
  author={Cheng, Jiale and Lu, Yida and Gu, Xiaotao and Ke, Pei and Liu, Xiao and Dong, Yuxiao and Wang, Hongning and Tang, Jie and Huang, Minlie},
  journal={arXiv preprint arXiv:2406.16714},
  year={2024}
}

@article{jones2022capturing,
  title={Capturing failures of large language models via human cognitive biases},
  author={Jones, Erik and Steinhardt, Jacob},
  journal={Advances in Neural Information Processing Systems},
  volume={35},
  pages={11785--11799},
  year={2022}
}

@inproceedings{kim2023prometheus,
  title={Prometheus: Inducing fine-grained evaluation capability in language models},
  author={Kim, Seungone and Shin, Jamin and Cho, Yejin and Jang, Joel and Longpre, Shayne and Lee, Hwaran and Yun, Sangdoo and Shin, Seongjin and Kim, Sungdong and Thorne, James and others},
  booktitle={The Twelfth International Conference on Learning Representations},
  year={2023}
}

@article{yang2024assessing,
  title={Assessing adversarial robustness of large language models: An empirical study},
  author={Yang, Zeyu and Meng, Zhao and Zheng, Xiaochen and Wattenhofer, Roger},
  journal={arXiv preprint arXiv:2405.02764},
  year={2024}
}

@article{chang2024language,
  title={Language model behavior: A comprehensive survey},
  author={Chang, Tyler A and Bergen, Benjamin K},
  journal={Computational Linguistics},
  volume={50},
  number={1},
  pages={293--350},
  year={2024},
  publisher={MIT Press One Broadway, 12th Floor, Cambridge, Massachusetts 02142, USA~…}
}

@inproceedings{kahng2024llm,
  title={{LLM} comparator: Visual analytics for side-by-side evaluation of large language models},
  author={Kahng, Minsuk and Tenney, Ian and Pushkarna, Mahima and Liu, Michael Xieyang and Wexler, James and Reif, Emily and Kallarackal, Krystal and Chang, Minsuk and Terry, Michael and Dixon, Lucas},
  booktitle={Extended Abstracts of the CHI Conference on Human Factors in Computing Systems},
  pages={1--7},
  year={2024}
}

@misc{hussein2024can,
  author       = {Nada, Hussein and Asma, Ghandeharioun and Ryan, Mullins and Emily, Reif and Jimbo, Wilson and Nithum, Thain and Lucas, Dixon},
  title        = {Can Large Language Models Explain Their Internal Mechanisms?},
  howpublished = {Explorable, Google PAIR},
  year         = {2024},
  month        = {jul},
  note         = {\url{https://pair.withgoogle.com/explorables/patchscopes/}}
}

@inproceedings{yin2024comparative,
  title={Comparative study of large language model architectures on frontier},
  author={Yin, Junqi and Bose, Avishek and Cong, Guojing and Lyngaas, Isaac and Anthony, Quentin},
  booktitle={2024 IEEE International Parallel and Distributed Processing Symposium (IPDPS)},
  pages={556--569},
  year={2024},
  organization={IEEE}
}

@article{chen2024automated,
  title={Automated data curation for robust language model fine-tuning},
  author={Chen, Jiuhai and Mueller, Jonas},
  journal={arXiv preprint arXiv:2403.12776},
  year={2024}
}

@article{yin2026fairness,
  title={Fairness Definitions in Language Models Explained},
  author={Yin, Zhipeng and Wang, Zichong and Palikhe, Avash and Zhang, Wenbin},
  journal={Wiley Interdisciplinary Reviews: Data Mining and Knowledge Discovery},
  volume={16},
  number={1},
  pages={e70063},
  year={2026},
  publisher={Wiley Online Library}
}

@article{manerba2023social,
  title={Social bias probing: Fairness benchmarking for language models},
  author={Manerba, Marta Marchiori and Sta{\'n}czak, Karolina and Guidotti, Riccardo and Augenstein, Isabelle},
  journal={arXiv preprint arXiv:2311.09090},
  year={2023}
}

@article{koo2023benchmarking,
  title={Benchmarking cognitive biases in large language models as evaluators},
  author={Koo, Ryan and Lee, Minhwa and Raheja, Vipul and Park, Jong Inn and Kim, Zae Myung and Kang, Dongyeop},
  journal={arXiv preprint arXiv:2309.17012},
  year={2023}
}

@article{calian2025datarater,
  title={{DataRater}: Meta-Learned Dataset Curation},
  author={Calian, Dan A and Farquhar, Gregory and Kemaev, Iurii and Zintgraf, Luisa M and Hessel, Matteo and Shar, Jeremy and Oh, Junhyuk and Gy{\"o}rgy, Andr{\'a}s and Schaul, Tom and Dean, Jeffrey and others},
  journal={arXiv preprint arXiv:2505.17895},
  year={2025}
}

@article{shi2025route,
  title={Route sparse autoencoder to interpret large language models},
  author={Shi, Wei and Li, Sihang and Liang, Tao and Wan, Mingyang and Ma, Guojun and Wang, Xiang and He, Xiangnan},
  journal={arXiv preprint arXiv:2503.08200},
  year={2025}
}

@article{li2024crowdsourced,
  title={From crowdsourced data to high-quality benchmarks: Arena-hard and benchbuilder pipeline},
  author={Li, Tianle and Chiang, Wei-Lin and Frick, Evan and Dunlap, Lisa and Wu, Tianhao and Zhu, Banghua and Gonzalez, Joseph E and Stoica, Ion},
  journal={arXiv preprint arXiv:2406.11939},
  year={2024}
}

@article{miller2025evaluating,
  title={Evaluating {LLM} metrics through real-world capabilities},
  author={Miller, Justin K and Tang, Wenjia},
  journal={arXiv preprint arXiv:2505.08253},
  year={2025}
}

@article{mizrahi2024state,
  title={State of what art? A call for multi-prompt {LLM} evaluation},
  author={Mizrahi, Moran and Kaplan, Guy and Malkin, Dan and Dror, Rotem and Shahaf, Dafna and Stanovsky, Gabriel},
  journal={Transactions of the Association for Computational Linguistics},
  volume={12},
  pages={933--949},
  year={2024},
  publisher={MIT Press 255 Main Street, 9th Floor, Cambridge, Massachusetts 02142, USA}
}

@article{chang2024survey,
  title={A survey on evaluation of large language models},
  author={Chang, Yupeng and Wang, Xu and Wang, Jindong and Wu, Yuan and Yang, Linyi and Zhu, Kaijie and Chen, Hao and Yi, Xiaoyuan and Wang, Cunxiang and Wang, Yidong and others},
  journal={ACM transactions on intelligent systems and technology},
  volume={15},
  number={3},
  pages={1--45},
  year={2024},
  publisher={ACM New York, NY}
}

@article{elhage2022toy,
  title={Toy models of superposition},
  author={Elhage, Nelson and Hume, Tristan and Olsson, Catherine and Schiefer, Nicholas and Henighan, Tom and Kravec, Shauna and Hatfield-Dodds, Zac and Lasenby, Robert and Drain, Dawn and Chen, Carol and others},
  journal={arXiv preprint arXiv:2209.10652},
  year={2022}
}

@article{balcells2024evolution,
  title={Evolution of {SAE} features across layers in {LLMs}},
  author={Balcells, Daniel and Lerner, Benjamin and Oesterle, Michael and Ucar, Ediz and Heimersheim, Stefan},
  journal={arXiv preprint arXiv:2410.08869},
  year={2024}
}

@article{mcgrath2024understanding,
  author = {McGrath, Thomas and Balsam, Daniel and Deng, Myra and Ho, Eric},
  title = {Understanding and Steering {Llama 3} with Sparse Autoencoders},
  journal = {Goodfire Research},
  year = {2024},
  month = {September}
}

@article{guo2023evaluating,
  title={Evaluating large language models: A comprehensive survey},
  author={Guo, Zishan and Jin, Renren and Liu, Chuang and Huang, Yufei and Shi, Dan and Yu, Linhao and Liu, Yan and Li, Jiaxuan and Xiong, Bojian and Xiong, Deyi and others},
  journal={arXiv preprint arXiv:2310.19736},
  year={2023}
}

@misc{hardt2025emerging,
  author = {Moritz Hardt},
  title = {The Emerging Science of Machine Learning Benchmarks},
  year = {2025},
  howpublished = {Online at \url{https://mlbenchmarks.org}},
  note = {Manuscript}
}

@article{burnell2023rethink,
  title={Rethink reporting of evaluation results in {AI}},
  author={Burnell, Ryan and Schellaert, Wout and Burden, John and Ullman, Tomer D and Martinez-Plumed, Fernando and Tenenbaum, Joshua B and Rutar, Danaja and Cheke, Lucy G and Sohl-Dickstein, Jascha and Mitchell, Melanie and others},
  journal={Science},
  volume={380},
  number={6641},
  pages={136--138},
  year={2023},
  publisher={American Association for the Advancement of Science}
}

@article{didolkar2024metacognitive,
  title={Metacognitive capabilities of {LLMs}: An exploration in mathematical problem solving},
  author={Didolkar, Aniket and Goyal, Anirudh and Ke, Nan Rosemary and Guo, Siyuan and Valko, Michal and Lillicrap, Timothy and Jimenez Rezende, Danilo and Bengio, Yoshua and Mozer, Michael C and Arora, Sanjeev},
  journal={Advances in Neural Information Processing Systems},
  volume={37},
  pages={19783--19812},
  year={2024}
}

@misc{lieberum_gemma_2024,
	title = {{Gemma} Scope: Open Sparse Autoencoders Everywhere All At Once on {Gemma} 2},
	shorttitle = {Gemma {Scope}},
	url = {http://arxiv.org/abs/2408.05147},
	doi = {10.48550/arXiv.2408.05147},
	urldate = {2025-09-23},
	publisher = {arXiv},
	author = {Lieberum, Tom and Rajamanoharan, Senthooran and Conmy, Arthur and Smith, Lewis and Sonnerat, Nicolas and Varma, Vikrant and Kramár, János and Dragan, Anca and Shah, Rohin and Nanda, Neel},
	month = aug,
	year = {2024},
	note = {arXiv:2408.05147},
}

@misc{balsam2025announcing,
    title = {Announcing Open-Source {SAEs} for {Llama} 3.3 {70B} and {Llama} 3.1 {8B}},
    author = {Balsam, Daniel and McGrath, Thomas and Gorton, Liv and Nguyen, Nam and Deng, Myra and Ho, Eric},
    year = {2025},
    url = {https://www.goodfire.ai/blog/sae-open-source-announcement},
    language = {en},
}

@misc{gemma2024,
	title = {{Gemma} 2: Improving Open Language Models  at a Practical Size},
	shorttitle = {Gemma 2},
	url = {http://arxiv.org/abs/2408.00118},
	doi = {10.48550/arXiv.2408.00118},
	urldate = {2025-09-23},
	publisher = {arXiv},
	author = {Gemma Team and Riviere, Morgane and Pathak, Shreya and Sessa, Pier Giuseppe and Hardin, Cassidy and Bhupatiraju, Surya and Hussenot, Léonard and Mesnard, Thomas and Shahriari, Bobak and Ramé, Alexandre and Ferret, Johan and Liu, Peter and Tafti, Pouya and Friesen, Abe and Casbon, Michelle and Ramos, Sabela and Kumar, Ravin and Lan, Charline Le and Jerome, Sammy and Tsitsulin, Anton and Vieillard, Nino and Stanczyk, Piotr and Girgin, Sertan and Momchev, Nikola and Hoffman, Matt and Thakoor, Shantanu and Grill, Jean-Bastien and Neyshabur, Behnam and Bachem, Olivier and Walton, Alanna and Severyn, Aliaksei and Parrish, Alicia and Ahmad, Aliya and Hutchison, Allen and Abdagic, Alvin and Carl, Amanda and Shen, Amy and Brock, Andy and Coenen, Andy and Laforge, Anthony and Paterson, Antonia and Bastian, Ben and Piot, Bilal and Wu, Bo and Royal, Brandon and Chen, Charlie and Kumar, Chintu and Perry, Chris and Welty, Chris and Choquette-Choo, Christopher A. and Sinopalnikov, Danila and Weinberger, David and Vijaykumar, Dimple and Rogozińska, Dominika and Herbison, Dustin and Bandy, Elisa and Wang, Emma and Noland, Eric and Moreira, Erica and Senter, Evan and Eltyshev, Evgenii and Visin, Francesco and Rasskin, Gabriel and Wei, Gary and Cameron, Glenn and Martins, Gus and Hashemi, Hadi and Klimczak-Plucińska, Hanna and Batra, Harleen and Dhand, Harsh and Nardini, Ivan and Mein, Jacinda and Zhou, Jack and Svensson, James and Stanway, Jeff and Chan, Jetha and Zhou, Jin Peng and Carrasqueira, Joana and Iljazi, Joana and Becker, Jocelyn and Fernandez, Joe and Amersfoort, Joost van and Gordon, Josh and Lipschultz, Josh and Newlan, Josh and Ji, Ju-yeong and Mohamed, Kareem and Badola, Kartikeya and Black, Kat and Millican, Katie and McDonell, Keelin and Nguyen, Kelvin and Sodhia, Kiranbir and Greene, Kish and Sjoesund, Lars Lowe and Usui, Lauren and Sifre, Laurent and Heuermann, Lena and Lago, Leticia and McNealus, Lilly and Soares, Livio Baldini and Kilpatrick, Logan and Dixon, Lucas and Martins, Luciano and Reid, Machel and Singh, Manvinder and Iverson, Mark and Görner, Martin and Velloso, Mat and Wirth, Mateo and Davidow, Matt and Miller, Matt and Rahtz, Matthew and Watson, Matthew and Risdal, Meg and Kazemi, Mehran and Moynihan, Michael and Zhang, Ming and Kahng, Minsuk and Park, Minwoo and Rahman, Mofi and Khatwani, Mohit and Dao, Natalie and Bardoliwalla, Nenshad and Devanathan, Nesh and Dumai, Neta and Chauhan, Nilay and Wahltinez, Oscar and Botarda, Pankil and Barnes, Parker and Barham, Paul and Michel, Paul and Jin, Pengchong and Georgiev, Petko and Culliton, Phil and Kuppala, Pradeep and Comanescu, Ramona and Merhej, Ramona and Jana, Reena and Rokni, Reza Ardeshir and Agarwal, Rishabh and Mullins, Ryan and Saadat, Samaneh and Carthy, Sara Mc and Cogan, Sarah and Perrin, Sarah and Arnold, Sébastien M. R. and Krause, Sebastian and Dai, Shengyang and Garg, Shruti and Sheth, Shruti and Ronstrom, Sue and Chan, Susan and Jordan, Timothy and Yu, Ting and Eccles, Tom and Hennigan, Tom and Kocisky, Tomas and Doshi, Tulsee and Jain, Vihan and Yadav, Vikas and Meshram, Vilobh and Dharmadhikari, Vishal and Barkley, Warren and Wei, Wei and Ye, Wenming and Han, Woohyun and Kwon, Woosuk and Xu, Xiang and Shen, Zhe and Gong, Zhitao and Wei, Zichuan and Cotruta, Victor and Kirk, Phoebe and Rao, Anand and Giang, Minh and Peran, Ludovic and Warkentin, Tris and Collins, Eli and Barral, Joelle and Ghahramani, Zoubin and Hadsell, Raia and Sculley, D. and Banks, Jeanine and Dragan, Anca and Petrov, Slav and Vinyals, Oriol and Dean, Jeff and Hassabis, Demis and Kavukcuoglu, Koray and Farabet, Clement and Buchatskaya, Elena and Borgeaud, Sebastian and Fiedel, Noah and Joulin, Armand and Kenealy, Kathleen and Dadashi, Robert and Andreev, Alek},
	month = oct,
	year = {2024},
	note = {arXiv:2408.00118},
}

@misc{grattafiori2024llama,
	title = {The {Llama} 3 Herd of Models},
	url = {http://arxiv.org/abs/2407.21783},
	doi = {10.48550/arXiv.2407.21783},
	urldate = {2025-03-29},
	publisher = {arXiv},
	author = {Grattafiori, Aaron and Dubey, Abhimanyu and Jauhri, Abhinav and Pandey, Abhinav and Kadian, Abhishek and Al-Dahle, Ahmad and Letman, Aiesha and Mathur, Akhil and Schelten, Alan and Vaughan, Alex and Yang, Amy and Fan, Angela and Goyal, Anirudh and Hartshorn, Anthony and Yang, Aobo and Mitra, Archi and Sravankumar, Archie and Korenev, Artem and Hinsvark, Arthur and Rao, Arun and Zhang, Aston and Rodriguez, Aurelien and Gregerson, Austen and Spataru, Ava and Roziere, Baptiste and Biron, Bethany and Tang, Binh and Chern, Bobbie and Caucheteux, Charlotte and Nayak, Chaya and Bi, Chloe and Marra, Chris and McConnell, Chris and Keller, Christian and Touret, Christophe and Wu, Chunyang and Wong, Corinne and Ferrer, Cristian Canton and Nikolaidis, Cyrus and Allonsius, Damien and Song, Daniel and Pintz, Danielle and Livshits, Danny and Wyatt, Danny and Esiobu, David and Choudhary, Dhruv and Mahajan, Dhruv and Garcia-Olano, Diego and Perino, Diego and Hupkes, Dieuwke and Lakomkin, Egor and AlBadawy, Ehab and Lobanova, Elina and Dinan, Emily and Smith, Eric Michael and Radenovic, Filip and Guzmán, Francisco and Zhang, Frank and Synnaeve, Gabriel and Lee, Gabrielle and Anderson, Georgia Lewis and Thattai, Govind and Nail, Graeme and Mialon, Gregoire and Pang, Guan and Cucurell, Guillem and Nguyen, Hailey and Korevaar, Hannah and Xu, Hu and Touvron, Hugo and Zarov, Iliyan and Ibarra, Imanol Arrieta and Kloumann, Isabel and Misra, Ishan and Evtimov, Ivan and Zhang, Jack and Copet, Jade and Lee, Jaewon and Geffert, Jan and Vranes, Jana and Park, Jason and Mahadeokar, Jay and Shah, Jeet and Linde, Jelmer van der and Billock, Jennifer and Hong, Jenny and Lee, Jenya and Fu, Jeremy and Chi, Jianfeng and Huang, Jianyu and Liu, Jiawen and Wang, Jie and Yu, Jiecao and Bitton, Joanna and Spisak, Joe and Park, Jongsoo and Rocca, Joseph and Johnstun, Joshua and Saxe, Joshua and Jia, Junteng and Alwala, Kalyan Vasuden and Prasad, Karthik and Upasani, Kartikeya and Plawiak, Kate and Li, Ke and Heafield, Kenneth and Stone, Kevin and El-Arini, Khalid and Iyer, Krithika and Malik, Kshitiz and Chiu, Kuenley and Bhalla, Kunal and Lakhotia, Kushal and Rantala-Yeary, Lauren and Maaten, Laurens van der and Chen, Lawrence and Tan, Liang and Jenkins, Liz and Martin, Louis and Madaan, Lovish and Malo, Lubo and Blecher, Lukas and Landzaat, Lukas and Oliveira, Luke de and Muzzi, Madeline and Pasupuleti, Mahesh and Singh, Mannat and Paluri, Manohar and Kardas, Marcin and Tsimpoukelli, Maria and Oldham, Mathew and Rita, Mathieu and Pavlova, Maya and Kambadur, Melanie and Lewis, Mike and Si, Min and Singh, Mitesh Kumar and Hassan, Mona and Goyal, Naman and Torabi, Narjes and Bashlykov, Nikolay and Bogoychev, Nikolay and Chatterji, Niladri and Zhang, Ning and Duchenne, Olivier and Çelebi, Onur and Alrassy, Patrick and Zhang, Pengchuan and Li, Pengwei and Vasic, Petar and Weng, Peter and Bhargava, Prajjwal and Dubal, Pratik and Krishnan, Praveen and Koura, Punit Singh and Xu, Puxin and He, Qing and Dong, Qingxiao and Srinivasan, Ragavan and Ganapathy, Raj and Calderer, Ramon and Cabral, Ricardo Silveira and Stojnic, Robert and Raileanu, Roberta and Maheswari, Rohan and Girdhar, Rohit and Patel, Rohit and Sauvestre, Romain and Polidoro, Ronnie and Sumbaly, Roshan and Taylor, Ross and Silva, Ruan and Hou, Rui and Wang, Rui and Hosseini, Saghar and Chennabasappa, Sahana and Singh, Sanjay and Bell, Sean and Kim, Seohyun Sonia and Edunov, Sergey and Nie, Shaoliang and Narang, Sharan and Raparthy, Sharath and Shen, Sheng and Wan, Shengye and Bhosale, Shruti and Zhang, Shun and Vandenhende, Simon and Batra, Soumya and Whitman, Spencer and Sootla, Sten and Collot, Stephane and Gururangan, Suchin and Borodinsky, Sydney and Herman, Tamar and Fowler, Tara and Sheasha, Tarek and Georgiou, Thomas and Scialom, Thomas and Speckbacher, Tobias and Mihaylov, Todor and Xiao, Tong and Karn, Ujjwal and Goswami, Vedanuj and Gupta, Vibhor and Ramanathan, Vignesh and Kerkez, Viktor and Gonguet, Vincent and Do, Virginie and Vogeti, Vish and Albiero, Vítor and Petrovic, Vladan and Chu, Weiwei and Xiong, Wenhan and Fu, Wenyin and Meers, Whitney and Martinet, Xavier and Wang, Xiaodong and Wang, Xiaofang and Tan, Xiaoqing Ellen and Xia, Xide and Xie, Xinfeng and Jia, Xuchao and Wang, Xuewei and Goldschlag, Yaelle and Gaur, Yashesh and Babaei, Yasmine and Wen, Yi and Song, Yiwen and Zhang, Yuchen and Li, Yue and Mao, Yuning and Coudert, Zacharie Delpierre and Yan, Zheng and Chen, Zhengxing and Papakipos, Zoe and Singh, Aaditya and Srivastava, Aayushi and Jain, Abha and Kelsey, Adam and Shajnfeld, Adam and Gangidi, Adithya and Victoria, Adolfo and Goldstand, Ahuva and Menon, Ajay and Sharma, Ajay and Boesenberg, Alex and Baevski, Alexei and Feinstein, Allie and Kallet, Amanda and Sangani, Amit and Teo, Amos and Yunus, Anam and Lupu, Andrei and Alvarado, Andres and Caples, Andrew and Gu, Andrew and Ho, Andrew and Poulton, Andrew and Ryan, Andrew and Ramchandani, Ankit and Dong, Annie and Franco, Annie and Goyal, Anuj and Saraf, Aparajita and Chowdhury, Arkabandhu and Gabriel, Ashley and Bharambe, Ashwin and Eisenman, Assaf and Yazdan, Azadeh and James, Beau and Maurer, Ben and Leonhardi, Benjamin and Huang, Bernie and Loyd, Beth and Paola, Beto De and Paranjape, Bhargavi and Liu, Bing and Wu, Bo and Ni, Boyu and Hancock, Braden and Wasti, Bram and Spence, Brandon and Stojkovic, Brani and Gamido, Brian and Montalvo, Britt and Parker, Carl and Burton, Carly and Mejia, Catalina and Liu, Ce and Wang, Changhan and Kim, Changkyu and Zhou, Chao and Hu, Chester and Chu, Ching-Hsiang and Cai, Chris and Tindal, Chris and Feichtenhofer, Christoph and Gao, Cynthia and Civin, Damon and Beaty, Dana and Kreymer, Daniel and Li, Daniel and Adkins, David and Xu, David and Testuggine, Davide and David, Delia and Parikh, Devi and Liskovich, Diana and Foss, Didem and Wang, Dingkang and Le, Duc and Holland, Dustin and Dowling, Edward and Jamil, Eissa and Montgomery, Elaine and Presani, Eleonora and Hahn, Emily and Wood, Emily and Le, Eric-Tuan and Brinkman, Erik and Arcaute, Esteban and Dunbar, Evan and Smothers, Evan and Sun, Fei and Kreuk, Felix and Tian, Feng and Kokkinos, Filippos and Ozgenel, Firat and Caggioni, Francesco and Kanayet, Frank and Seide, Frank and Florez, Gabriela Medina and Schwarz, Gabriella and Badeer, Gada and Swee, Georgia and Halpern, Gil and Herman, Grant and Sizov, Grigory and Guangyi and Zhang and Lakshminarayanan, Guna and Inan, Hakan and Shojanazeri, Hamid and Zou, Han and Wang, Hannah and Zha, Hanwen and Habeeb, Haroun and Rudolph, Harrison and Suk, Helen and Aspegren, Henry and Goldman, Hunter and Zhan, Hongyuan and Damlaj, Ibrahim and Molybog, Igor and Tufanov, Igor and Leontiadis, Ilias and Veliche, Irina-Elena and Gat, Itai and Weissman, Jake and Geboski, James and Kohli, James and Lam, Janice and Asher, Japhet and Gaya, Jean-Baptiste and Marcus, Jeff and Tang, Jeff and Chan, Jennifer and Zhen, Jenny and Reizenstein, Jeremy and Teboul, Jeremy and Zhong, Jessica and Jin, Jian and Yang, Jingyi and Cummings, Joe and Carvill, Jon and Shepard, Jon and McPhie, Jonathan and Torres, Jonathan and Ginsburg, Josh and Wang, Junjie and Wu, Kai and U, Kam Hou and Saxena, Karan and Khandelwal, Kartikay and Zand, Katayoun and Matosich, Kathy and Veeraraghavan, Kaushik and Michelena, Kelly and Li, Keqian and Jagadeesh, Kiran and Huang, Kun and Chawla, Kunal and Huang, Kyle and Chen, Lailin and Garg, Lakshya and A, Lavender and Silva, Leandro and Bell, Lee and Zhang, Lei and Guo, Liangpeng and Yu, Licheng and Moshkovich, Liron and Wehrstedt, Luca and Khabsa, Madian and Avalani, Manav and Bhatt, Manish and Mankus, Martynas and Hasson, Matan and Lennie, Matthew and Reso, Matthias and Groshev, Maxim and Naumov, Maxim and Lathi, Maya and Keneally, Meghan and Liu, Miao and Seltzer, Michael L. and Valko, Michal and Restrepo, Michelle and Patel, Mihir and Vyatskov, Mik and Samvelyan, Mikayel and Clark, Mike and Macey, Mike and Wang, Mike and Hermoso, Miquel Jubert and Metanat, Mo and Rastegari, Mohammad and Bansal, Munish and Santhanam, Nandhini and Parks, Natascha and White, Natasha and Bawa, Navyata and Singhal, Nayan and Egebo, Nick and Usunier, Nicolas and Mehta, Nikhil and Laptev, Nikolay Pavlovich and Dong, Ning and Cheng, Norman and Chernoguz, Oleg and Hart, Olivia and Salpekar, Omkar and Kalinli, Ozlem and Kent, Parkin and Parekh, Parth and Saab, Paul and Balaji, Pavan and Rittner, Pedro and Bontrager, Philip and Roux, Pierre and Dollar, Piotr and Zvyagina, Polina and Ratanchandani, Prashant and Yuvraj, Pritish and Liang, Qian and Alao, Rachad and Rodriguez, Rachel and Ayub, Rafi and Murthy, Raghotham and Nayani, Raghu and Mitra, Rahul and Parthasarathy, Rangaprabhu and Li, Raymond and Hogan, Rebekkah and Battey, Robin and Wang, Rocky and Howes, Russ and Rinott, Ruty and Mehta, Sachin and Siby, Sachin and Bondu, Sai Jayesh and Datta, Samyak and Chugh, Sara and Hunt, Sara and Dhillon, Sargun and Sidorov, Sasha and Pan, Satadru and Mahajan, Saurabh and Verma, Saurabh and Yamamoto, Seiji and Ramaswamy, Sharadh and Lindsay, Shaun and Lindsay, Shaun and Feng, Sheng and Lin, Shenghao and Zha, Shengxin Cindy and Patil, Shishir and Shankar, Shiva and Zhang, Shuqiang and Zhang, Shuqiang and Wang, Sinong and Agarwal, Sneha and Sajuyigbe, Soji and Chintala, Soumith and Max, Stephanie and Chen, Stephen and Kehoe, Steve and Satterfield, Steve and Govindaprasad, Sudarshan and Gupta, Sumit and Deng, Summer and Cho, Sungmin and Virk, Sunny and Subramanian, Suraj and Choudhury, Sy and Goldman, Sydney and Remez, Tal and Glaser, Tamar and Best, Tamara and Koehler, Thilo and Robinson, Thomas and Li, Tianhe and Zhang, Tianjun and Matthews, Tim and Chou, Timothy and Shaked, Tzook and Vontimitta, Varun and Ajayi, Victoria and Montanez, Victoria and Mohan, Vijai and Kumar, Vinay Satish and Mangla, Vishal and Ionescu, Vlad and Poenaru, Vlad and Mihailescu, Vlad Tiberiu and Ivanov, Vladimir and Li, Wei and Wang, Wenchen and Jiang, Wenwen and Bouaziz, Wes and Constable, Will and Tang, Xiaocheng and Wu, Xiaojian and Wang, Xiaolan and Wu, Xilun and Gao, Xinbo and Kleinman, Yaniv and Chen, Yanjun and Hu, Ye and Jia, Ye and Qi, Ye and Li, Yenda and Zhang, Yilin and Zhang, Ying and Adi, Yossi and Nam, Youngjin and Yu and Wang and Zhao, Yu and Hao, Yuchen and Qian, Yundi and Li, Yunlu and He, Yuzi and Rait, Zach and DeVito, Zachary and Rosnbrick, Zef and Wen, Zhaoduo and Yang, Zhenyu and Zhao, Zhiwei and Ma, Zhiyu},
	month = nov,
	year = {2024},
	note = {arXiv:2407.21783},
}

@inproceedings{rein2024gpqa,
  title={{GPQA}: A graduate-level {Google}-proof {Q\&A} benchmark},
  author={Rein, David and Hou, Betty Li and Stickland, Asa Cooper and Petty, Jackson and Pang, Richard Yuanzhe and Dirani, Julien and Michael, Julian and Bowman, Samuel R},
  booktitle={First Conference on Language Modeling},
  year={2024}
}

@inproceedings{perez2023discovering,
  title={Discovering language model behaviors with model-written evaluations},
  author={Perez, Ethan and Ringer, Sam and Lukosiute, Kamile and Nguyen, Karina and Chen, Edwin and Heiner, Scott and Pettit, Craig and Olsson, Catherine and Kundu, Sandipan and Kadavath, Saurav and others},
  booktitle={Findings of the association for computational linguistics: ACL 2023},
  pages={13387--13434},
  year={2023}
}

@article{zheng2023judging,
  title={Judging {LLM}-as-a-judge with {MT-Bench} and chatbot arena},
  author={Zheng, Lianmin and Chiang, Wei-Lin and Sheng, Ying and Zhuang, Siyuan and Wu, Zhanghao and Zhuang, Yonghao and Lin, Zi and Li, Zhuohan and Li, Dacheng and Xing, Eric and others},
  journal={Advances in neural information processing systems},
  volume={36},
  pages={46595--46623},
  year={2023}
}

@inproceedings{menon2025analyzing,
  title={Analyzing (in)abilities of {SAEs} via formal languages},
  author={Menon, Abhinav and Shrivastava, Manish and Krueger, David and Lubana, Ekdeep Singh},
  booktitle={Proceedings of the 2025 Conference of the Nations of the Americas Chapter of the Association for Computational Linguistics: Human Language Technologies (Volume 1: Long Papers)},
  pages={4837--4862},
  year={2025}
}

@article{paulo2025sparse,
  title={Sparse autoencoders trained on the same data learn different features},
  author={Paulo, Gon{\c{c}}alo and Belrose, Nora},
  journal={arXiv preprint arXiv:2501.16615},
  year={2025}
}

@article{hendrycks2020measuring,
  title={Measuring massive multitask language understanding},
  author={Hendrycks, Dan and Burns, Collin and Basart, Steven and Zou, Andy and Mazeika, Mantas and Song, Dawn and Steinhardt, Jacob},
  journal={arXiv preprint arXiv:2009.03300},
  year={2020}
}

@inproceedings{jimenez2023swebench,
  title={{SWE}-bench: Can language models resolve real-world {GitHub} issues?},
  author={Jimenez, Carlos E and Yang, John and Wettig, Alexander and Yao, Shunyu and Pei, Kexin and Press, Ofir and Narasimhan, Karthik},
  booktitle={International Conference on Learning Representations},
  volume={2024},
  pages={54107--54157},
  year={2024}
}

@article{guo2025deepseek,
  title={Deepseek-{R1}: Incentivizing reasoning capability in {LLMs} via reinforcement learning},
  author={Guo, Daya and Yang, Dejian and Zhang, Haowei and Song, Junxiao and Wang, Peiyi and Zhu, Qihao and Xu, Runxin and Zhang, Ruoyu and Ma, Shirong and Bi, Xiao and others},
  journal={arXiv preprint arXiv:2501.12948},
  year={2025}
}

@article{he2024llamascope,
  title={{Llama Scope}: Extracting millions of features from {Llama-3.1-8B} with sparse autoencoders},
  author={He, Zhengfu and Shu, Wentao and Ge, Xuyang and Chen, Lingjie and Wang, Junxuan and Zhou, Yunhua and Liu, Frances and Guo, Qipeng and Huang, Xuanjing and Wu, Zuxuan and others},
  journal={arXiv preprint arXiv:2410.20526},
  year={2024}
}

@article{jiang2023mistral,
  title={{Mistral 7B}},
  author={Jiang, Albert Q. and Sablayrolles, Alexandre and Mensch, Arthur and Bamford, Chris and Chaplot, Devendra Singh and de las Casas, Diego and Bressand, Florian and Lengyel, Gianna and Lample, Guillaume and Saulnier, Lucile and Lavaud, L{\'e}lio Renard and Lachaux, Marie-Anne and Stock, Pierre and Le Scao, Teven and Lavril, Thibaut and Wang, Thomas and Lacroix, Timoth{\'e}e and El Sayed, William},
  journal={arXiv preprint arXiv:2310.06825},
  year={2023}
}

@article{yang2025qwen3,
  title={Qwen3 technical report},
  author={Yang, An and Li, Anfeng and Yang, Baosong and Zhang, Beichen and Hui, Binyuan and Zheng, Bo and Yu, Bowen and Gao, Chang and Huang, Chengen and Lv, Chenxu and others},
  journal={arXiv preprint arXiv:2505.09388},
  year={2025}
}

@misc{cosgrove2024mistralsae,
  title={Mistral-{7B} Sparse Autoencoder (Layer 16)},
  author={Cosgrove, Tyler},
  year={2024},
  howpublished={\url{https://huggingface.co/tylercosgrove/mistral-7b-sparse-autoencoder-layer16}}
}

@misc{hanna2024qwen3transcoders,
  title={Qwen3-{4B} Transcoders},
  author={Hanna, Michael},
  year={2024},
  howpublished={\url{https://huggingface.co/mwhanna/qwen3-4b-transcoders}}
}

@article{zeng2025evaltree,
  title={{EvalTree}: Profiling language model weaknesses via hierarchical capability trees},
  author={Zeng, Zhiyuan and Wang, Yizhong and Hajishirzi, Hannaneh and Koh, Pang Wei},
  journal={arXiv preprint arXiv:2503.08893},
  year={2025}
}
\bibliographystyle{icml2026}

\newpage
\onecolumn

\addcontentsline{toc}{section}{Appendix}
{%
  \hypersetup{linkcolor=black, citecolor=black, filecolor=black, urlcolor=black}%
  \setcounter{secnumdepth}{-2}
  \part{Appendix}%
  \setcounter{secnumdepth}{3}
  \parttoc
}

\appendix


\newpage


\section{Sparse Autoencoder (SAE) Primer}
\label{app:sae_primer}

We summarize the SAE terminology used throughout the paper. A sparse autoencoder is trained to reconstruct the internal activations of a host model $M$ at a chosen layer, using an over-complete dictionary of $K$ features with an explicit sparsity penalty~\citep{bricken2023towards,cunningham2023sparse}. The goal is to recover monosemantic ``concepts'' that decompose superposed neuron activations into interpretable axes~\citep{elhage2022toy}.

We refer to each of the $K$ learned feature directions as a \emph{concept}, the full set as the \emph{concept dictionary} $C_{SAE}$, and the space they span as the \emph{SAE space}. The activation of concept $c$ on token $x_{i,j}$ is denoted $SAE(x_{i,j}, c) \in \mathbb{R}_{\ge 0}$. For each concept, an autointerpretability pipeline assigns a textual label $l_c$ by prompting an LLM with high-activation exemplars~\citep{mcgrath2024understanding}.

The dictionary size $K$ is not a hyperparameter we tune; it is fixed by the pre-trained SAE's \emph{expansion factor} over the host model's hidden dimension, with prior scaling work suggesting $K$ should grow with model size~\citep{templeton2024scaling,gao2024scaling}. The Goodfire SAE for Llama 3.1 8B uses $K \approx 65{,}536$ concepts; the Gemma Scope SAE for Gemma 2 2B uses $K \approx 16{,}384$. Larger dictionaries yield finer-grained concepts but leave our analysis pipeline unchanged.

Running an SAE at a single layer adds one linear projection of size $d_{\text{model}} \times K$ per token plus a sparsity step. The host LLM, by comparison, performs many such projections per layer across all of its layers, plus attention --- so the SAE contributes roughly one extra ``layer's worth'' of compute on top of the full forward pass, typically a few percent for an 8B model at a single residual-stream layer.


\newpage
\section{Additional Results: Gemma 2 2B Instruct}
\label{app:add_results_gemma_2_2b_it}

\subsection{Benchmark Gaps}

\begin{figure}[h]
    \centering
    \includegraphics[width=0.49\textwidth]{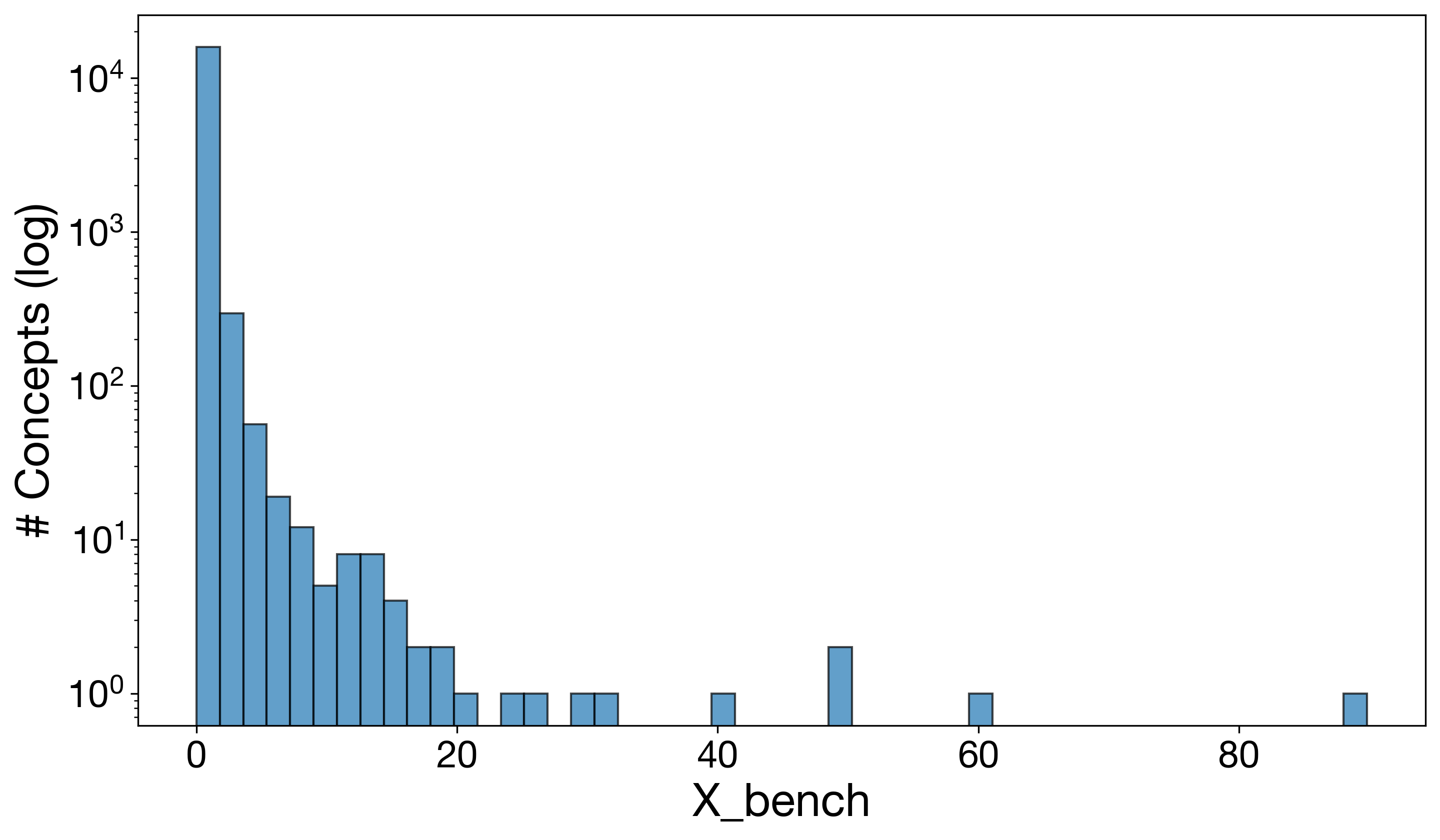} %
    \caption{\textbf{Cross-Benchmark Coverage.} The distribution of $\bm{\mathit{X}}_{\text{bench}}^{(c)}$ scores obtained for the $10$ evaluated benchmarks, using the SAE of Gemma 2 2B.}
    \label{fig:gemma_overall_coverage}
\end{figure}

\begin{figure*}[h]
    \centering
    \includegraphics[width=\linewidth]{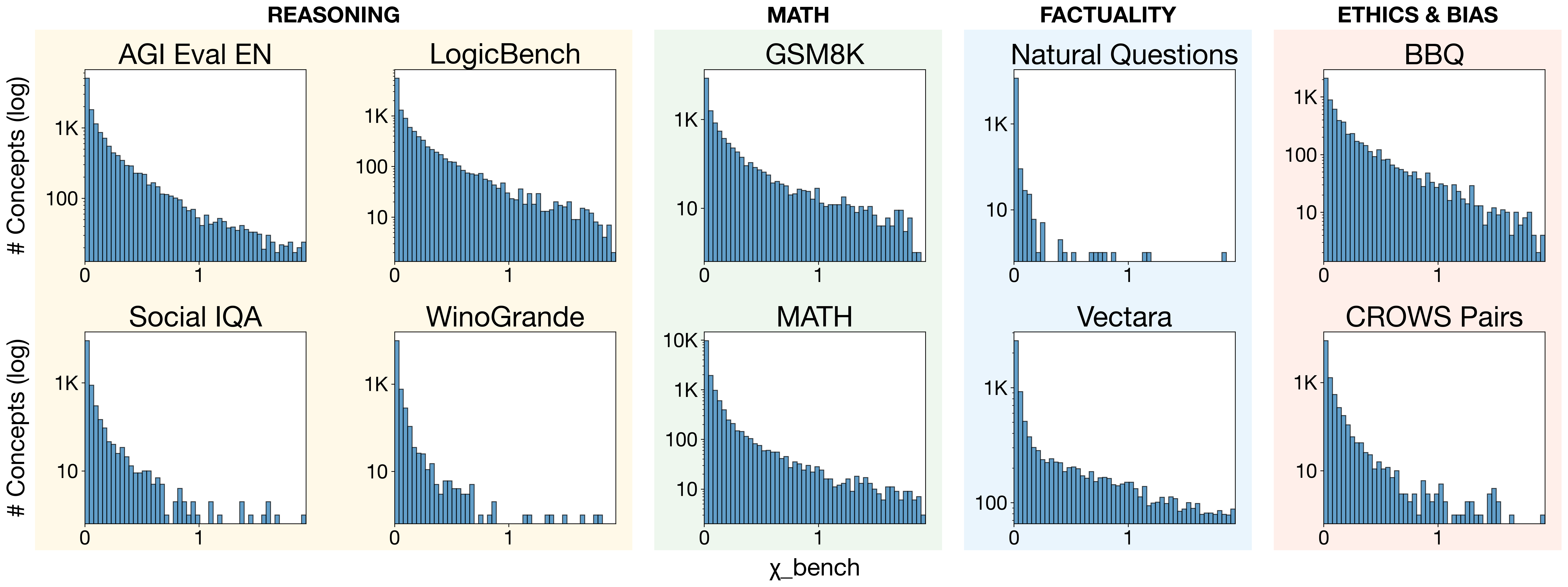} %
    \caption{\textbf{Coverage Within Individual Benchmarks.} A breakdown of $\chi_{\text{bench}}^{(b,c)}$ score distributions for individual benchmarks obtained via Gemma 2 2B.}
    \label{fig:gemma_bench_lvl_coverage}
\end{figure*}

\begin{figure}[h]
    \centering
    \includegraphics[width=0.49\textwidth]{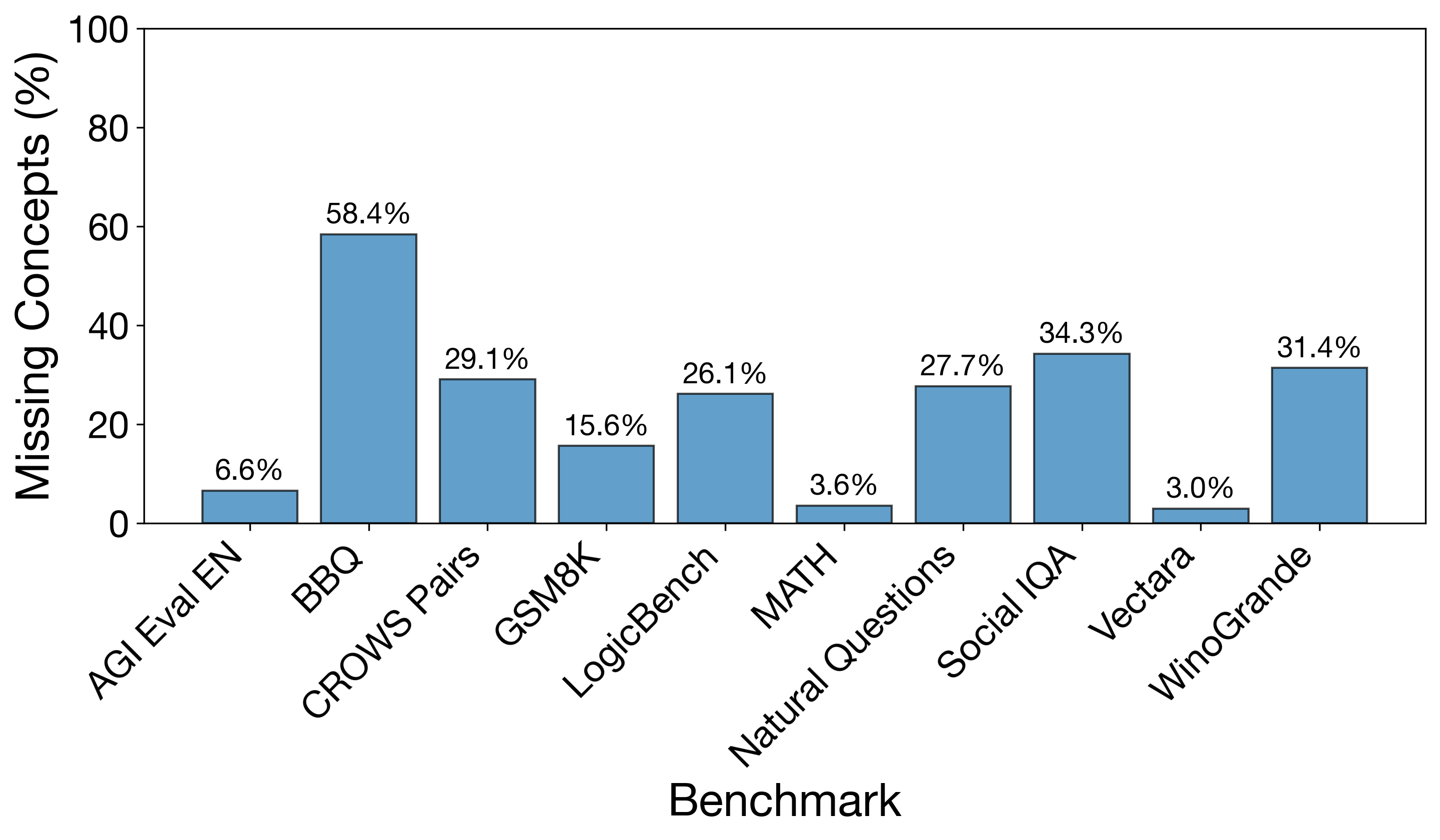} %
    \caption{\textbf{Missing Concepts.} Proportion of the SAE concept dictionary for Gemma 2 2B that is not tested by the respective benchmarks.}
    \label{fig:gemma_missing_concept_ratio}
\end{figure}

\begin{table}[h]
    \centering
    \caption{\textbf{Examples of Missing Concepts from the Full Benchmark Suite.}}
    \label{tab:gemma_missing_concepts_full_suite}
    \resizebox{0.8\textwidth}{!}{ 
    \begin{tabular}{cl}
    \toprule
    \textbf{Concept ID} & \textbf{Concept Description} \\
    \toprule
    \addlinespace[0.6ex]
    \conc{\textbf{(5169)}} & patterns of repeated characters or symbols \\
    \cmidrule{1-2}
    \conc{\textbf{(3674)}} & phrases related to coding or programming syntax \\
    \toprule
    \conc{\textbf{(9514)}} & detailed references and citations in academic writing \\
    \cmidrule{1-2}
    \conc{\textbf{(10108)}} & questions and references to uncertainty or confusion \\
    \cmidrule{1-2}
    \conc{\textbf{(5102)}} & specific programming or technical terminology related to data storage and handling \\
    \addlinespace[0.6ex]
    \bottomrule
    \end{tabular}
    }
    
\end{table}

\begin{table}[h]
\centering
\caption{\textbf{Examples of Missing Concepts from Individual Benchmarks.}}
\label{tab:gemma_missing_concepts}
\resizebox{0.93\textwidth}{!}{ 
\begin{tabular}{lcl}
\toprule
\textbf{Benchmark} & \textbf{Concept ID} & \textbf{Concept Description} \\
\toprule
\addlinespace[0.6ex]
\textit{AGI Eval} & \conc{\textbf{(12792)}} & phrases related to problem-solving or troubleshooting \\
\cmidrule{2-3}
& \conc{\textbf{(12436)}} & error handling and debugging statements in programming code \\
\addlinespace[0.6ex]
\toprule
 \textit{LogicBench} & \conc{\textbf{(7873)}} & phrases that indicate conditions or states related to certainty or necessity \\
\cmidrule{2-3}
& \conc{\textbf{(7264)}} & occurrences of mathematical or formal logic terms and control structures in the text \\
\addlinespace[0.6ex]
\toprule
\textit{Social IQA} & \conc{\textbf{(2863)}} & concepts related to social dynamics and collaborative efforts \\
\cmidrule{2-3}
& \conc{\textbf{(12897)}} & concepts related to socio-cultural analysis and individualized experiences \\
\bottomrule
\end{tabular}
}

\end{table}

\FloatBarrier

\subsection{Model Gaps}

\begin{figure}[h]
    \centering
    \includegraphics[width=0.49\textwidth]{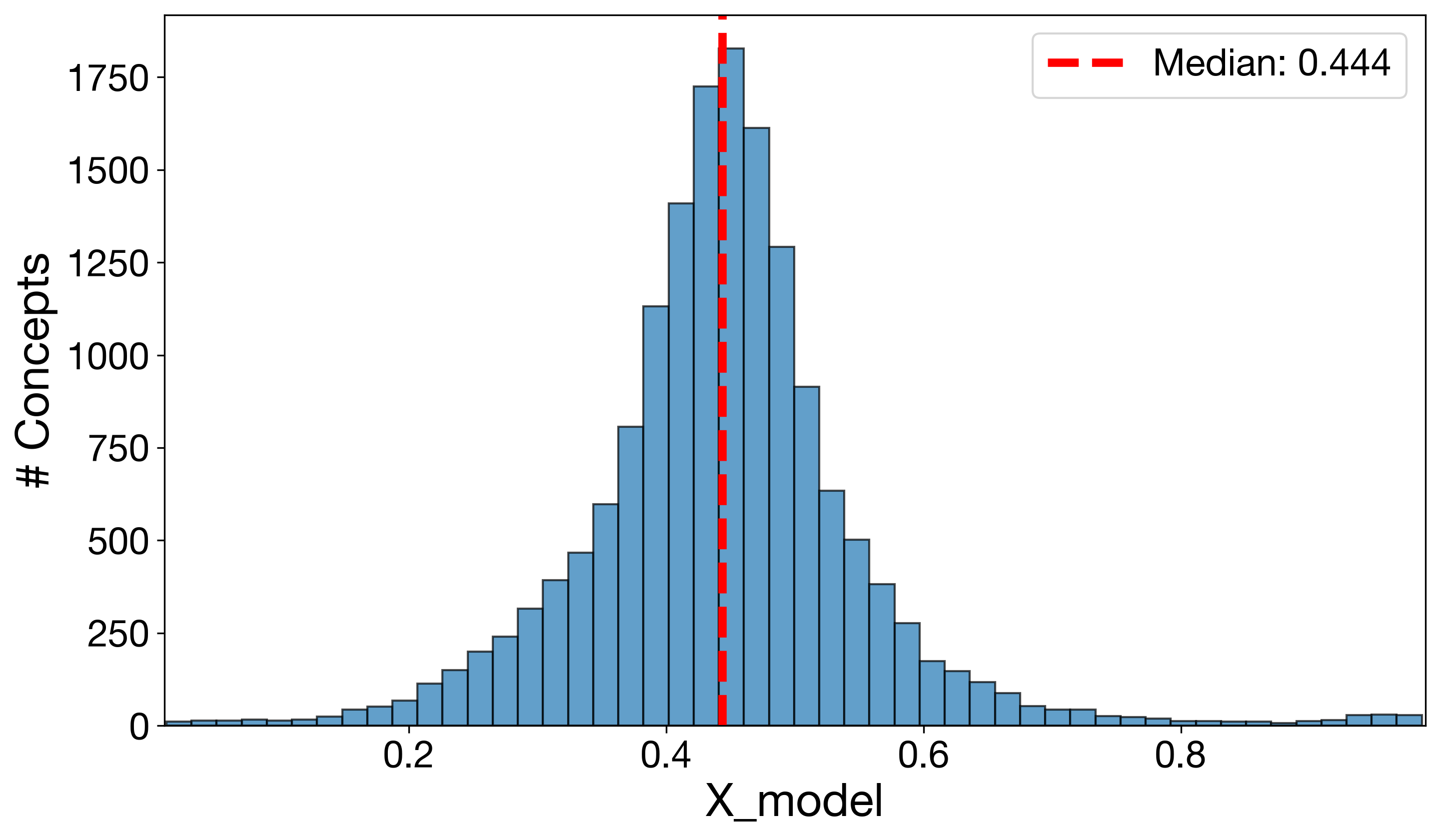} %
    \caption{\textbf{Cross-Benchmark Performance.} The distribution of $\bm{\mathit{X}}_{\text{model}}^{(c)}$ scores obtained for for Gemma 2 2B.}
    \label{fig:gemma_overall_performance}
\end{figure}

\begin{figure*}[h]
    \centering
    \includegraphics[width=\linewidth]{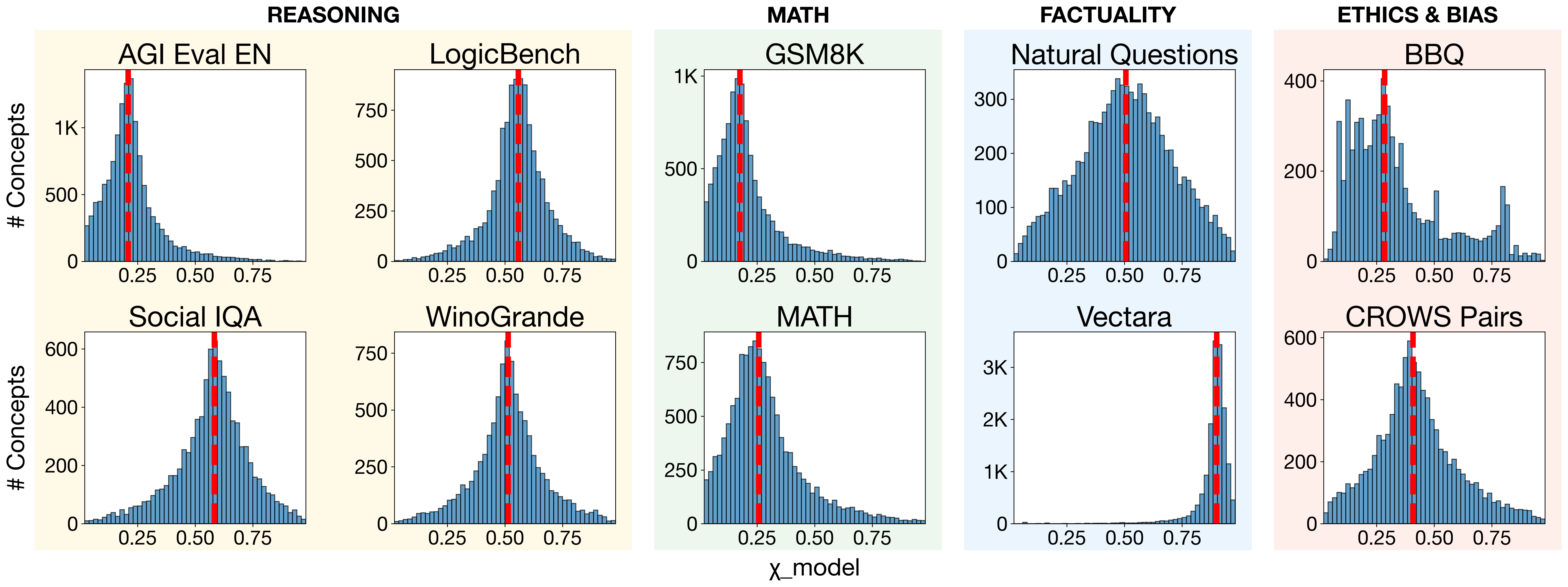} %
    \caption{\textbf{Per-Benchmark Distributions for Model Performance.}  A breakdown of model performance $\chi_{\text{model}}^{(b,c)}$ score distributions for individual benchmarks obtained for Gemma 2 2B. The red line indicates the median.}
    \label{fig:gemma_bench_lvl_performance}
\end{figure*}

\FloatBarrier

{ \subsection{Robustness} }

\begin{table*}[h!]
\centering
\setlength{\tabcolsep}{4pt}

\caption{ { \textbf{Examples of Concept Correspondences in Best/Worst Performance and Coverage Analyses.} The analysis was conducted for Llama 3.1 8B: the Llama SAE results used the model's own activation and custom SAE; the Gemma SAE results used Gemma 2 2B's activations and SAE. They demonstrate reasonable correspondence when the SAEs are swapped.} }
\label{tab:llama_gemma_concepts}

{\small
\begin{tabular}{>{\raggedright\arraybackslash\hyphenpenalty=10000\exhyphenpenalty=10000}p{1.6cm} p{5.5cm} p{5.5cm}}
\toprule
\textbf{Analysis} & \textbf{Llama SAE} & \textbf{Gemma SAE} \\
\midrule

{\textit{Best Performance}}
& {\conc{\textbf{(45314)}} Legal reasoning and argumentation patterns in multiple choice questions}
& {\conc{\textbf{(5471)}} References to legal cases and procedural aspects of law} \\
& {\conc{\textbf{(27510)}} Code patterns for including JavaScript resources in web pages}
& {\conc{\textbf{(8196)}} Code constructs or reserved keywords in programming languages} \\
\midrule

{\textit{Worst Performance}}
& {\conc{\textbf{(2874)}} Mathematical differentiation operators and notation}
& {\conc{\textbf{(13908)}} Numerical values, counts or measurements} \\
& {\conc{\textbf{(2872)}} Explaining time requirements and duration}
& {\conc{\textbf{(9936)}} Dates and numeric sequences} \\
\midrule

{\textit{Best Coverage}}
& {\conc{\textbf{(41290)}} New conversation or topic segment boundary marker}
& {\conc{\textbf{(11527)}} The start of a document} \\
& {\conc{\textbf{(902)}} Step-by-step mathematical explanations and calculations}
& {\conc{\textbf{(11880)}} Mathematical expressions and calculations related to derivatives and factors} \\
\midrule

{\textit{Worst Coverage}}
& {\conc{\textbf{(27900)}} Discussions of factual accuracy and consistency checking}
& {\conc{\textbf{(5657)}} Terms related to correctness and accuracy in responses or answers} \\
& {\conc{\textbf{(47946)}} The assistant explains how it processes and handles information}
& {\conc{\textbf{(1797)}} Phrases related to instructions or operational processes} \\
\bottomrule
\end{tabular}
} 
\end{table*}

{ \paragraph{Perturbation Consistency.} We re-ran the full analysis 100 times, each time randomly dropping 20\% of the examples per benchmark. The resulting standard deviations were: 0.012 for $\bm{\mathit{X}}_{\text{model}}$ and 0.011 for $\bm{\mathit{X}}_{\text{bench}}$. }

{ \paragraph{Adversarial Perturbations.} Upon removal of the most salient 100 datapoints associated with top 100 best-performing concepts across all benchmarks lowered median $\bm{\mathit{X}}_{\text{model}}$ on average by 0.8\%. On the other hand, removing the most salient 100 datapoints associated with top 100 worst-performing concepts across all benchmarks increased median $\bm{\mathit{X}}_{\text{model}}$ on average by 0.5\%. This process was repeated 10 times. }

\FloatBarrier


\newpage
~
\newpage
\section{Additional Results: DeepSeek-R1-Distill-Llama-8B}
\label{app:add_results_deepseek}

We apply CG to \texttt{DeepSeek-R1-Distill-Llama-8B}~\citep{guo2025deepseek} using the Llama-Scope-R1-Distill SAE (residual stream, layer 20)~\citep{he2024llamascope}.

\subsection{Benchmark Gaps}

\begin{figure}[h]
    \centering
    \includegraphics[width=0.49\textwidth]{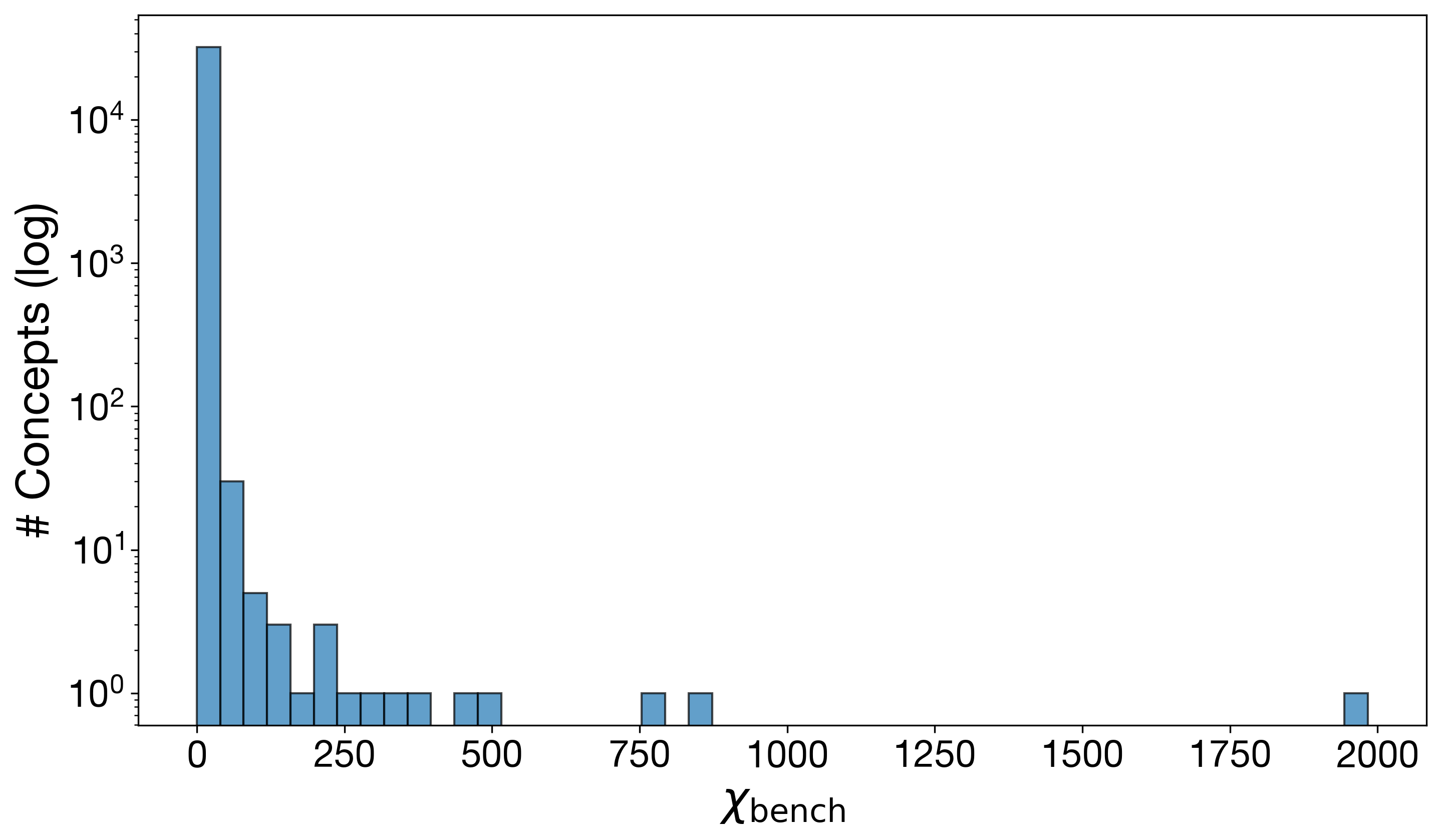}
    \caption{\textbf{Cross-Benchmark Coverage.} The distribution of $\bm{\mathit{X}}_{\text{bench}}^{(c)}$ scores obtained for the $10$ evaluated benchmarks, using the SAE of DeepSeek-R1-Distill-Llama-8B.}
    \label{fig:deepseek_overall_coverage}
\end{figure}

\begin{figure*}[h]
    \centering
    \includegraphics[width=\linewidth]{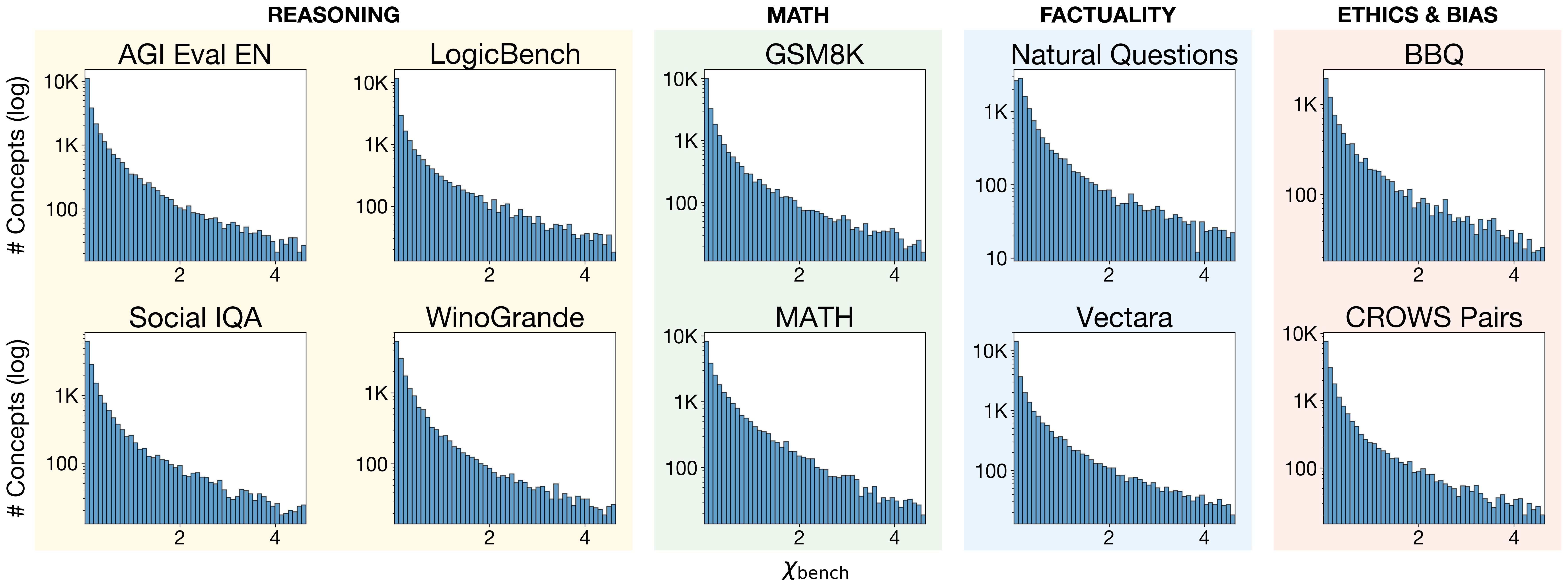}
    \caption{\textbf{Coverage Within Individual Benchmarks.} A breakdown of $\chi_{\text{bench}}^{(b,c)}$ score distributions for individual benchmarks obtained via DeepSeek-R1-Distill-Llama-8B.}
    \label{fig:deepseek_bench_lvl_coverage}
\end{figure*}

\begin{figure}[h]
    \centering
    \includegraphics[width=0.49\textwidth]{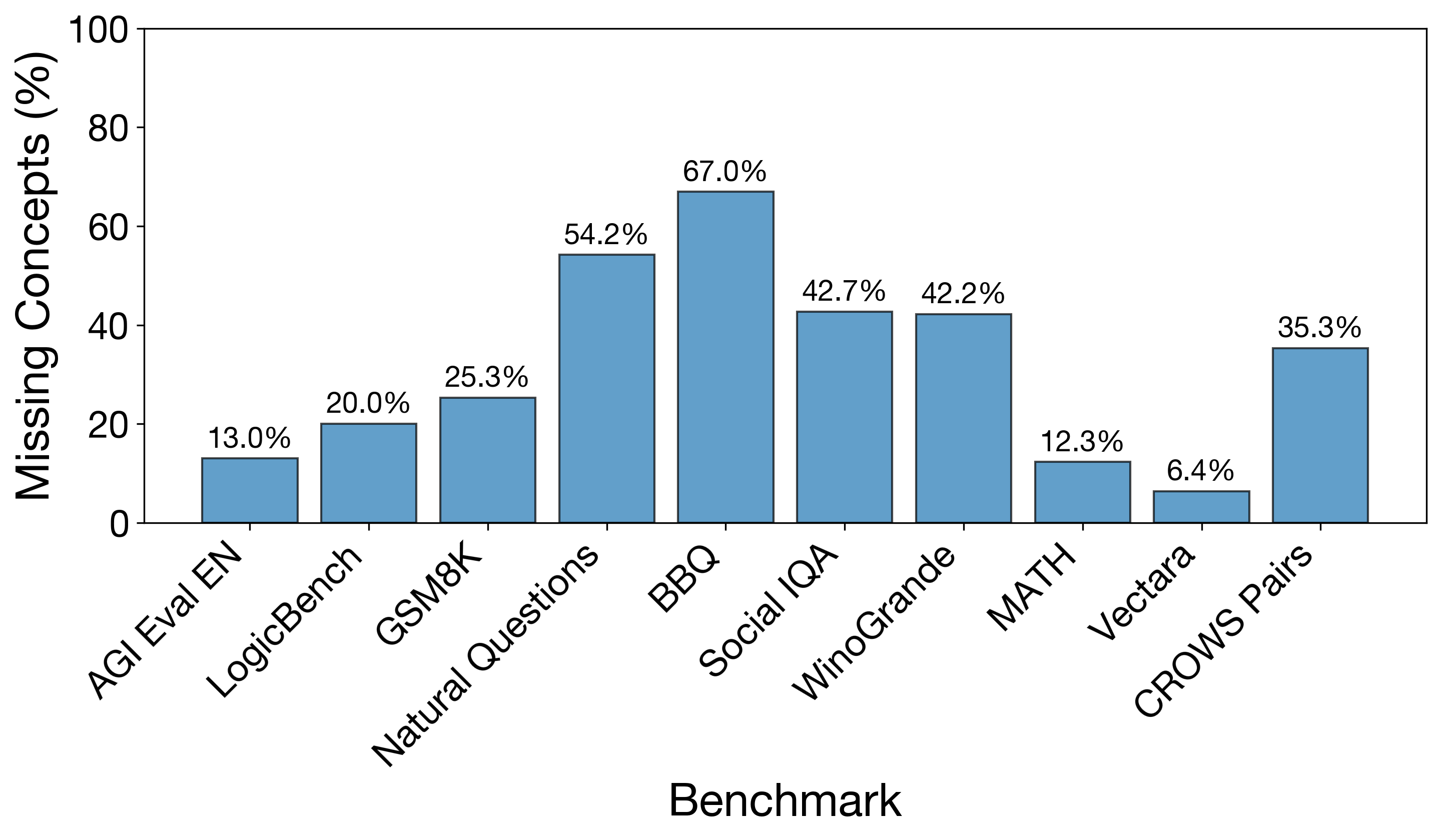}
    \caption{\textbf{Missing Concepts.} Proportion of the SAE concept dictionary for DeepSeek-R1-Distill-Llama-8B that is not tested by the respective benchmarks.}
    \label{fig:deepseek_missing_concept_ratio}
\end{figure}

\begin{table}[h]
    \centering
    \caption{\textbf{Examples of Missing Concepts from the Full Benchmark Suite (DeepSeek-R1-Distill-Llama-8B).}}
    \label{tab:deepseek_missing_concepts_full_suite}
    \resizebox{0.8\textwidth}{!}{
    \begin{tabular}{cl}
    \toprule
    \textbf{Concept ID} & \textbf{Concept Description} \\
    \toprule
    \addlinespace[0.6ex]
    \conc{\textbf{(4492)}} & mathematical reasoning and trigonometry \\
    \cmidrule{1-2}
    \conc{\textbf{(5054)}} & text that contains mathematical reasoning including variables, angles, equations, and calculations \\
    \cmidrule{1-2}
    \conc{\textbf{(3936)}} & the word ``opinion'' and its derivatives \\
    \cmidrule{1-2}
    \conc{\textbf{(12843)}} & code snippets relating to requests, tags, and parameters \\
    \cmidrule{1-2}
    \conc{\textbf{(799)}} & words ending in ``ic, ics, ist, ied, al'' or their inflections \\
    \addlinespace[0.6ex]
    \bottomrule
    \end{tabular}
    }
\end{table}

\begin{table}[h]
\centering
\caption{\textbf{Examples of Missing Concepts from Individual Benchmarks (DeepSeek-R1-Distill-Llama-8B).}}
\label{tab:deepseek_missing_concepts}
\resizebox{0.93\textwidth}{!}{
\begin{tabular}{lcl}
\toprule
\textbf{Benchmark} & \textbf{Concept ID} & \textbf{Concept Description} \\
\toprule
\addlinespace[0.6ex]
\textit{AGI Eval} & \conc{\textbf{(699)}} & mentions of museums or libraries, or potentially things you might find in those places \\
\cmidrule{2-3}
& \conc{\textbf{(30886)}} & information about Arabic media sources \\
\addlinespace[0.6ex]
\toprule
\textit{LogicBench} & \conc{\textbf{(24511)}} & words related to math problems and solutions \\
\cmidrule{2-3}
& \conc{\textbf{(8523)}} & the word ``multiple'' or ``multiples'' appearing in mathematical contexts \\
\addlinespace[0.6ex]
\toprule
\textit{Social IQA} & \conc{\textbf{(8971)}} & code comments and code fragments \\
\cmidrule{2-3}
& \conc{\textbf{(17303)}} & the words ``root'' or ``roots'', as well as legal terms \\
\bottomrule
\end{tabular}
}
\end{table}

\FloatBarrier

\subsection{Model Gaps}

\begin{figure}[h]
    \centering
    \includegraphics[width=0.49\textwidth]{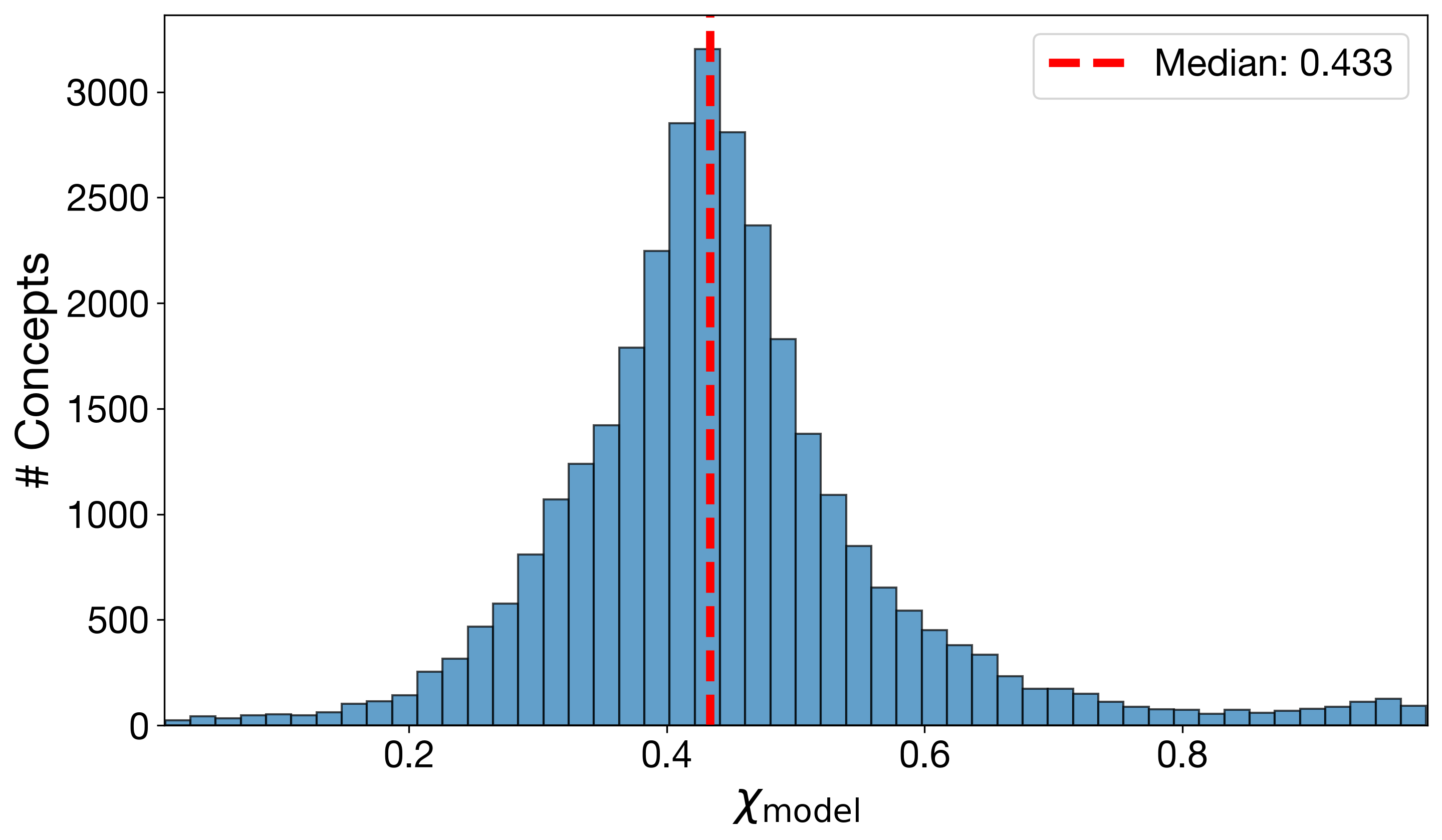}
    \caption{\textbf{Cross-Benchmark Performance.} The distribution of $\bm{\mathit{X}}_{\text{model}}^{(c)}$ scores obtained for DeepSeek-R1-Distill-Llama-8B.}
    \label{fig:deepseek_overall_performance}
\end{figure}

\begin{figure*}[h]
    \centering
    \includegraphics[width=\linewidth]{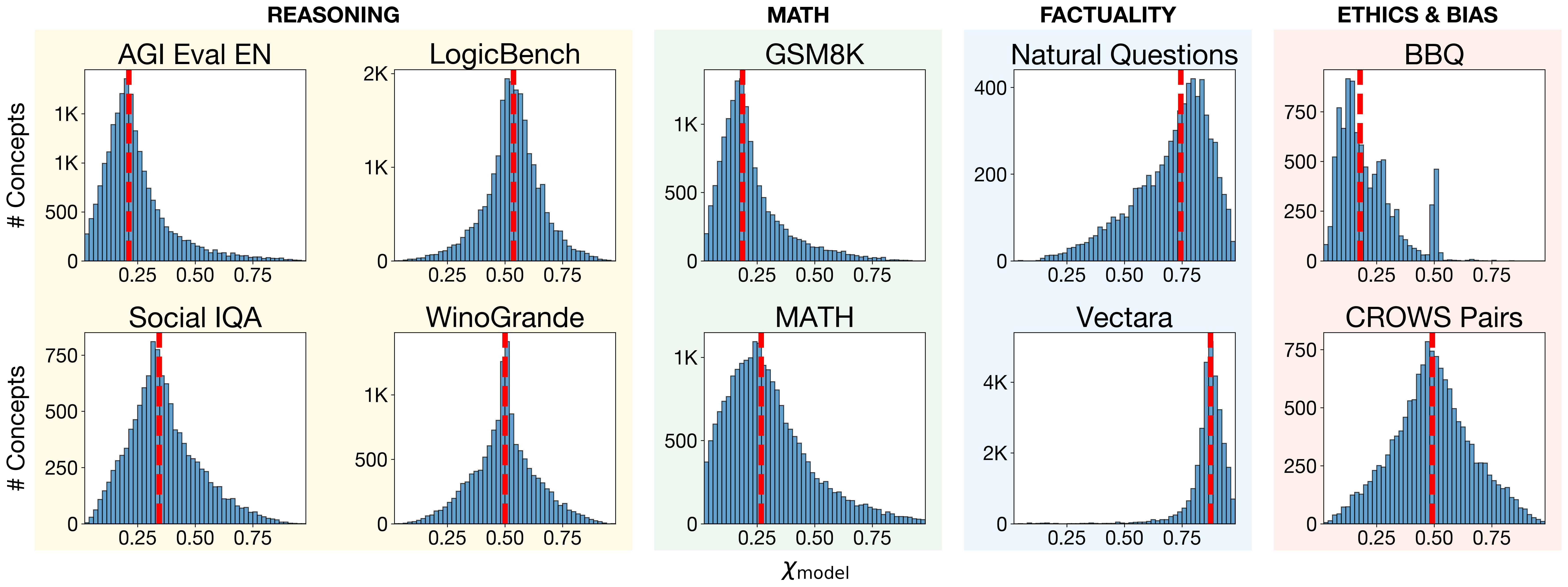}
    \caption{\textbf{Per-Benchmark Distributions for Model Performance.} A breakdown of model performance $\chi_{\text{model}}^{(b,c)}$ score distributions for individual benchmarks obtained for DeepSeek-R1-Distill-Llama-8B. The red line indicates the median.}
    \label{fig:deepseek_bench_lvl_performance}
\end{figure*}

\begin{table}[h]
\centering
\caption{\textbf{Examples of Best- and Worst-Performing Concepts (DeepSeek-R1-Distill-Llama-8B).} Filtered to concepts that activate in at least 5 benchmarks with $\ge 50$ supporting datapoints. The pattern is suggestive: R1-Distill scores well on the surface forms of math reasoning (variables, LaTeX, reasoning-step transitions) but poorly on its substance (multi-digit arithmetic, boxed numerical answers, code syntax).}
\label{tab:deepseek_perf_concepts}
\resizebox{0.93\textwidth}{!}{
\begin{tabular}{lcl}
\toprule
\textbf{} & \textbf{Concept ID} & \textbf{Concept Description} \\
\toprule
\addlinespace[0.6ex]
\textit{Best Performance} & \conc{\textbf{(14585)}} & mentions of variables, specifically within mathematical or logical reasoning \\
\cmidrule{2-3}
& \conc{\textbf{(17004)}} & LaTeX math code \\
\cmidrule{2-3}
& \conc{\textbf{(17718)}} & the word ``Alternatively'' which is used to introduce different reasoning steps to solve math problems \\
\addlinespace[0.6ex]
\toprule
\textit{Worst Performance} & \conc{\textbf{(32637)}} & boxed numerical answers in a mathematical context \\
\cmidrule{2-3}
& \conc{\textbf{(24736)}} & multi-digit numbers \\
\cmidrule{2-3}
& \conc{\textbf{(9878)}} & code blocks in C\# with curly braces \\
\bottomrule
\end{tabular}
}
\end{table}

\FloatBarrier

\newpage
~
\newpage
\section{Additional Results: Mistral-7B-Instruct-v0.1}
\label{app:add_results_mistral}

We apply CG to \texttt{Mistral-7B-Instruct-v0.1}~\citep{jiang2023mistral} using a publicly available SAE attached at layer 16 (MLP output)~\citep{cosgrove2024mistralsae}. The autointerpretability labels for this SAE are short, terse fragments rather than full descriptions, so the example tables below should be read as illustrative of \emph{which} concepts CG surfaces rather than as fully interpretable concept names.

\subsection{Benchmark Gaps}

\begin{figure}[h]
    \centering
    \includegraphics[width=0.49\textwidth]{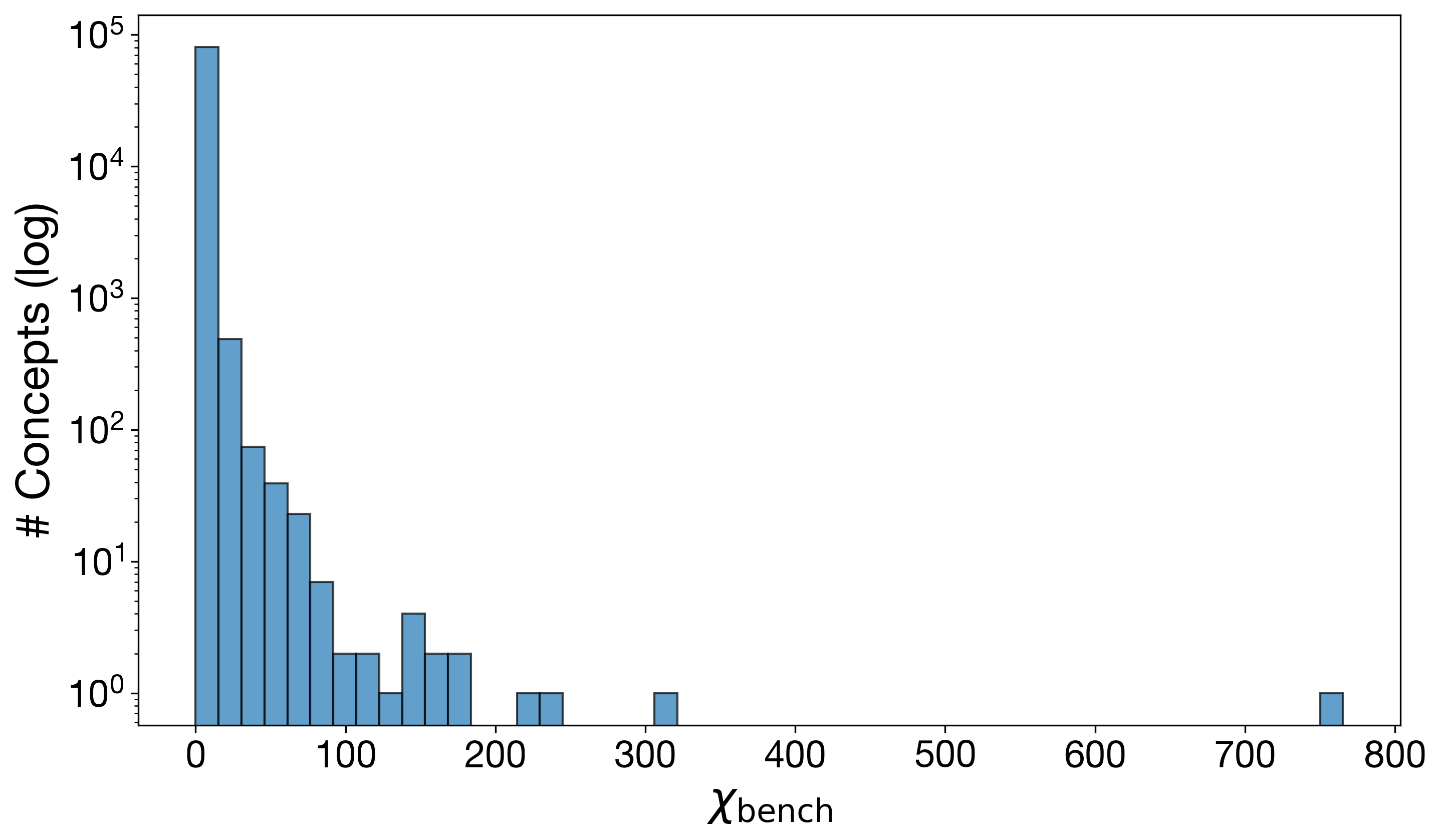}
    \caption{\textbf{Cross-Benchmark Coverage.} The distribution of $\bm{\mathit{X}}_{\text{bench}}^{(c)}$ scores obtained for the $10$ evaluated benchmarks, using the SAE of Mistral-7B-Instruct-v0.1.}
    \label{fig:mistral_overall_coverage}
\end{figure}

\begin{figure*}[h]
    \centering
    \includegraphics[width=\linewidth]{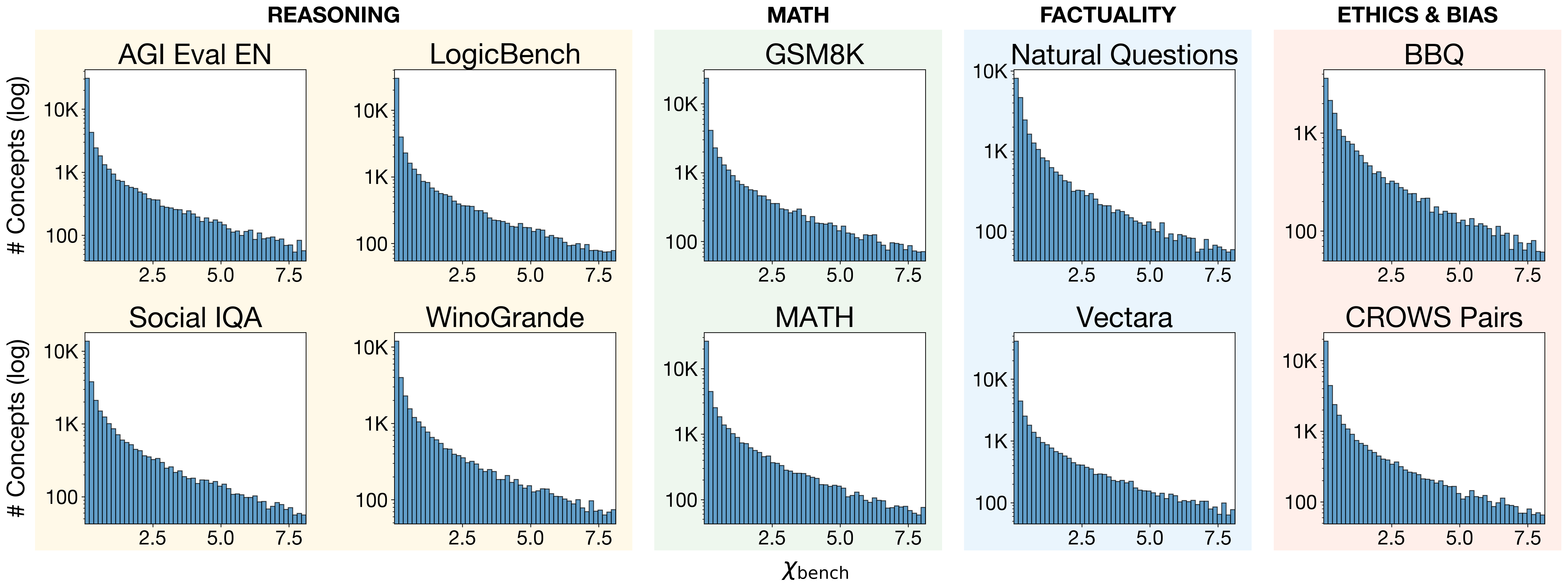}
    \caption{\textbf{Coverage Within Individual Benchmarks.} A breakdown of $\chi_{\text{bench}}^{(b,c)}$ score distributions for individual benchmarks obtained via Mistral-7B-Instruct-v0.1.}
    \label{fig:mistral_bench_lvl_coverage}
\end{figure*}

\begin{figure}[h]
    \centering
    \includegraphics[width=0.49\textwidth]{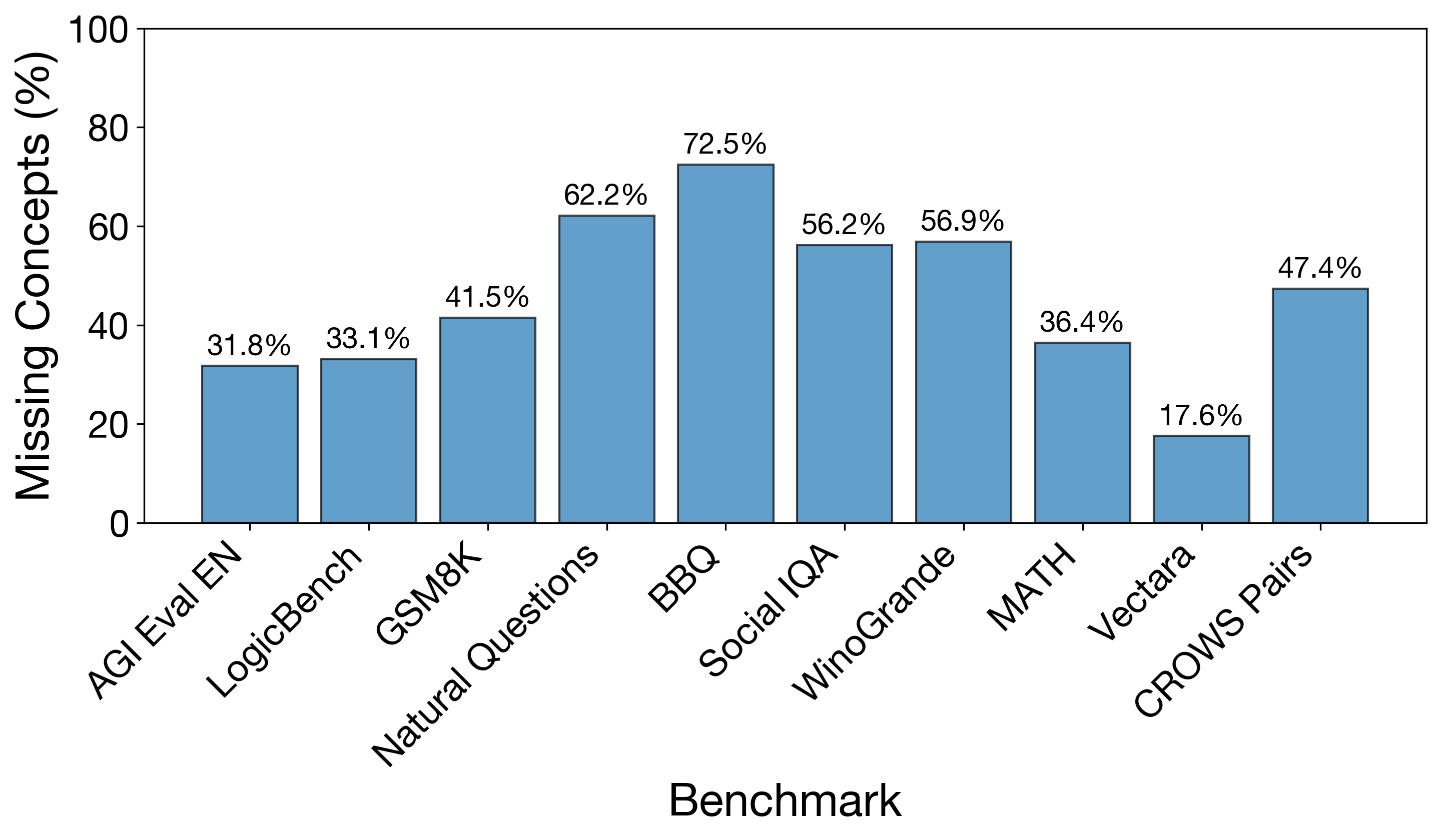}
    \caption{\textbf{Missing Concepts.} Proportion of the SAE concept dictionary for Mistral-7B-Instruct-v0.1 that is not tested by the respective benchmarks.}
    \label{fig:mistral_missing_concept_ratio}
\end{figure}

\FloatBarrier

\subsection{Model Gaps}

\begin{figure}[h]
    \centering
    \includegraphics[width=0.49\textwidth]{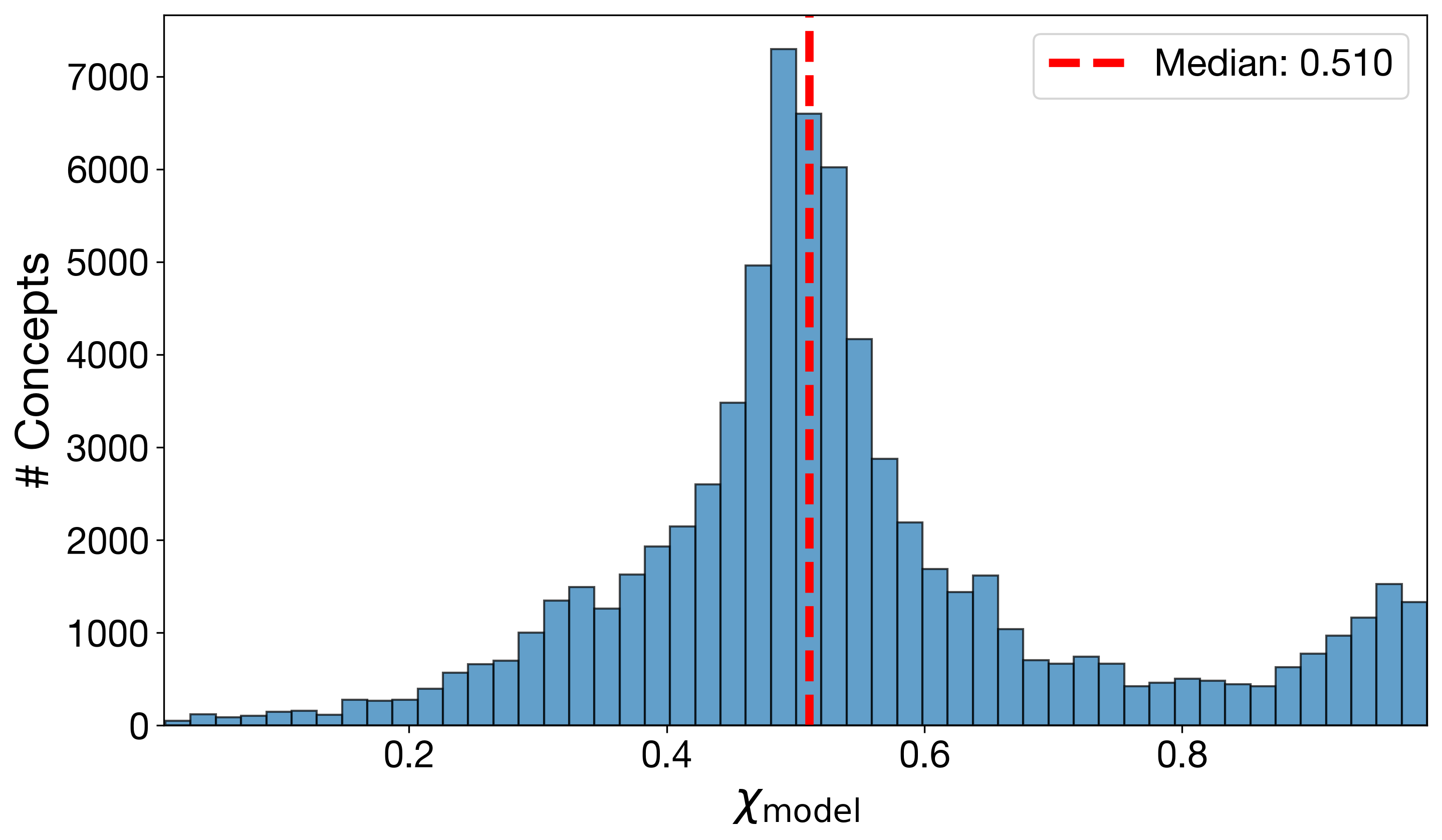}
    \caption{\textbf{Cross-Benchmark Performance.} The distribution of $\bm{\mathit{X}}_{\text{model}}^{(c)}$ scores obtained for Mistral-7B-Instruct-v0.1.}
    \label{fig:mistral_overall_performance}
\end{figure}

\begin{figure*}[h]
    \centering
    \includegraphics[width=\linewidth]{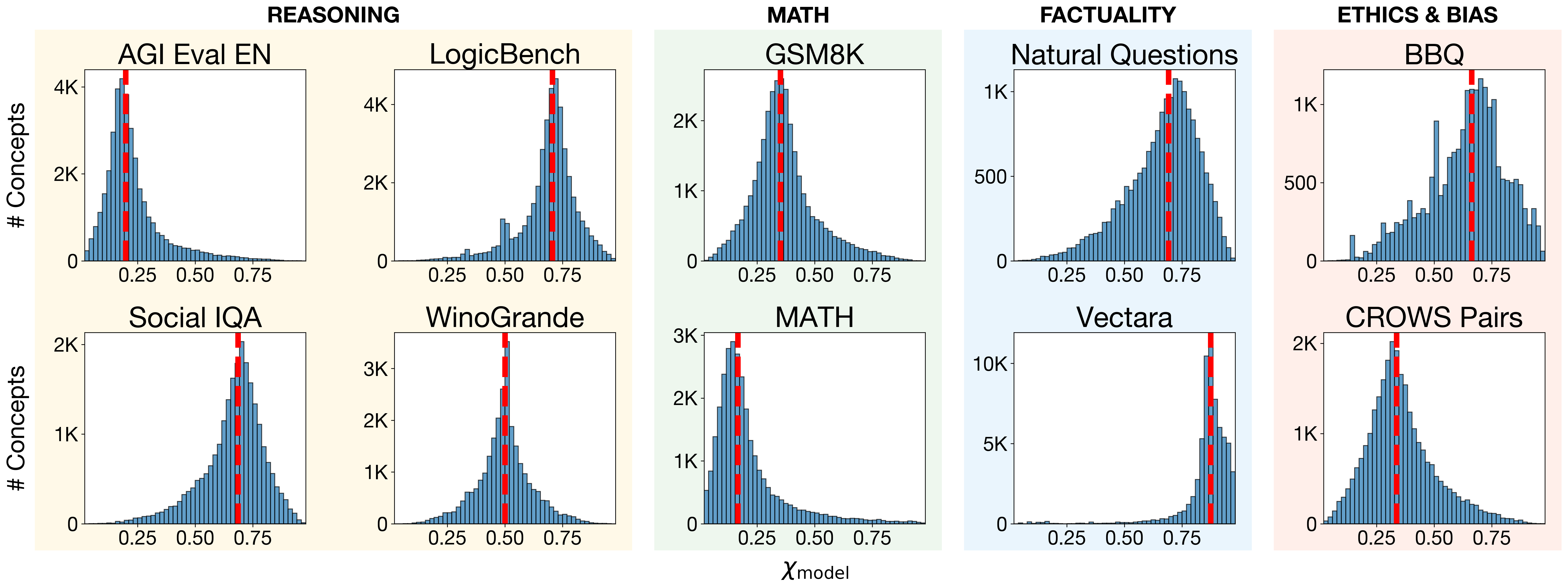}
    \caption{\textbf{Per-Benchmark Distributions for Model Performance.} A breakdown of model performance $\chi_{\text{model}}^{(b,c)}$ score distributions for individual benchmarks obtained for Mistral-7B-Instruct-v0.1. The red line indicates the median.}
    \label{fig:mistral_bench_lvl_performance}
\end{figure*}

\begin{table}[h]
\centering
\caption{\textbf{Examples of Best- and Worst-Performing Concepts (Mistral-7B-Instruct-v0.1).} Filtered to concepts that activate in at least 5 benchmarks with $\ge 50$ supporting datapoints, and further restricted to entries with coherent semantic labels. Coherent labels are sparse for this SAE, so the table is shorter than for the other models.}
\label{tab:mistral_perf_concepts}

\begin{tabular}{lcl}
\toprule
\textbf{} & \textbf{Concept ID} & \textbf{Concept Description} \\
\toprule
\addlinespace[0.6ex]
\textit{Best Performance} & \conc{\textbf{(14155)}} & Months of the year \\
\cmidrule{2-3}
& \conc{\textbf{(101008)}} & automaticity, self-acting \\
\cmidrule{2-3}
& \conc{\textbf{(64910)}} & drawing attributes \\
\addlinespace[0.6ex]
\toprule
\textit{Worst Performance} & \conc{\textbf{(18081)}} & Invention and innovation \\
\cmidrule{2-3}
& \conc{\textbf{(97261)}} & vehicles and assembly \\
\bottomrule
\end{tabular}

\end{table}

\FloatBarrier


\newpage
~
\newpage
\section{Additional Results: Qwen3-4B}
\label{app:add_results_qwen3}

We apply CG to \texttt{Qwen3-4B}~\citep{yang2025qwen3} using a publicly available set of transcoders on its residual stream~\citep{hanna2024qwen3transcoders}. Autointerpretability labels for these transcoders are available for a subset of concepts and tend to be short fragments; the example tables below restrict to that subset and to concepts with broad cross-benchmark support.

\subsection{Benchmark Gaps}

\begin{figure}[h]
    \centering
    \includegraphics[width=0.49\textwidth]{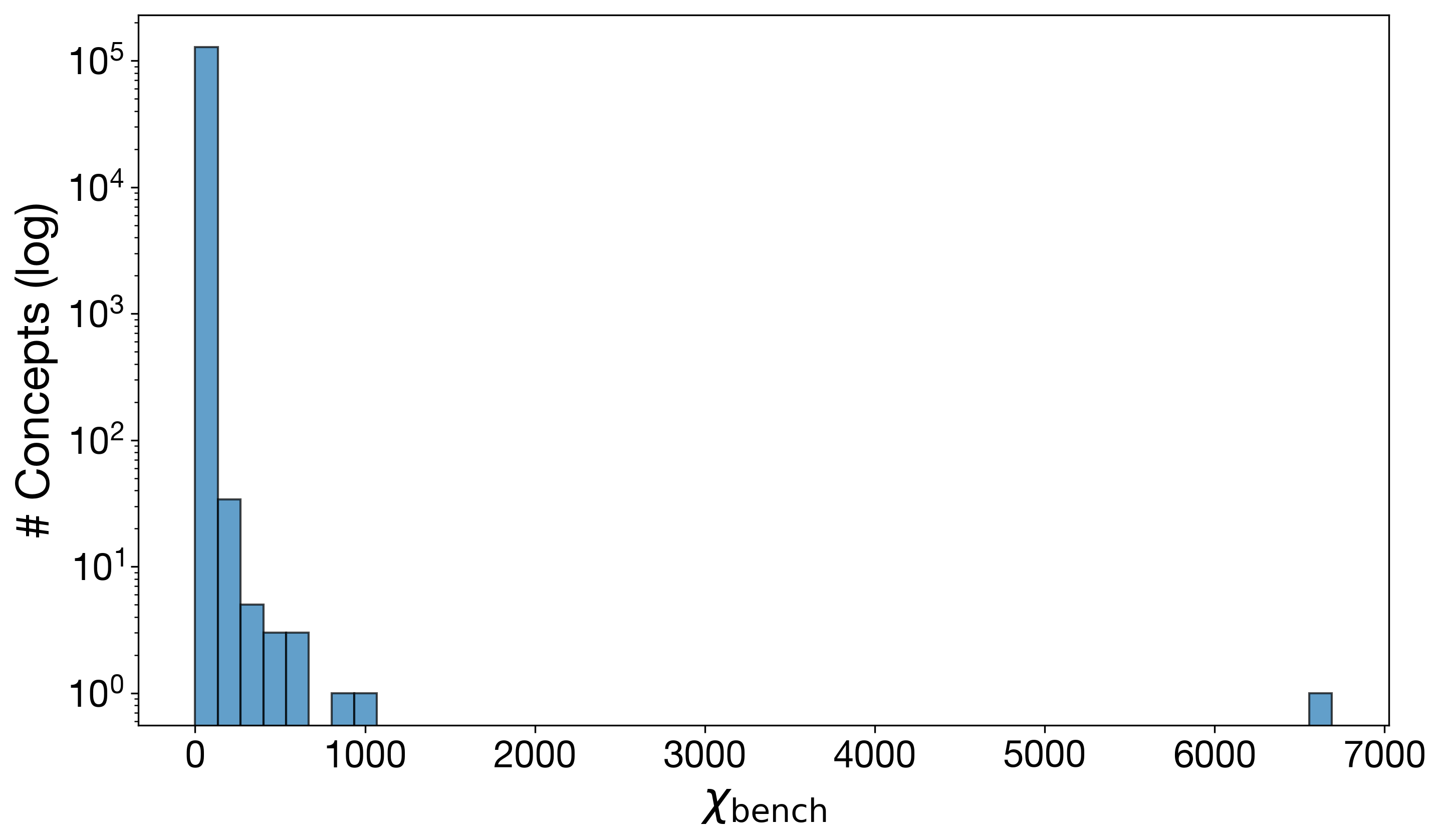}
    \caption{\textbf{Cross-Benchmark Coverage.} The distribution of $\bm{\mathit{X}}_{\text{bench}}^{(c)}$ scores obtained for the $10$ evaluated benchmarks, using the SAE of Qwen3-4B.}
    \label{fig:qwen3_overall_coverage}
\end{figure}

\begin{figure*}[h]
    \centering
    \includegraphics[width=\linewidth]{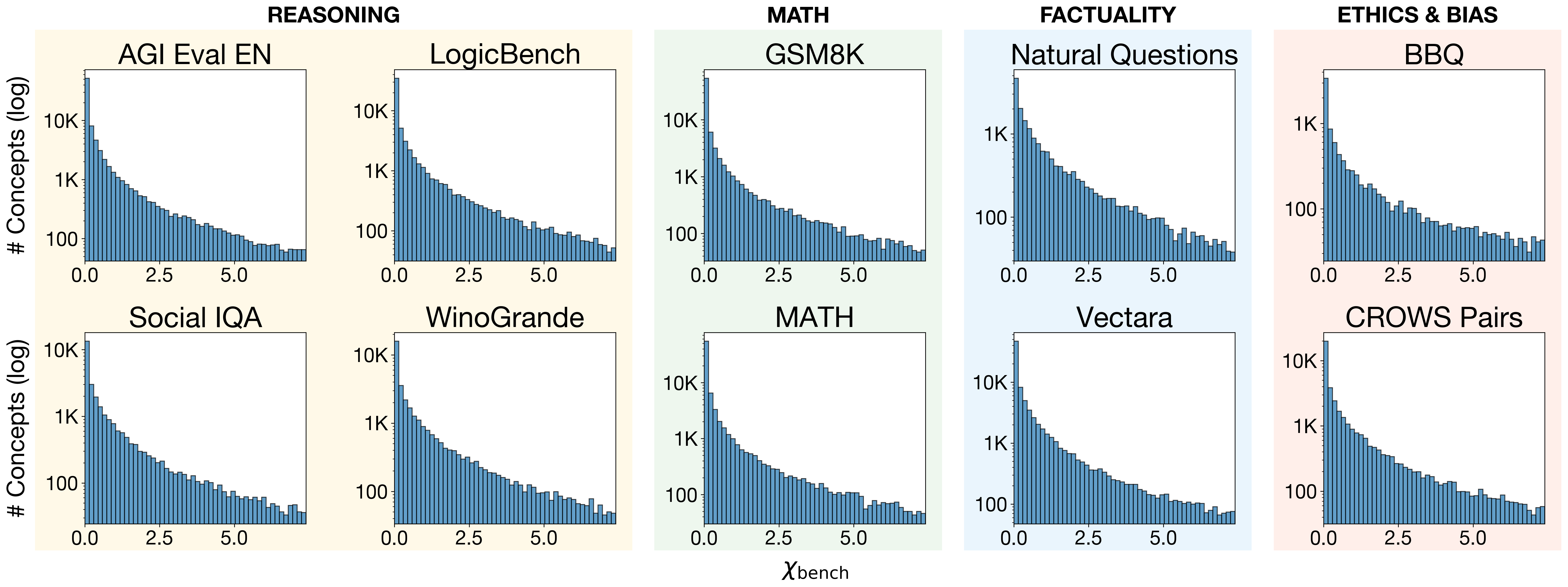}
    \caption{\textbf{Coverage Within Individual Benchmarks.} A breakdown of $\chi_{\text{bench}}^{(b,c)}$ score distributions for individual benchmarks obtained via Qwen3-4B.}
    \label{fig:qwen3_bench_lvl_coverage}
\end{figure*}

\begin{figure}[h]
    \centering
    \includegraphics[width=0.49\textwidth]{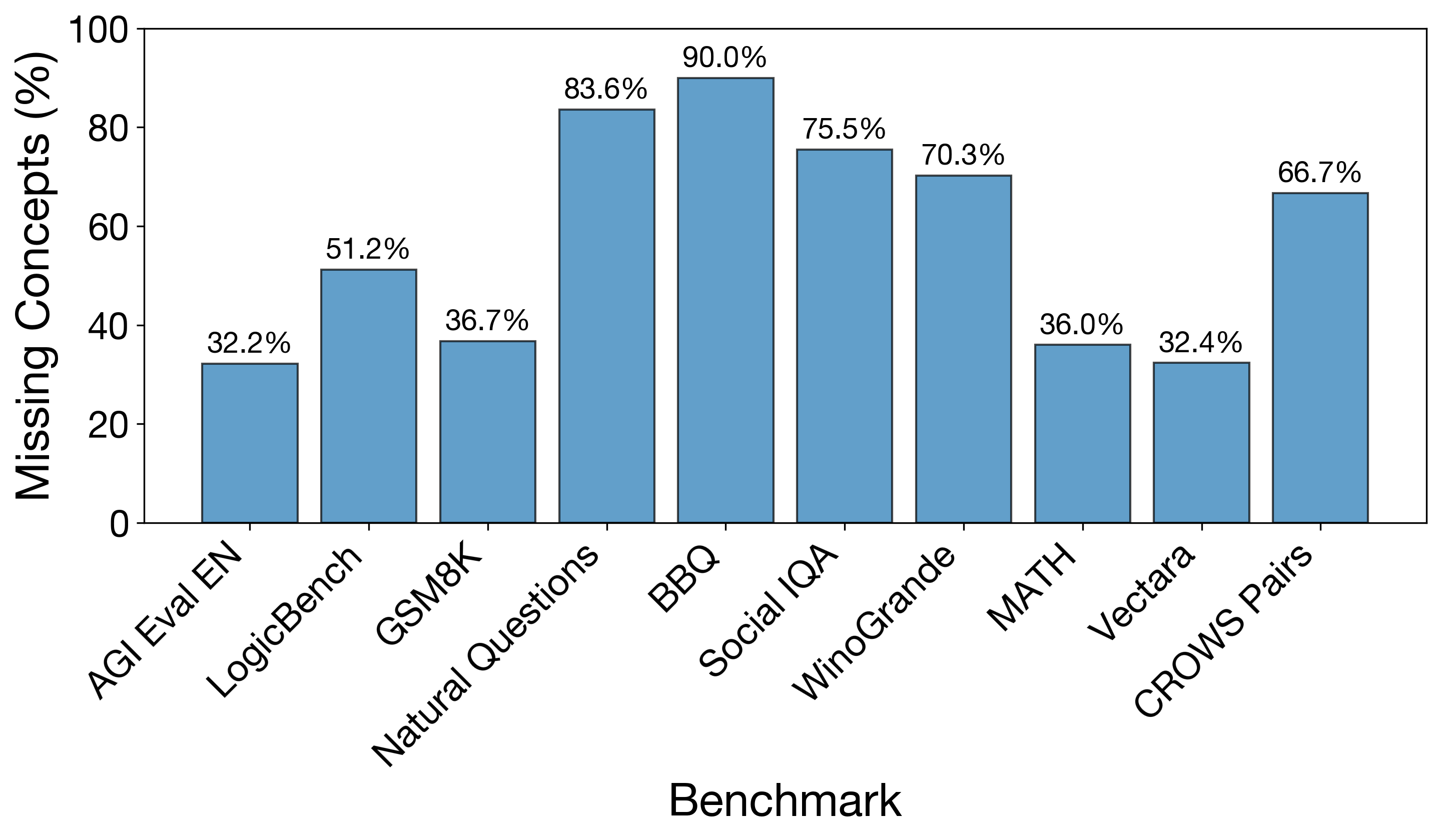}
    \caption{\textbf{Missing Concepts.} Proportion of the SAE concept dictionary for Qwen3-4B that is not tested by the respective benchmarks.}
    \label{fig:qwen3_missing_concept_ratio}
\end{figure}

\FloatBarrier

\subsection{Model Gaps}

\begin{figure}[h]
    \centering
    \includegraphics[width=0.49\textwidth]{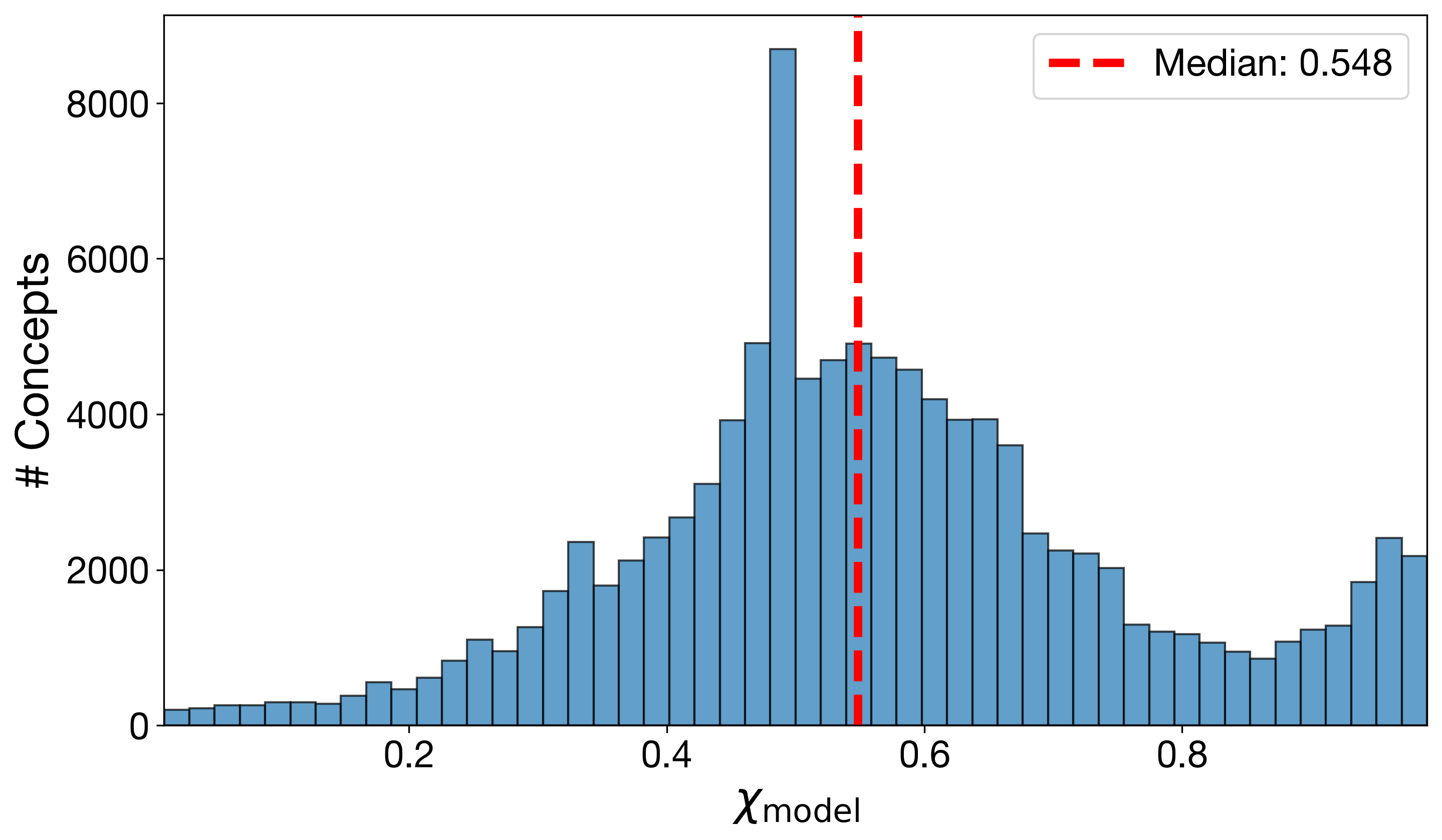}
    \caption{\textbf{Cross-Benchmark Performance.} The distribution of $\bm{\mathit{X}}_{\text{model}}^{(c)}$ scores obtained for Qwen3-4B.}
    \label{fig:qwen3_overall_performance}
\end{figure}

\begin{figure*}[h]
    \centering
    \includegraphics[width=\linewidth]{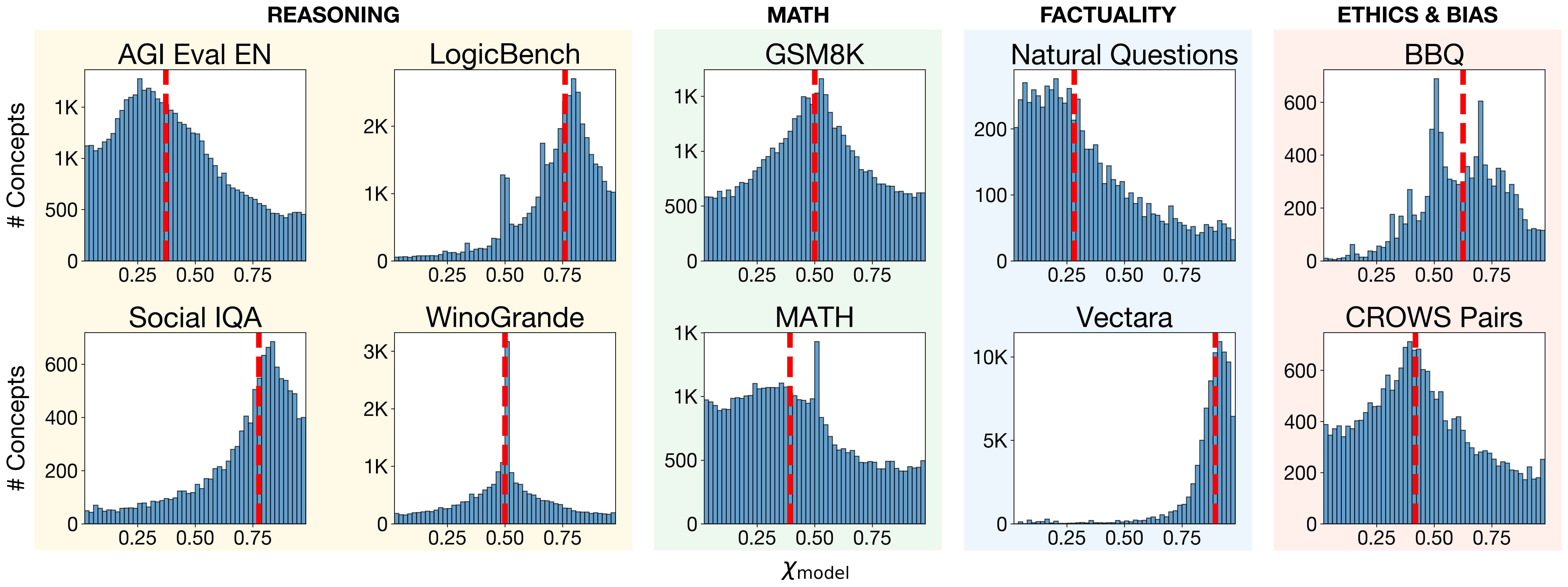}
    \caption{\textbf{Per-Benchmark Distributions for Model Performance.} A breakdown of model performance $\chi_{\text{model}}^{(b,c)}$ score distributions for individual benchmarks obtained for Qwen3-4B. The red line indicates the median.}
    \label{fig:qwen3_bench_lvl_performance}
\end{figure*}

\begin{table}[h]
\centering
\caption{\textbf{Examples of Best- and Worst-Performing Concepts (Qwen3-4B).} Filtered to concepts that activate in at least 5 benchmarks with $\ge 50$ supporting datapoints, and further restricted to entries with coherent semantic labels. Notably, the worst-performing set includes time-related words --- echoing the time-as-a-model-gap finding for Llama 3.1 8B in the main results.}
\label{tab:qwen3_perf_concepts}
\begin{tabular}{lcl}
\toprule
\textbf{} & \textbf{Concept ID} & \textbf{Concept Description} \\
\toprule
\addlinespace[0.6ex]
\textit{Best Performance} & \conc{\textbf{(23176)}} & Hedging \\
\cmidrule{2-3}
& \conc{\textbf{(17871)}} & Positive/innovative descriptions \\
\cmidrule{2-3}
& \conc{\textbf{(22392)}} & Negative deviations from ideal state \\
\addlinespace[0.6ex]
\toprule
\textit{Worst Performance} & \conc{\textbf{(24345)}} & Time-related words \\
\cmidrule{2-3}
& \conc{\textbf{(18102)}} & code snippets \\
\cmidrule{2-3}
& \conc{\textbf{(7346)}} & government and locations \\
\bottomrule
\end{tabular}
\end{table}

\subsection{Causal Validation: Foreign-Name Bias}
\label{app:qwen3_foreign_name_bias}

Beyond surfacing concepts associated with poor performance, CG can also guide targeted experiments to causally validate the identified model gaps. As an illustration, we examined concept \conc{\textbf{(19086)}} ``\texttt{Foreign languages, names}'', which received a low $\bm{\mathit{X}}_{\text{model}}^{(c)}$ score for Qwen3-4B. We constructed a LogicBench-style prompt with two protagonists and ran it $10$ times in each of two settings: (A) protagonists unnamed, and (B) protagonists named ``Omar'' and ``Mahmoud''. The prompts were otherwise identical. Qwen3-4B answered correctly $80\%$ of the time in setting (A) but only $30\%$ of the time in setting (B), providing direct evidence that the foreign-name model gap surfaced by CG corresponds to a real, behaviorally observable weakness.

\FloatBarrier


\newpage
~
\newpage
\section{Additional Results: Llama 3.1 8B Instruct}
\label{app:add_results_llama_3_1_8b_it}

\subsection{Benchmark Gaps}

\begin{figure*}[h]
    \centering    
    \includegraphics[width=\linewidth]{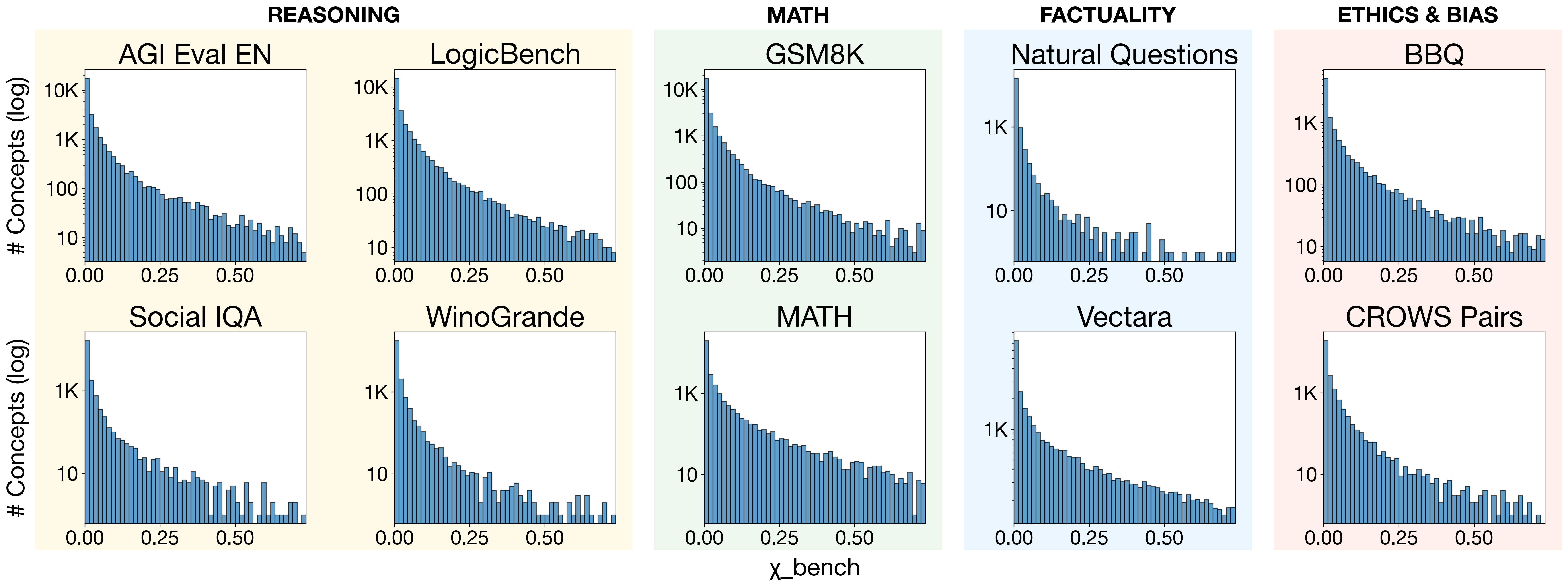}
    \caption{\textbf{Coverage Within Individual Benchmarks.} A breakdown of $\chi_{\text{bench}}^{(b,c)}$ score distributions for individual benchmarks obtained via Llama 3.1 8B. These distributions all show strong right skew, such that average performance on each benchmark is strongly dominated by a small number of concepts with high coverage (high $\chi_{\text{bench}}^{(b,c)}$).}
    \label{fig:bench_lvl_coverage}
\end{figure*}

\begin{figure*}[h]
    \centering    
    \includegraphics[width=0.49\textwidth]{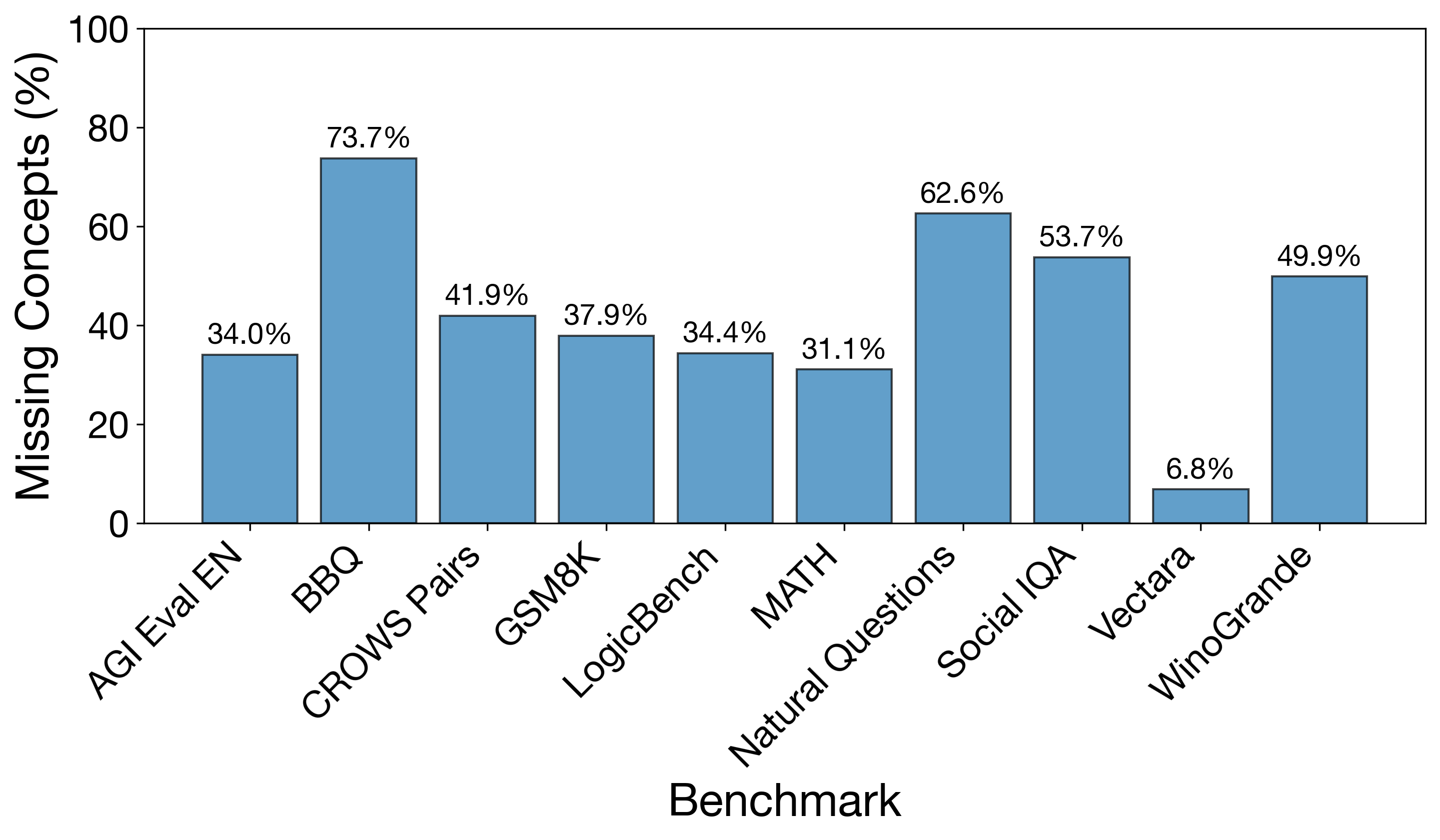}
    \caption{\textbf{Proportion Missing Concepts, for Individual Benchmarks.} Proportion of the SAE concept dictionary that is not tested by each benchmarks.}
    \label{fig:missing_concept_ratio}
\end{figure*}

\begin{figure*}[h]
    \centering    
    \includegraphics[width=0.85\textwidth]{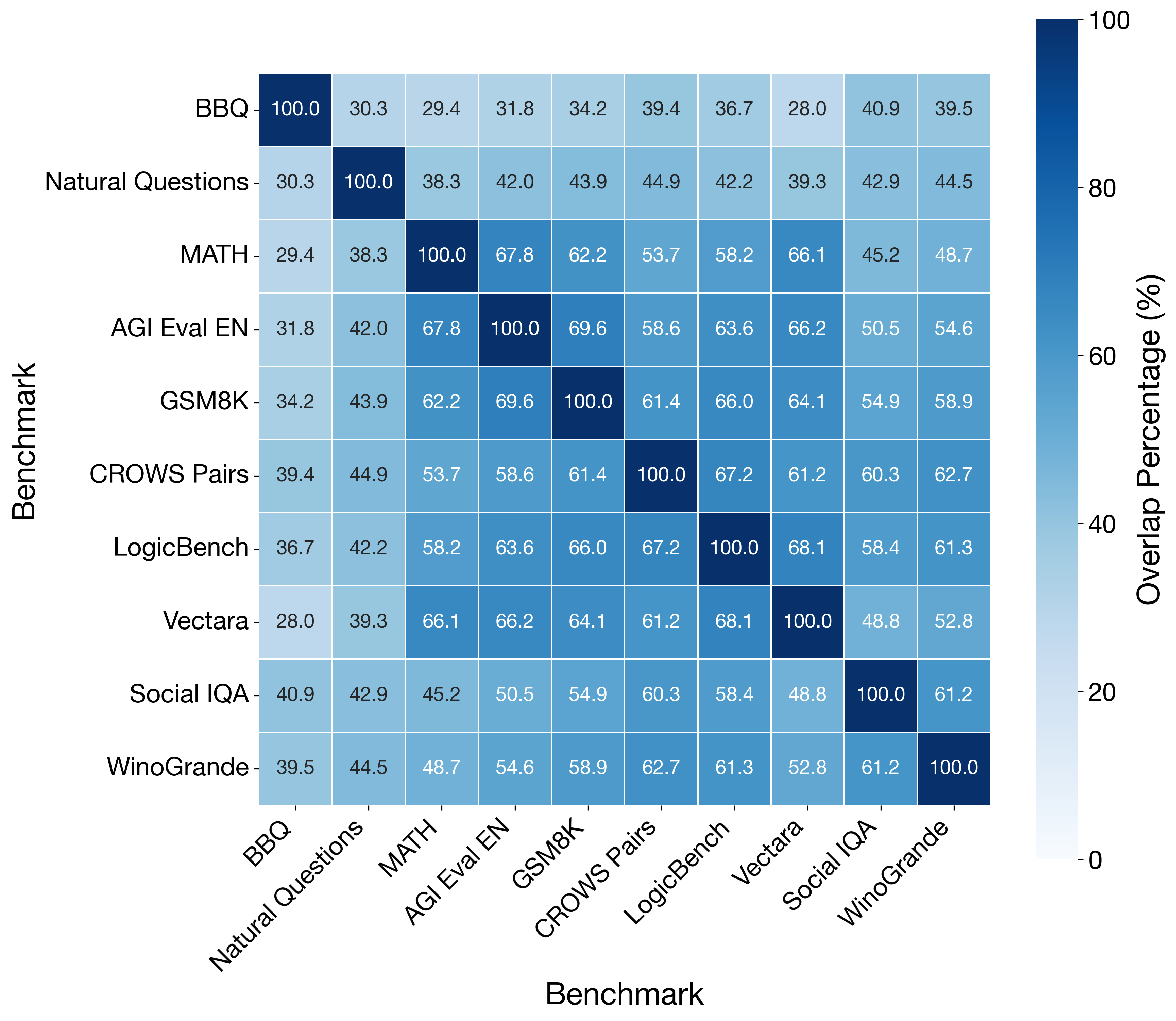}
    \caption{{\textbf{Cross-Benchmark Concept Overlap.} Jaccard similarity of $\bm{\mathit{X}}_{\text{bench}}^{(c)}$ coverage profiles between benchmark pairs, obtained through Llama 3.1 8B, showing which benchmarks share similar concept coverage.}}
    \label{fig:llama_bench_overlap}
\end{figure*}

\newpage
~
\newpage
\subsection{Model Gaps}

\begin{figure}[h]
\vspace{-5pt}
\centering
\begin{subfigure}{0.49\columnwidth}
  \centering
  \includegraphics[width=0.9\linewidth]{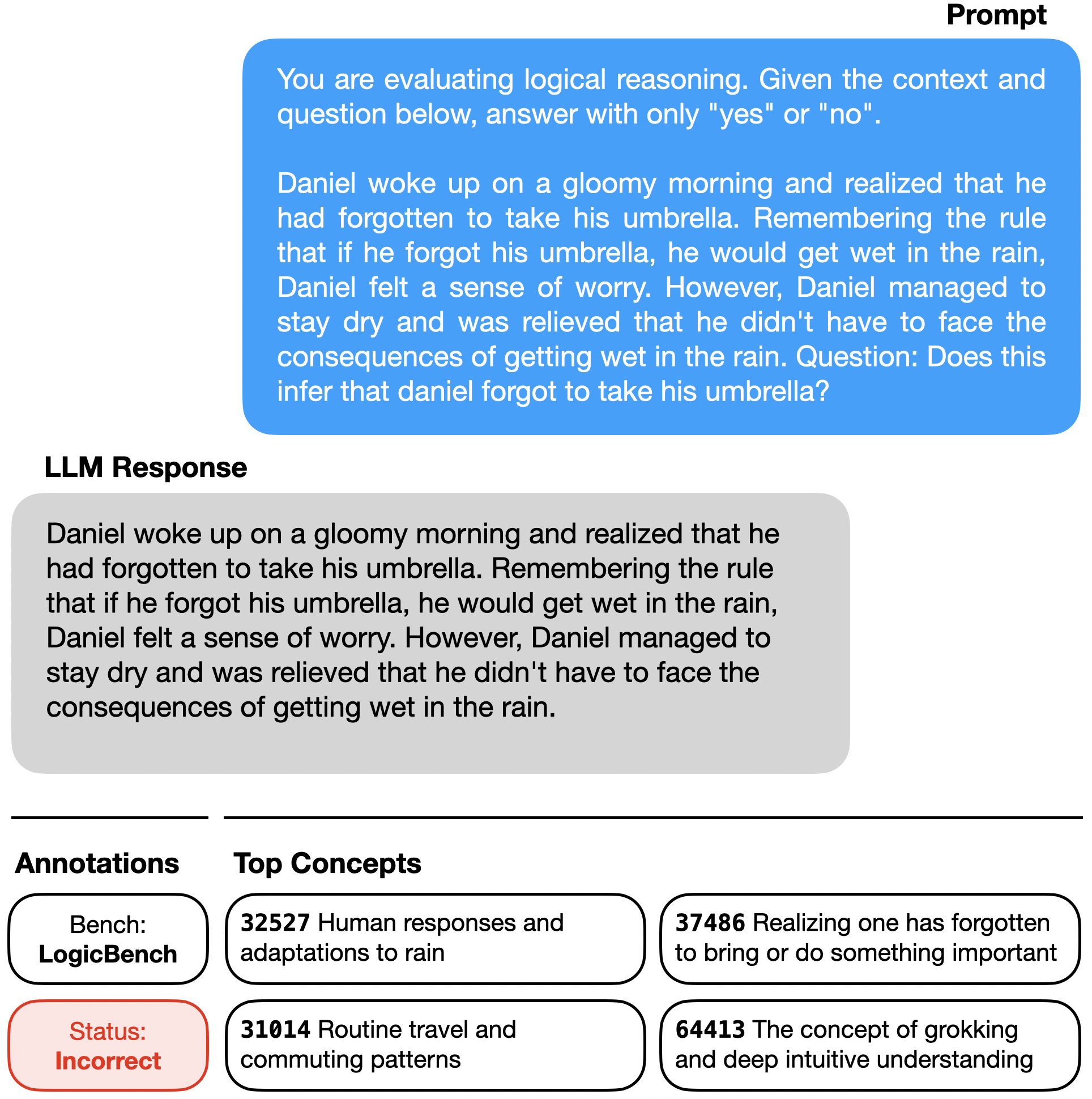}
  \caption{\textbf{LogicBench.}
  Example item associated with an ``intuitive understanding'' concept (64113). Llama 3.1 8B answered incorrectly, consistent with this concept being a model gap.}
  \label{fig:model_gap_logicbench}
\end{subfigure}
\hfill
\begin{subfigure}{0.49\columnwidth}
  \centering
  \includegraphics[width=0.9\linewidth]{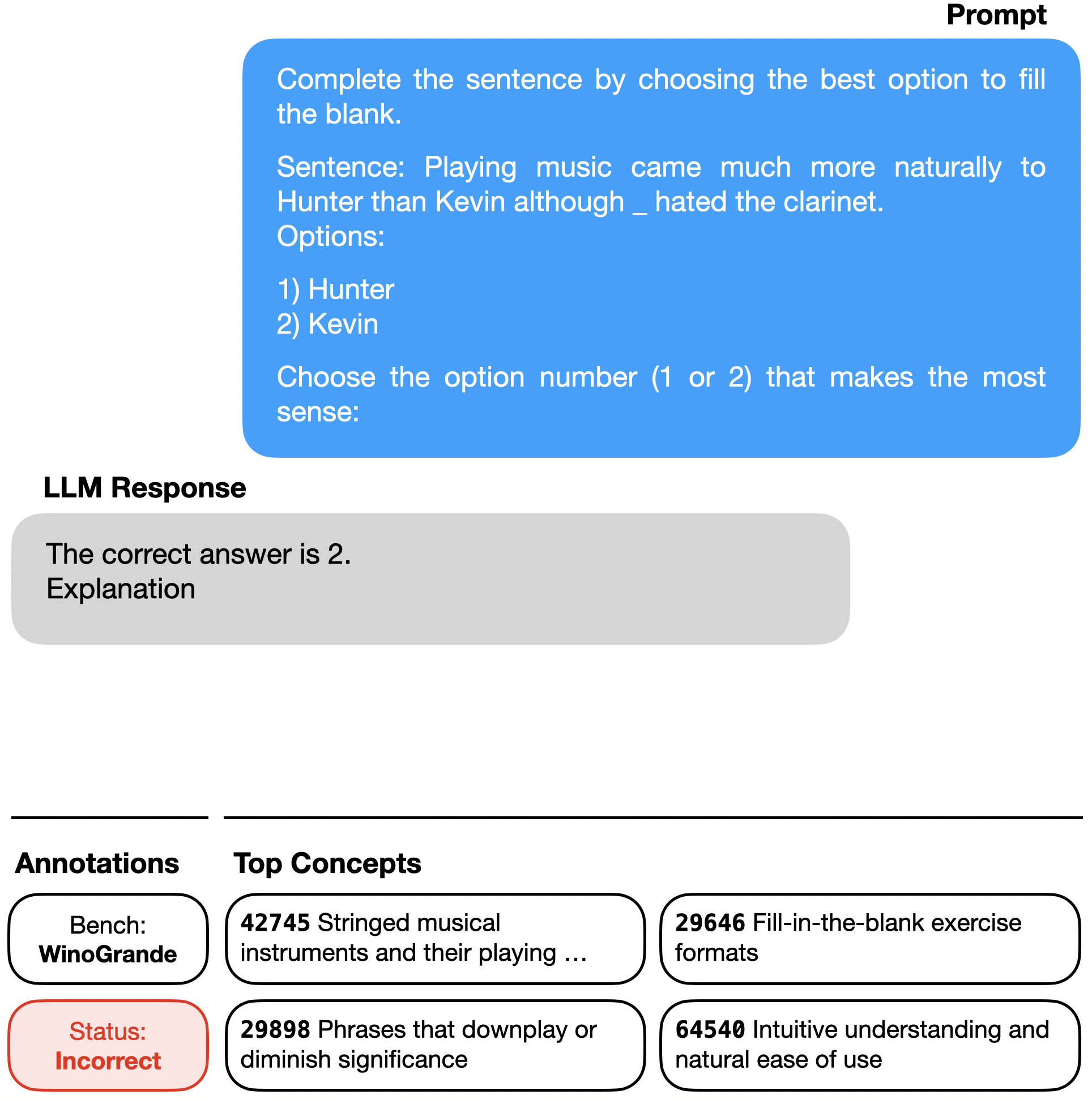}
  \caption{\textbf{WinoGrande.}
  Example item associated with an ``intuitive understanding'' concept (64540). Llama 3.1 8B answered incorrectly, consistent with this concept being a model gap.}
  \label{fig:model_gap_winogrande}
\end{subfigure}
\vspace{-2pt}
\caption{\textbf{Model Gaps Illustrated on Intuitive Understanding Datapoints.}
Two examples from different benchmarks where Llama 3.1 8B fails on items associated with intuitive understanding concepts.}
\label{fig:model_gap_examples}
\vspace{-3pt}
\end{figure}

\begin{figure*}[h]
    \vspace{-3pt}
    \centering
    \includegraphics[width=\linewidth]{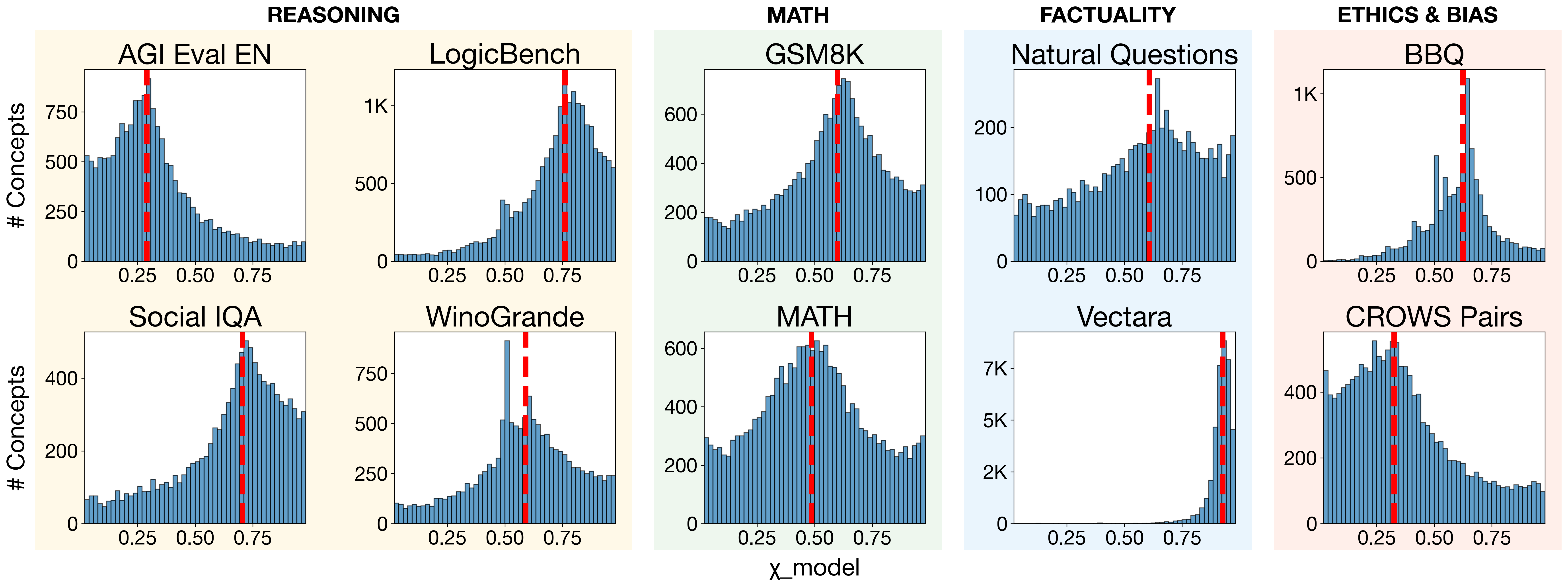}
    \vspace{-15pt}
    \caption{\textbf{Per-Benchmark Distributions for Model Performance.}  A breakdown of model performance $\chi_{\text{model}}^{(b,c)}$ score distributions for individual benchmarks obtained for Llama 3.1 8B. The red line indicates the median.}
    \label{fig:bench_lvl_performance}
\end{figure*}

\FloatBarrier


\newpage
\section{{Additional Results: LMSYS Chatbot Arena}}
\label{app:add_results_lmsys}

{To illustrate how our method can be applied to arena-style benchmarks, we apply CG on Llama 3.1 8B with LMSYS Chatbot Arena~\cite{zheng2023judging}.}

\subsection{{Implementation Details}}

{Unlike the rest of the benchmark datasets evaluated in this paper, which are static datapoints with a scoring policy, LMSYS Chatbot Arena~\cite{zheng2023judging} relies on preference annotations from humans presented with responses of two LLMs at a time, competing on the same input. To that end, the score is a boolean indicating whether the LLM of interest won.}

{We use the data from \url{https://huggingface.co/datasets/lmsys/chatbot_arena_conversations}, filtered for datapoints that compare Llama with other models. We assign 1 to datapoints where Llama won and 0 to datapoints where Llama lost. We extract the SAE concept activations from \emph{both} the input prompts and the model responses to ensure that the computed concept profile reflects the complete semantic footprint, including concepts introduced or emphasized by the model’s generation, which can meaningfully impact human preference. All other aspects of this analysis follow the standard methodology outlined in Section~\ref{sec:method}.}

\subsection{{Benchmark Gaps}}

\begin{figure}[h]
    \centering
    \includegraphics[width=0.49\textwidth]{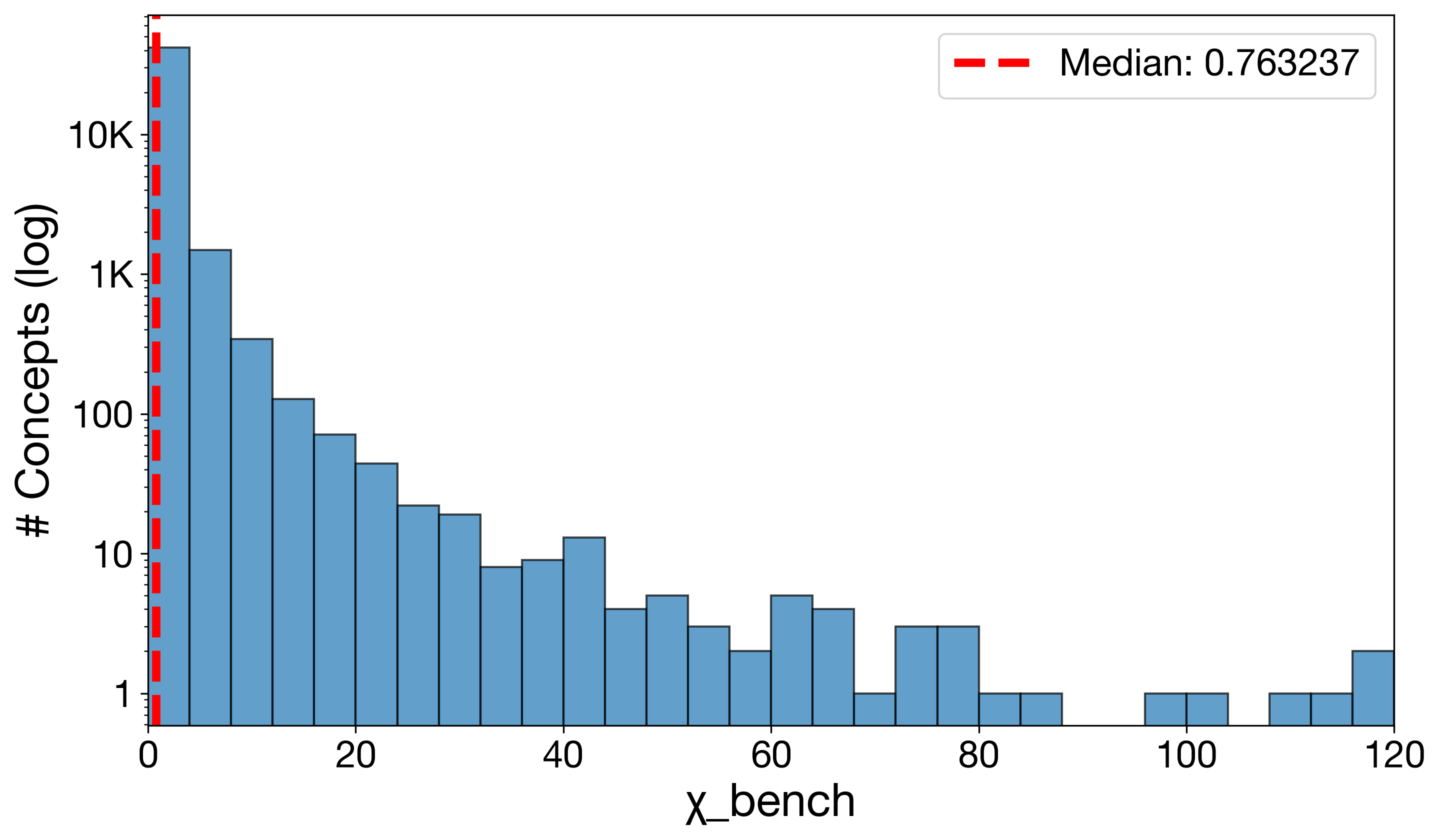} %
    \caption{{\textbf{Benchmark Coverage.} The distribution of $\chi_{\text{bench}}^{(b,c)}$ scores obtained for LMSYS Chatbot Arena, using the SAE of Llama 3.1 8B.}}
    \label{fig:lmsys_bench_coverage}
\end{figure}

\begin{table}[h]
\centering
\caption{{\textbf{Examples of Specific Concepts with the Best and Worst Coverage in LMSYS Chatbot Arena, Obtained Through Llama 3.1 8B.}}}
\label{tab:lmsys_coverage_examples}
\begin{tabular}{>{\raggedright\arraybackslash}p{0.15\linewidth} 
                p{0.12\linewidth} 
                p{0.62\linewidth}}
\toprule
\textbf{Benchmark} & \textbf{Concept ID} & \textbf{Concept Description} \\
\midrule
\addlinespace[0.6ex]
{\textit{Best Coverage}} 
    & \conc{{\textbf{(902)}}} 
    & {Step-by-step mathematical explanations and calculations} \\
& \conc{{\textbf{(9287)}}} 
    & {Numbered steps in instruction lists and process descriptions} \\
\addlinespace[0.8ex]
{\textit{Worst Coverage}} 
    & \conc{{\textbf{(27900)}}} 
    & {Discussions of factual accuracy and consistency checking} \\
& \conc{{\textbf{(14146)}}} 
    & {The assistant should reject inappropriate or NSFW requests} \\
\bottomrule
\end{tabular}
\end{table}

\newpage

\subsection{{Model Gaps}}

\begin{figure}[h]
    \centering
    \includegraphics[width=0.49\textwidth]{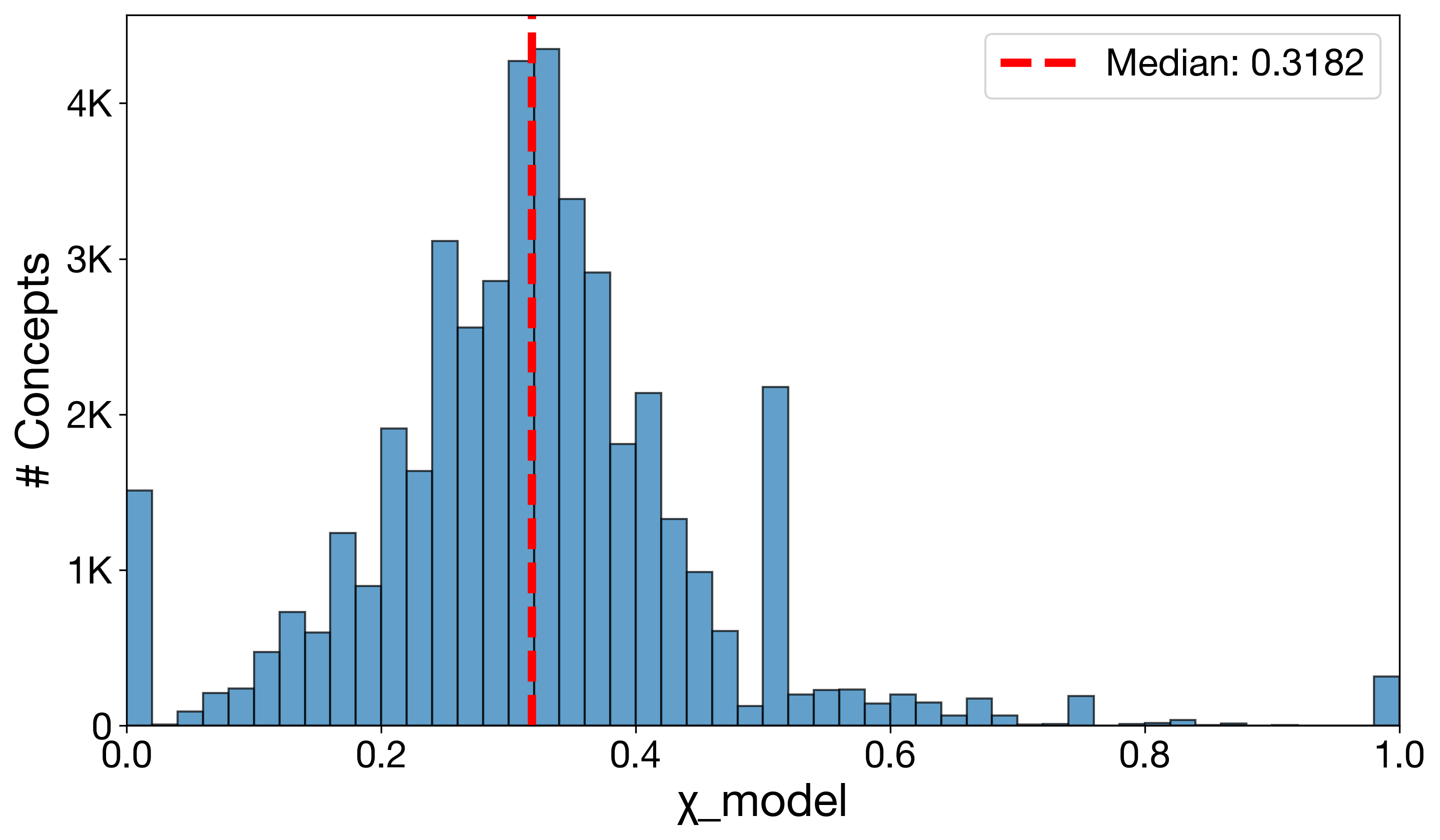} %
    \caption{{\textbf{Benchmark Performance.} The distribution of $\chi_{\text{model}}^{(b,c)}$ scores obtained for LMSYS Chatbot Arena, using the SAE of Llama 3.1 8B.}}
    \label{fig:lmsys_model_coverage}
\end{figure}

\begin{table}[h]
\caption{{\textbf{Examples of Specific Concepts within Llama 3.1 8B with the Best and Worst Performance on LMSYS Chatbot Arena.}}}
\label{tab:lmsys_performance_examples}
\centering
\begin{tabular}{>{\raggedright\arraybackslash}p{0.15\linewidth} 
                p{0.12\linewidth} 
                p{0.62\linewidth}}
\toprule
\textbf{Benchmark} & \textbf{Concept ID} & \textbf{Concept Description} \\
\midrule
\addlinespace[0.6ex]
{\textit{Best Performance}} 
    & \conc{{\textbf{(2691)}}} 
    & {Multiple choice format with options A (okay), B (good), C (wrong) for evaluating behaviors} \\
& \conc{{\textbf{(45314)}}} 
    & {Legal reasoning and argumentation patterns in multiple choice questions} \\
\addlinespace[0.8ex]
{\textit{Worst Performance}} 
    & \conc{{\textbf{(27171)}}} 
    & {The assistant is breaking down complex topics into fundamental concepts} \\
& \conc{{\textbf{(52258)}}} 
    & {Password-related security discussions and requests} \\
\bottomrule
\end{tabular}
\end{table}

\FloatBarrier


\newpage
\section{Additional Results: Conventional Benchmarks}
\label{app:add_results_conventional}

We additionally apply CG to a set of widely used benchmarks that were not part of our main suite: SWE-Bench~\citep{jimenez2023swebench}, Terminal-Bench~\citep{merrill2026terminal}, MMLU~\citep{hendrycks2020measuring}, GPQA~\citep{rein2024gpqa}, and Humanity's Last Exam (HLE)~\citep{phan2025humanity}. All results in this section use \texttt{Qwen3-4B} as the host model, with the same configuration as in Appendix~\ref{app:add_results_qwen3}.


\subsection{SWE-Bench}
\label{app:add_results_conventional_swe_bench}

\begin{figure}[h]
    \centering
    \includegraphics[width=0.49\textwidth]{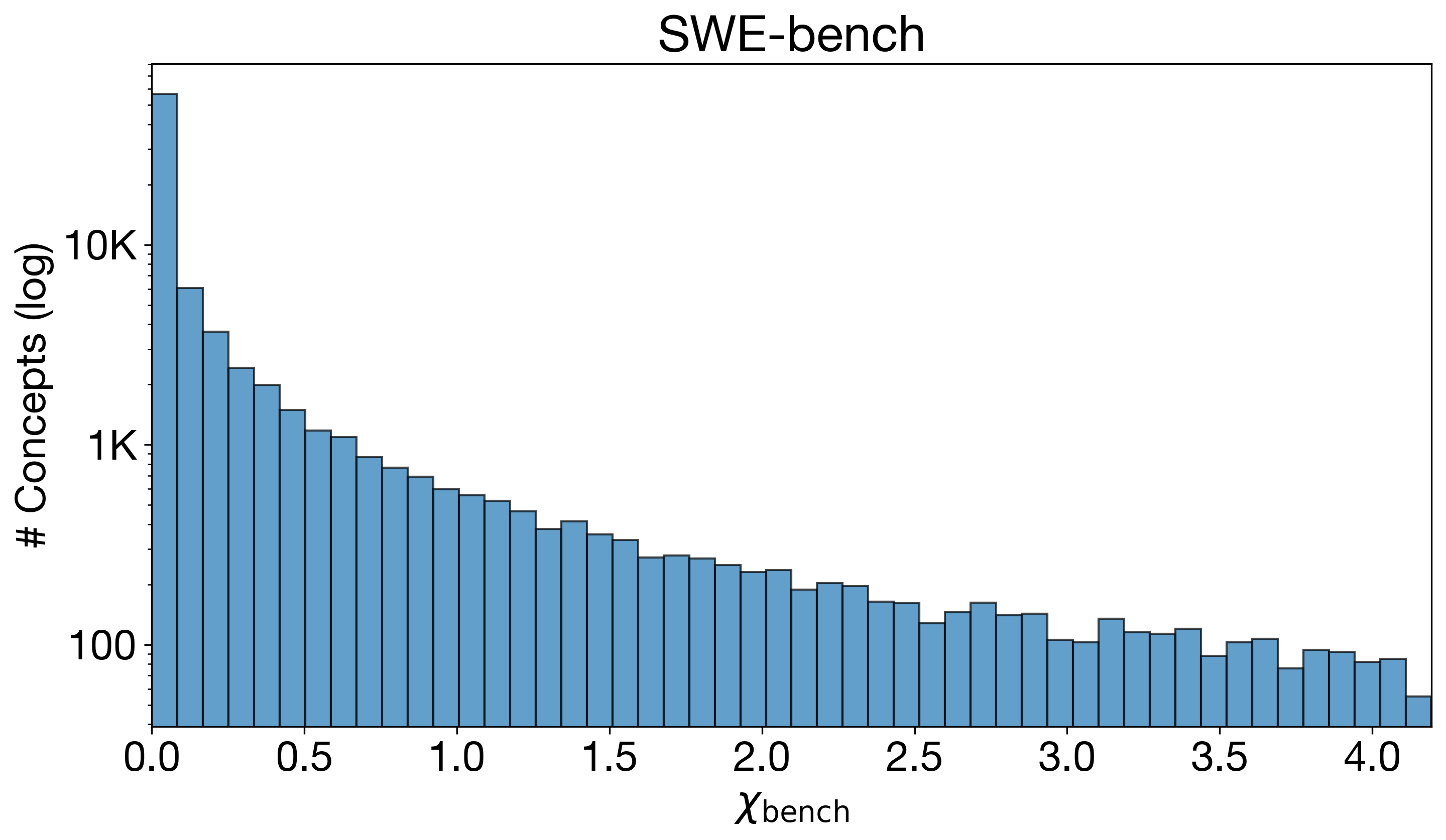}
    \caption{\textbf{Benchmark Coverage.} The distribution of $\chi_{\text{bench}}^{(b,c)}$ scores for SWE-Bench, using the SAE of Qwen3-4B.}
    \label{fig:swe_bench_coverage}
\end{figure}

\begin{table}[h]
\centering
\caption{\textbf{Examples of Best- and Worst-Coverage Concepts in SWE-Bench (Qwen3-4B).}}
\label{tab:swe_bench_coverage_examples}
\begin{tabular}{>{\raggedright\arraybackslash}p{0.15\linewidth}
                p{0.12\linewidth}
                p{0.62\linewidth}}
\toprule
\textbf{} & \textbf{Concept ID} & \textbf{Concept Description} \\
\midrule
\addlinespace[0.6ex]
\textit{Best Coverage}
    & \conc{\textbf{(17172)}}
    & coding errors \\
& \conc{\textbf{(9452)}}
    & bug \\
\addlinespace[0.8ex]
\textit{Worst Coverage}
    & \conc{\textbf{(49022)}}
    & code and licensing \\
& \conc{\textbf{(54116)}}
    & Random characters and code \\
\bottomrule
\end{tabular}
\end{table}

\begin{figure}[h]
    \centering
    \includegraphics[width=0.49\textwidth]{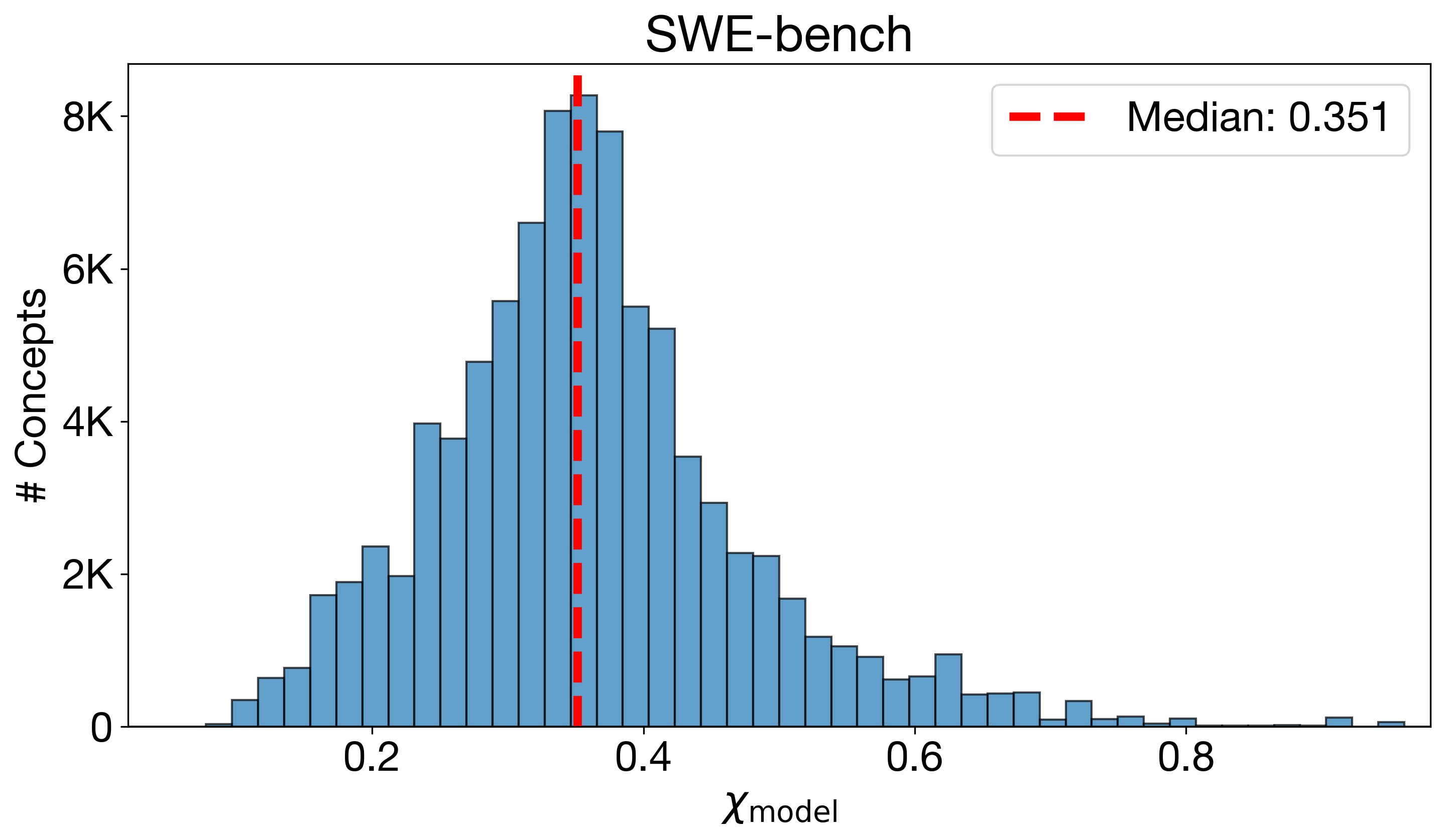}
    \caption{\textbf{Benchmark Performance.} The distribution of $\chi_{\text{model}}^{(b,c)}$ scores for SWE-Bench, using the SAE of Qwen3-4B.}
    \label{fig:swe_bench_performance}
\end{figure}

\begin{table}[h]
\centering
\caption{\textbf{Examples of Best- and Worst-Performing Concepts on SWE-Bench (Qwen3-4B).}}
\label{tab:swe_bench_performance_examples}
\begin{tabular}{>{\raggedright\arraybackslash}p{0.15\linewidth}
                p{0.12\linewidth}
                p{0.62\linewidth}}
\toprule
\textbf{} & \textbf{Concept ID} & \textbf{Concept Description} \\
\midrule
\addlinespace[0.6ex]
\textit{Best Performance}
    & \conc{\textbf{(45776)}}
    & Code/formatting issues \\
& \conc{\textbf{(49596)}}
    & Text excerpts \\
\addlinespace[0.8ex]
\textit{Worst Performance}
    & \conc{\textbf{(2511)}}
    & truth \\
& \conc{\textbf{(26712)}}
    & prepositions \\
\bottomrule
\end{tabular}
\end{table}

\paragraph{Repository Coverage Gaps.} CG identifies which kinds of software engineering topics SWE-Bench tests and which it omits. Among the most well-represented concepts are \conc{\textbf{(86659)}} ``\texttt{web development code}'', \conc{\textbf{(116868)}} ``\texttt{Data plots and analysis}'', and \conc{\textbf{(42119)}} ``\texttt{Code testing}'' --- consistent with the benchmark's inclusion of repositories like \texttt{django}, \texttt{matplotlib}, and \texttt{pytest}. Among the most underrepresented concepts are \conc{\textbf{(142338)}} ``\texttt{compilation}'' and \conc{\textbf{(141808)}} ``\texttt{computer security}''; consistent with this, none of the twelve repositories that make up SWE-Bench is a compiler or security project.

\FloatBarrier


\subsection{Terminal-Bench}
\label{app:add_results_conventional_terminal_bench}

\begin{figure}[h]
    \centering
    \includegraphics[width=0.49\textwidth]{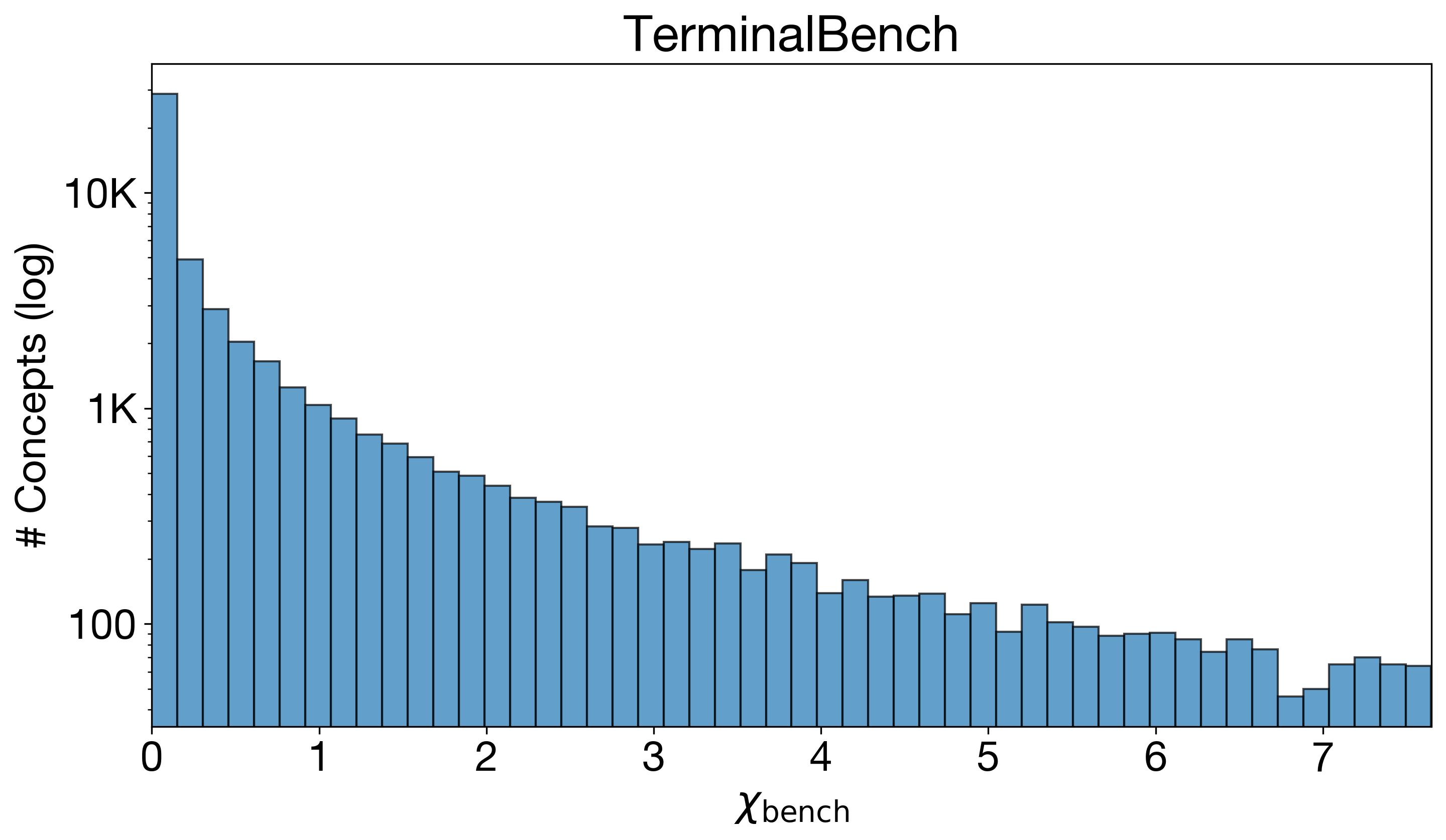}
    \caption{\textbf{Benchmark Coverage.} The distribution of $\chi_{\text{bench}}^{(b,c)}$ scores for Terminal-Bench, using the SAE of Qwen3-4B.}
    \label{fig:terminal_bench_coverage}
\end{figure}

\begin{figure}[h]
    \centering
    \includegraphics[width=0.49\textwidth]{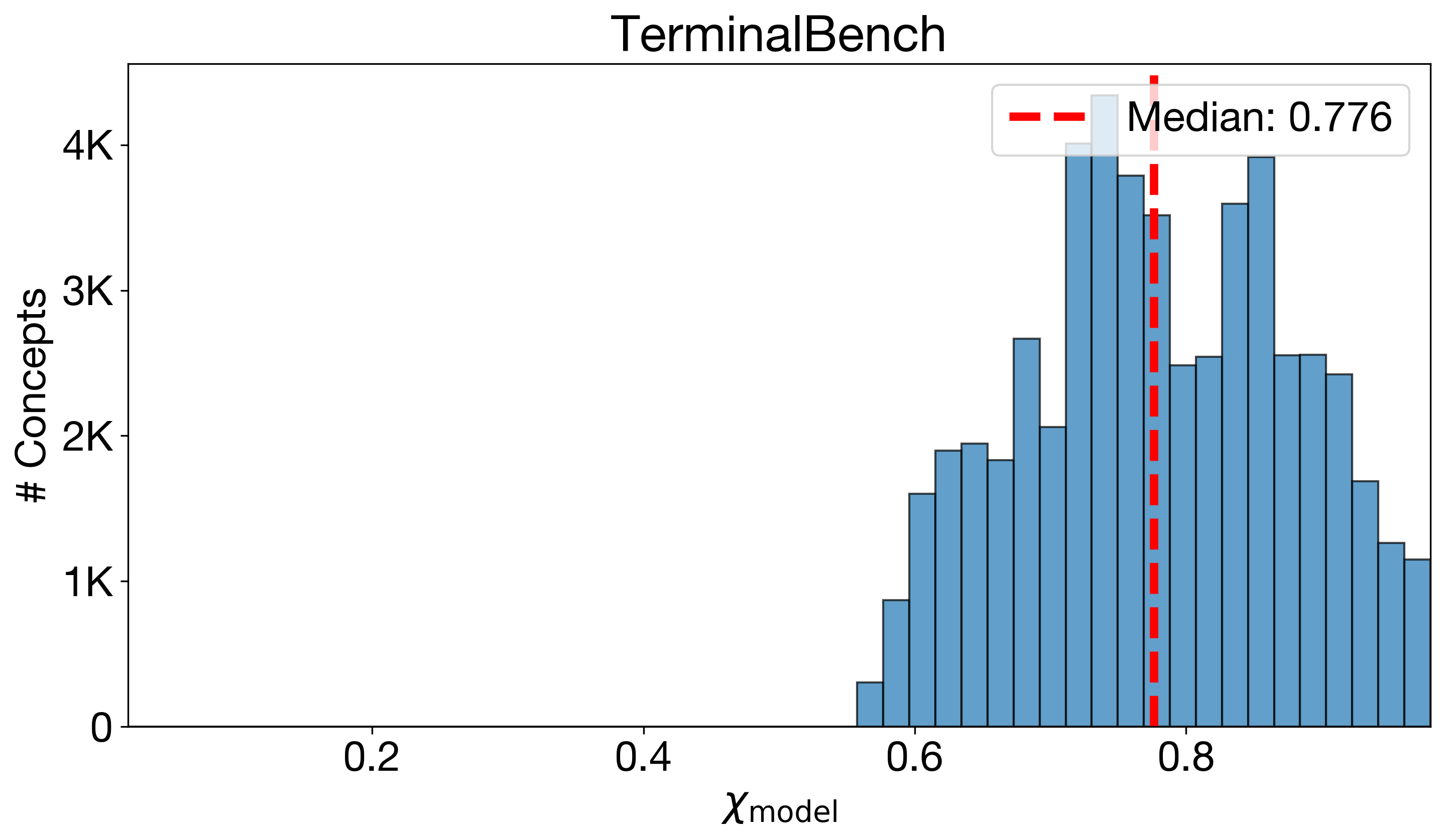}
    \caption{\textbf{Benchmark Performance.} The distribution of $\chi_{\text{model}}^{(b,c)}$ scores for Terminal-Bench, using the SAE of Qwen3-4B.}
    \label{fig:terminal_bench_performance}
\end{figure}

\FloatBarrier


\subsection{MMLU}
\label{app:add_results_conventional_mmlu}

\begin{figure}[h]
    \centering
    \includegraphics[width=0.49\textwidth]{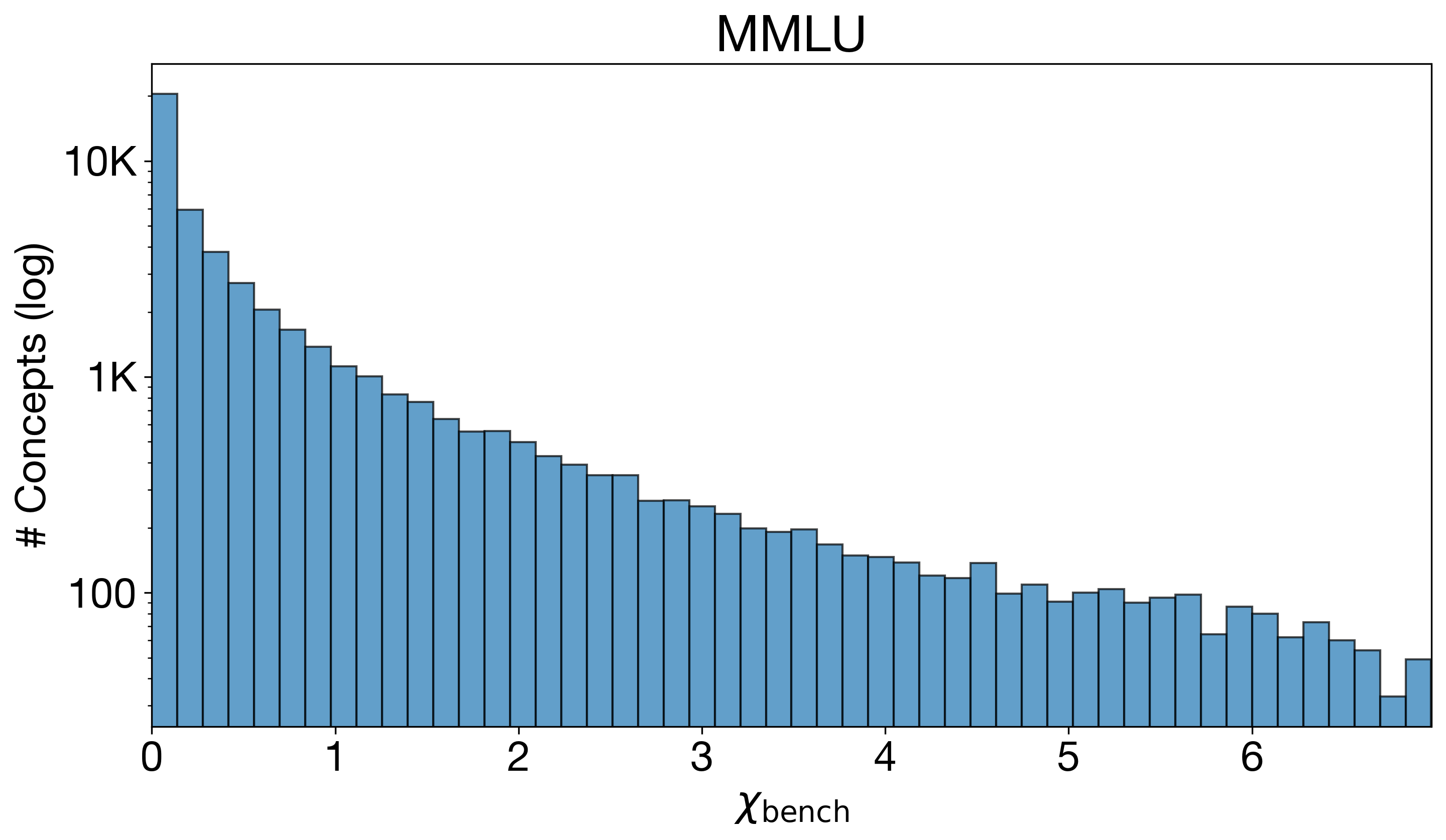}
    \caption{\textbf{Benchmark Coverage.} The distribution of $\chi_{\text{bench}}^{(b,c)}$ scores for MMLU, using the SAE of Qwen3-4B.}
    \label{fig:mmlu_coverage}
\end{figure}

\begin{table}[h]
\centering
\caption{\textbf{Examples of Best- and Worst-Coverage Concepts in MMLU (Qwen3-4B).}}
\label{tab:mmlu_coverage_examples}
\begin{tabular}{>{\raggedright\arraybackslash}p{0.15\linewidth}
                p{0.12\linewidth}
                p{0.62\linewidth}}
\toprule
\textbf{} & \textbf{Concept ID} & \textbf{Concept Description} \\
\midrule
\addlinespace[0.6ex]
\textit{Best Coverage}
    & \conc{\textbf{(2882)}}
    & abstract \\
& \conc{\textbf{(53686)}}
    & assertion/claim \\
\addlinespace[0.8ex]
\textit{Worst Coverage}
    & \conc{\textbf{(24577)}}
    & History articles \\
& \conc{\textbf{(46972)}}
    & Medical studies and patients \\
\bottomrule
\end{tabular}
\end{table}

\begin{figure}[h]
    \centering
    \includegraphics[width=0.49\textwidth]{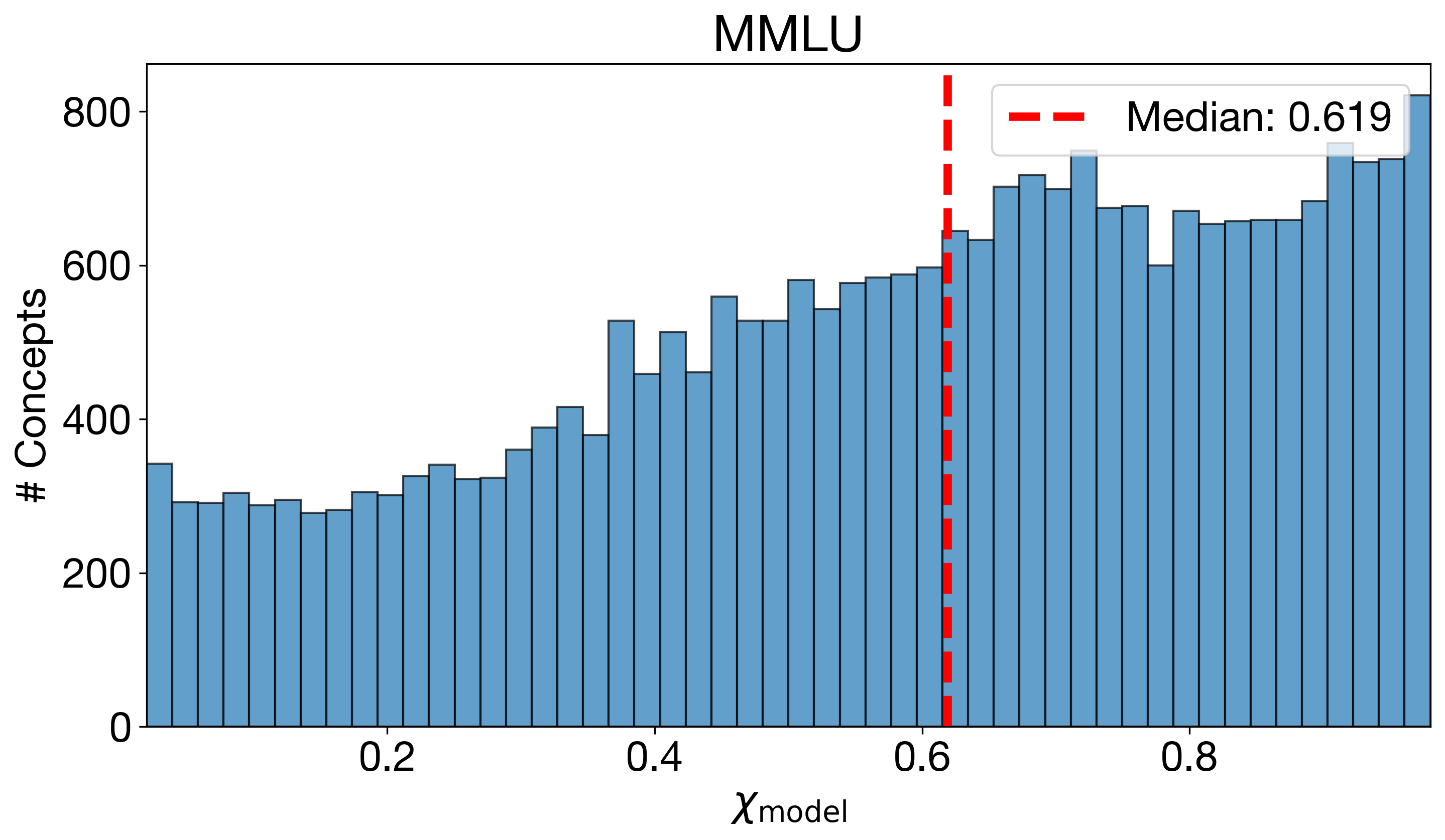}
    \caption{\textbf{Benchmark Performance.} The distribution of $\chi_{\text{model}}^{(b,c)}$ scores for MMLU, using the SAE of Qwen3-4B.}
    \label{fig:mmlu_performance}
\end{figure}

\begin{table}[h]
\centering
\caption{\textbf{Examples of Best- and Worst-Performing Concepts on MMLU (Qwen3-4B).}}
\label{tab:mmlu_performance_examples}
\begin{tabular}{>{\raggedright\arraybackslash}p{0.15\linewidth}
                p{0.12\linewidth}
                p{0.62\linewidth}}
\toprule
\textbf{} & \textbf{Concept ID} & \textbf{Concept Description} \\
\midrule
\addlinespace[0.6ex]
\textit{Best Performance}
    & \conc{\textbf{(44476)}}
    & Scientific/academic text \\
& \conc{\textbf{(45577)}}
    & Programming code \\
\addlinespace[0.8ex]
\textit{Worst Performance}
    & \conc{\textbf{(47139)}}
    & dates \\
& \conc{\textbf{(33078)}}
    & Punctuation \\
\bottomrule
\end{tabular}
\end{table}

\paragraph{Historical Coverage Gaps.} Inspecting the lowest combinations of $\chi_{\text{bench}}^{(b,c)}$ scores in MMLU surfaces two pairings of concepts with substantive historical content but near-zero co-coverage: \conc{\textbf{(159486)}} ``\texttt{Historical events and organizations}'' with \conc{\textbf{(71132)}} ``\texttt{Saudi Arabia/Middle East}'' (suggestive of the Arab Spring), and \conc{\textbf{(102854)}} ``\texttt{Cuba}'' with \conc{\textbf{(28317)}} ``\texttt{invasions or military conflict}'' (suggestive of the Cuban Missile Crisis). A keyword search and manual review of MMLU confirmed that the benchmark contains no questions about either topic, despite both being canonical in modern history curricula.

\FloatBarrier


\subsection{GPQA}
\label{app:add_results_conventional_gpqa}

\begin{figure}[h]
    \centering
    \includegraphics[width=0.49\textwidth]{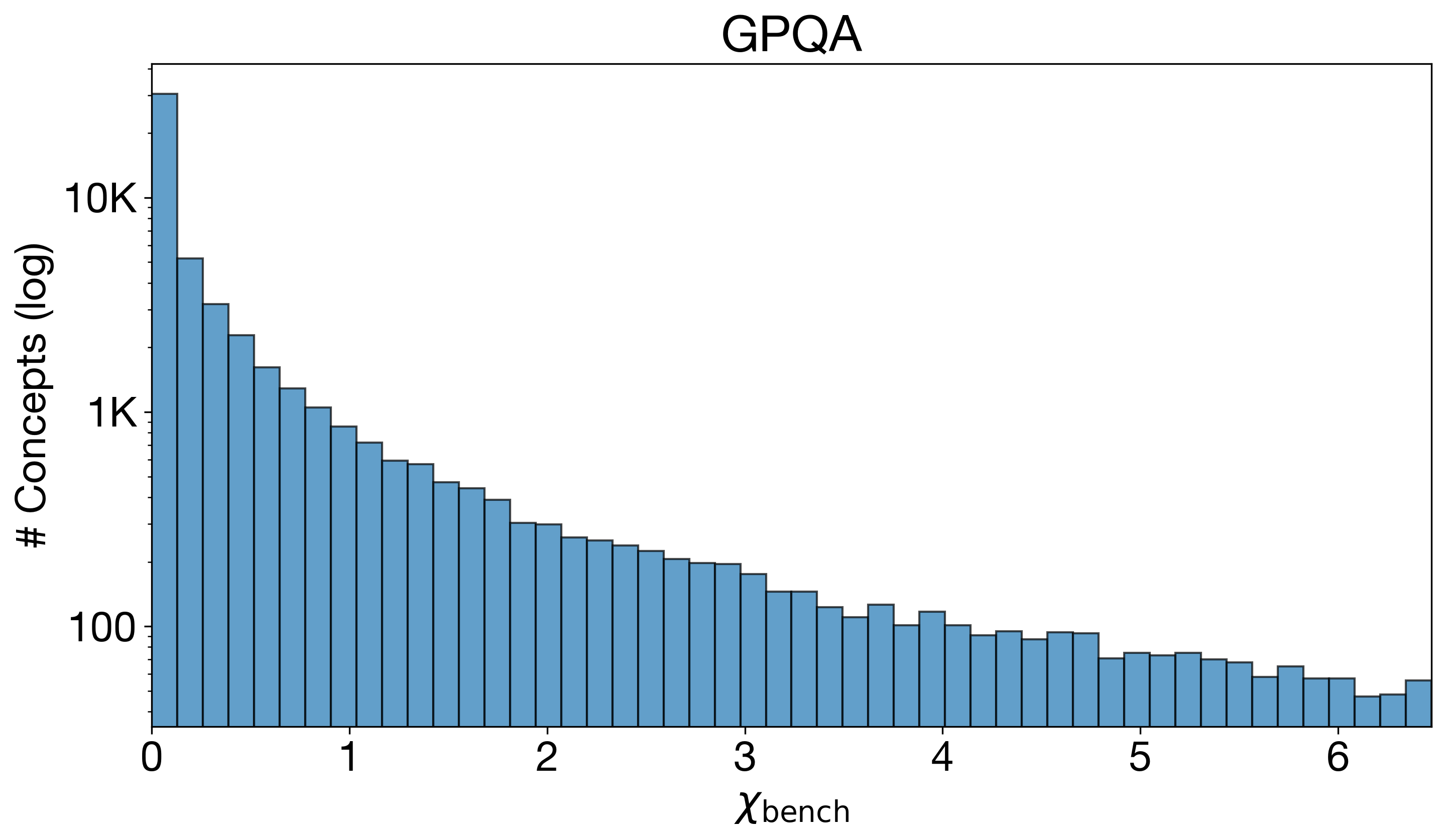}
    \caption{\textbf{Benchmark Coverage.} The distribution of $\chi_{\text{bench}}^{(b,c)}$ scores for GPQA, using the SAE of Qwen3-4B.}
    \label{fig:gpqa_coverage}
\end{figure}

\begin{figure}[h]
    \centering
    \includegraphics[width=0.49\textwidth]{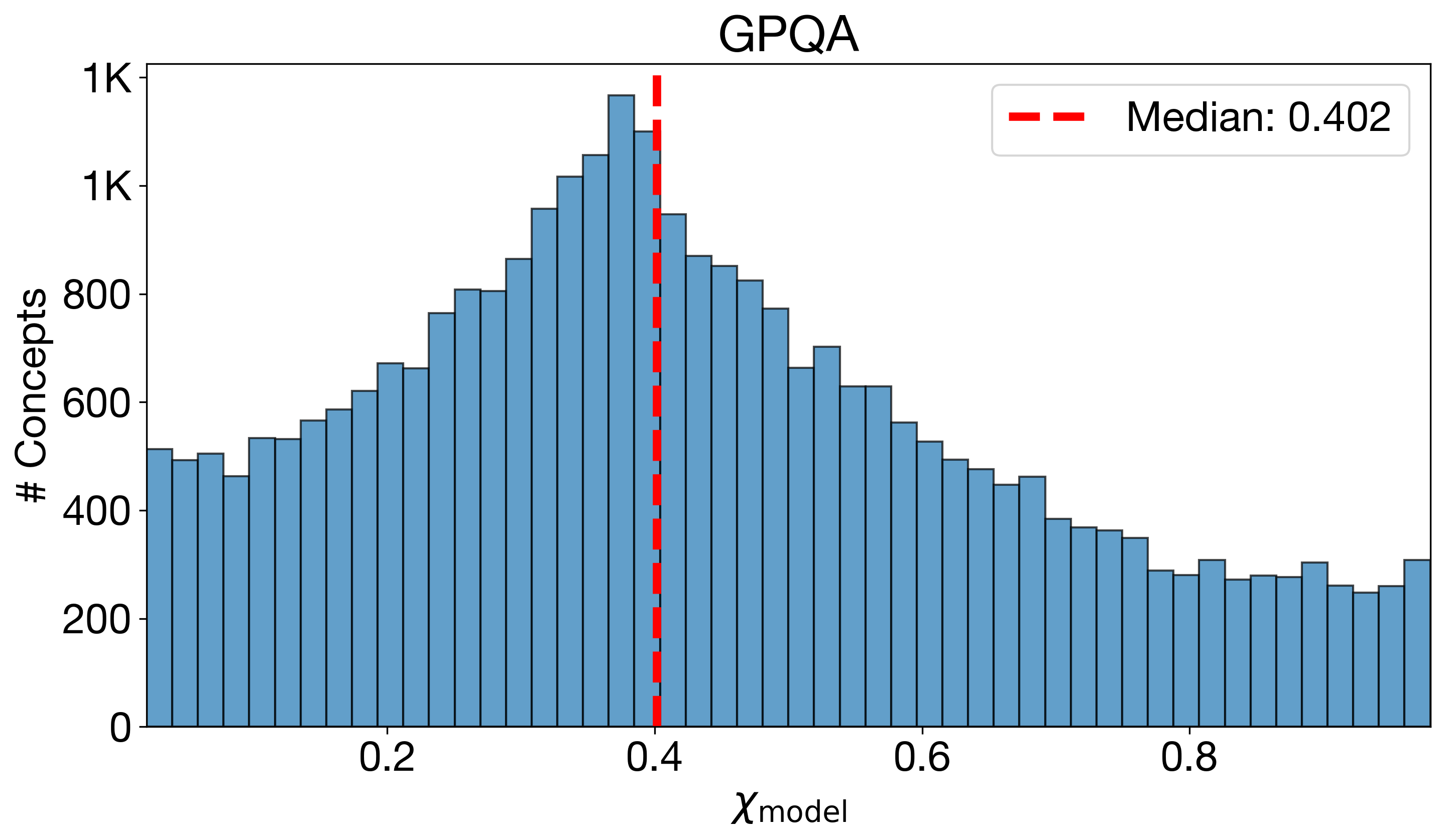}
    \caption{\textbf{Benchmark Performance.} The distribution of $\chi_{\text{model}}^{(b,c)}$ scores for GPQA, using the SAE of Qwen3-4B.}
    \label{fig:gpqa_performance}
\end{figure}

\FloatBarrier


\newpage
\subsection{Humanity's Last Exam (HLE)}
\label{app:add_results_conventional_hle}

\begin{figure}[h]
    \centering
    \includegraphics[width=0.49\textwidth]{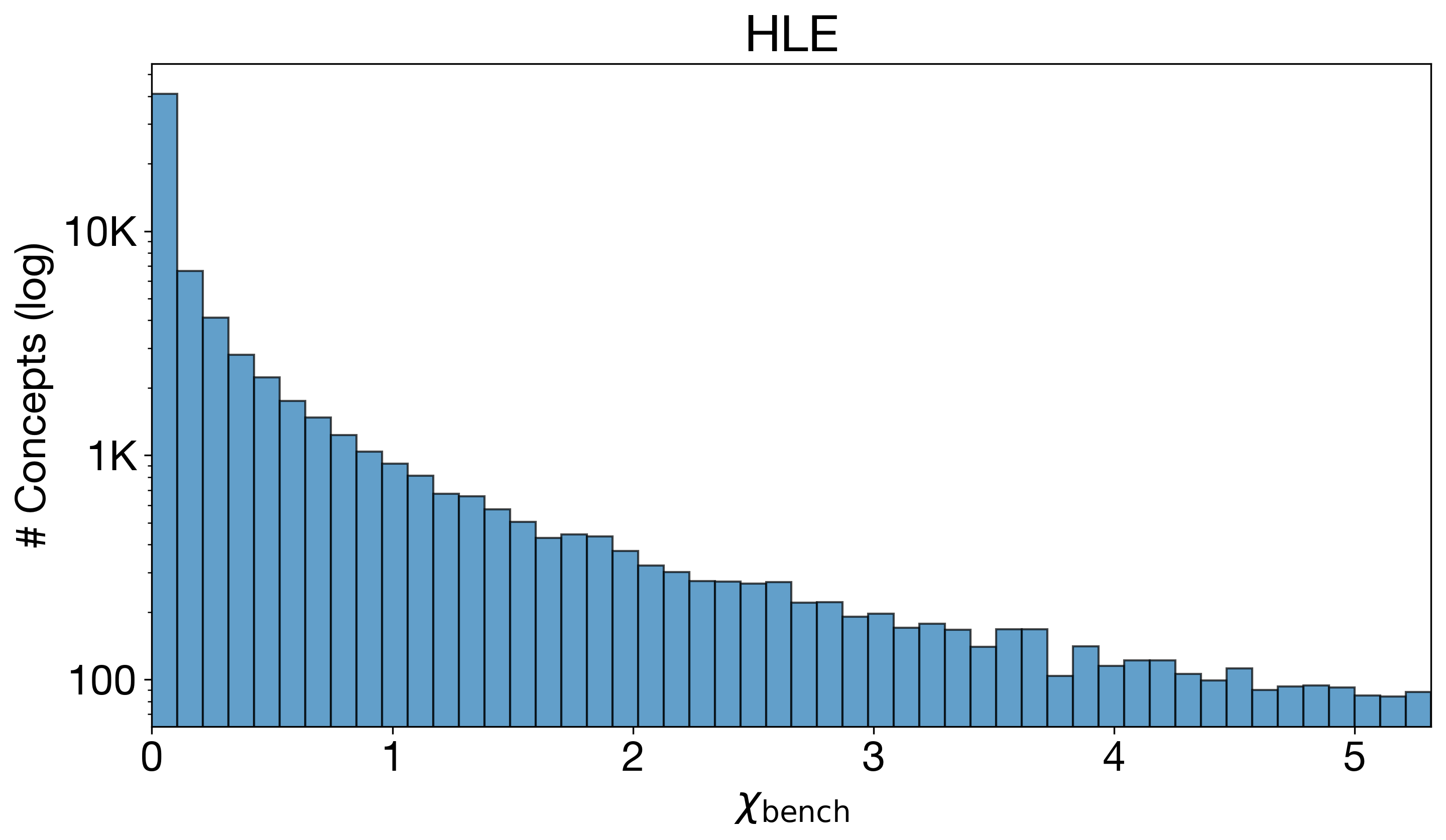}
    \caption{\textbf{Benchmark Coverage.} The distribution of $\chi_{\text{bench}}^{(b,c)}$ scores for HLE, using the SAE of Qwen3-4B.}
    \label{fig:hle_coverage}
\end{figure}

\begin{figure}[h]
    \centering
    \includegraphics[width=0.49\textwidth]{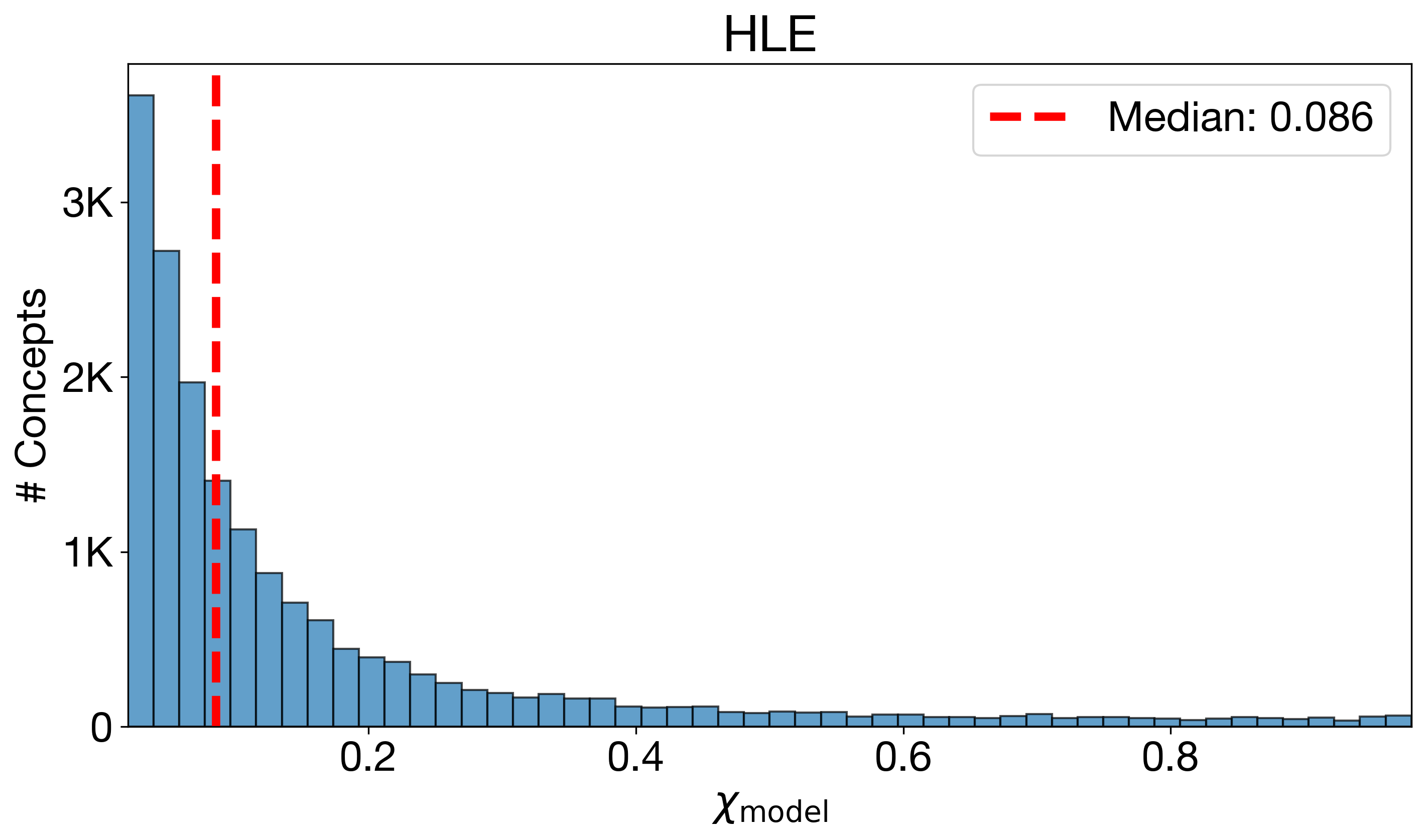}
    \caption{\textbf{Benchmark Performance.} The distribution of $\chi_{\text{model}}^{(b,c)}$ scores for HLE, using the SAE of Qwen3-4B.}
    \label{fig:hle_performance}
\end{figure}

\FloatBarrier


\newpage
\section{Prompts for LLM Analysis of CG Results}
\label{app:prompts}

To help filter through a large number of results, we used an LLM (Gemini 2.5 Pro) to sift through results. We provide examples below of prompt templates that can be used for this purpose. Segments highlighted like \texttt{<THIS>} are to be replaced with data for the case at hand.

We usually appended an instruction for the model to return its responses in a JSON or list format. Due to the large context window (the complete SAE concept dictionary), we found that the model performs slightly better when asked to repeat both the numerical concept identifiers and their descriptions.

\subsection{Benchmark Gaps: Missing Concepts (Cross-Benchmark)}
\label{appsub:prompt_missing_concepts_bench_specific}

\begin{promptbox}
Below is a list of concepts in a large language model. Each concept has an ID and a description. Are any of these concepts *critical* to the evaluation of large language models? Such concepts generally span topics of safety (toxic language, harm, bias, etc.), performance (reasoning ability, math, coding, etc.), and metacognition (ability to reject responses, reasoning about instructions, etc.).  Choose from the list of concepts below. List all such relevant concepts. Do not summarize or group; list all concepts verbatim as they appear below if they are relevant. \\

LLM CONCEPTS:

\texttt{<AVAILABLE\_CONCEPTS>}
\end{promptbox}

\subsection{Benchmark Gaps: Missing Concepts (Per-Benchmark)}
\label{appsub:prompt_missing_concepts_bench_specific}

\begin{promptbox}
Below is a list of concepts in a large language model. Each concept has an ID and a description. Are any of these concepts *absolutely critical* for the evaluation of the \texttt{<BENCHMARK\_NAME>} benchmark, as defined below? Choose from the list of concepts below. List all such relevant concepts. Do not summarize or group; list all concepts verbatim as they appear below if they are relevant. \\

BENCHMARK DEFINITION:

\texttt{<BENCHMARK\_DEFINITION>} \\

LLM CONCEPTS:

\texttt{<AVAILABLE\_CONCEPTS>}
\end{promptbox}

\subsection{Benchmark Gaps: Matching}
\label{appsub:prompt_matching}

\begin{promptbox}
Below, there is (1) a list of Competency Gaps concepts and (2) a list of \texttt{<OTHER\_FRAMEWORK>} categories. \\

For each category from \texttt{<OTHER\_FRAMEWORK>}, determine whether there are any corresponding Competency Gaps concepts. If no relevant concepts exist, leave this blank. \\

If there are multiple such concepts, include only the top \texttt{<MATCHING\_LIMIT>} most representative ones. Do not include more than \texttt{<MATCHING\_LIMIT>} concepts per category. \\

(1) COMPETENCY GAPS CONCEPTS:

\texttt{<AVAILABLE\_CONCEPTS>} \\

(2) \texttt{<OTHER\_FRAMEWORK>} CONCEPTS:

\texttt{<OTHER\_FRAMEWORK\_CONCEPTS>}
\end{promptbox}

\newpage
\onecolumn


\newpage
\section{LLM-as-Judge Per-Benchmark Concept Relevance}
\label{app:relevance}

To complement the unsupervised CG analysis, we run an LLM-as-judge step that labels each SAE concept as semantically relevant or irrelevant to each benchmark in the suite, yielding a per-concept boolean column per benchmark. We use Gemini 2.5 Flash, with concepts batched in groups of $20$ and all benchmark definitions included in the prompt; the model returns a JSON array of booleans, one per (concept, benchmark) pair. The exact relevance criterion given to the judge is whether \emph{activating on the concept would plausibly help or hurt benchmark accuracy}. Annotated relevance labels are released as a CSV alongside our code, and CG scores can be recomputed on the relevance-filtered concept subset as an alternative analysis option.

The prompt template used for the relevance annotation is reproduced below:

\begin{promptbox}
You are classifying SAE (sparse autoencoder) concepts by benchmark relevance. \\

\textbf{Benchmarks:}

\texttt{<BENCHMARK\_DEFINITIONS>} \\

For each concept below, output a JSON array where each element is an object with keys \texttt{<KEYS>} and boolean values (\texttt{true}/\texttt{false}) indicating whether that concept is likely relevant to each benchmark. A concept is relevant if activating on it would plausibly help or hurt accuracy. \\

\textbf{Concepts:}

\texttt{<CONCEPTS>} \\

Respond ONLY with a valid JSON array of \texttt{<N>} objects, one per concept, in order. No explanation, no markdown fences.
\end{promptbox}

Tables~\ref{tab:relevance_gemma} and~\ref{tab:relevance_llama} report the resulting per-benchmark relevance percentages for the Gemma 2 2B and Llama 3.1 8B SAE dictionaries respectively. The relevant fraction varies widely across benchmarks, reinforcing that no single benchmark exercises more than a slice of the concept space: open-ended factual benchmarks (Natural Questions, AGI Eval, Vectara) exercise the broadest slice, while narrow-format benchmarks (WinoGrande, GSM8K) are the most concentrated. Both models follow the same ranking pattern across benchmarks despite using independently trained SAEs, suggesting that the relevance structure is benchmark-driven rather than SAE-specific.

\begin{table}[h]
\centering
\caption{\textbf{LLM-as-Judge Per-Benchmark Concept Relevance: Gemma 2 2B.}}
\label{tab:relevance_gemma}
\small
\begin{tabular}{lrrr}
\toprule
\textbf{Benchmark} & \textbf{Relevant} & \textbf{Annotated} & \textbf{\%} \\
\midrule
Natural Questions  & 8{,}544 & 16{,}104 & 53.1\% \\
AGI Eval           & 7{,}733 & 16{,}104 & 48.0\% \\
Vectara            & 7{,}135 & 16{,}104 & 44.3\% \\
SWE-Bench          & 5{,}153 & 16{,}104 & 32.0\% \\
Terminal-Bench     & 3{,}670 & 16{,}104 & 22.8\% \\
Social IQA          & 3{,}632 & 16{,}104 & 22.6\% \\
BBQ                & 3{,}173 & 16{,}104 & 19.7\% \\
LogicBench         & 3{,}061 & 16{,}104 & 19.0\% \\
CROWS Pairs        & 2{,}433 & 16{,}104 & 15.1\% \\
MATH               & 2{,}198 & 16{,}104 & 13.6\% \\
GSM8K              & 1{,}908 & 16{,}104 & 11.8\% \\
WinoGrande         & 1{,}320 & 16{,}104 &  8.2\% \\
\bottomrule
\end{tabular}
\end{table}

\begin{table}[h]
\centering
\caption{\textbf{LLM-as-Judge Per-Benchmark Concept Relevance: Llama 3.1 8B.}}
\label{tab:relevance_llama}
\small
\begin{tabular}{lrrr}
\toprule
\textbf{Benchmark} & \textbf{Relevant} & \textbf{Annotated} & \textbf{\%} \\
\midrule
Vectara            & 15{,}616 & 45{,}199 & 34.5\% \\
AGI Eval           & 14{,}900 & 45{,}199 & 33.0\% \\
Natural Questions  & 13{,}871 & 45{,}199 & 30.7\% \\
SWE-Bench          & 13{,}308 & 45{,}199 & 29.4\% \\
Social IQA          &  9{,}445 & 45{,}199 & 20.9\% \\
Terminal-Bench     &  9{,}374 & 45{,}199 & 20.7\% \\
LogicBench         &  7{,}846 & 45{,}199 & 17.4\% \\
BBQ                &  6{,}415 & 45{,}199 & 14.2\% \\
CROWS Pairs        &  5{,}383 & 45{,}199 & 11.9\% \\
MATH               &  5{,}041 & 45{,}199 & 11.2\% \\
GSM8K              &  4{,}423 & 45{,}199 &  9.8\% \\
WinoGrande         &  2{,}232 & 45{,}199 &  4.9\% \\
\bottomrule
\end{tabular}
\end{table}

\FloatBarrier


\newpage
~
\newpage
\section{Exploratory Web Application: Additional Screenshots}
\label{app:screenshots}

\begin{figure*}[h]
    \centering
    \boxed{\includegraphics[width=0.8\linewidth]{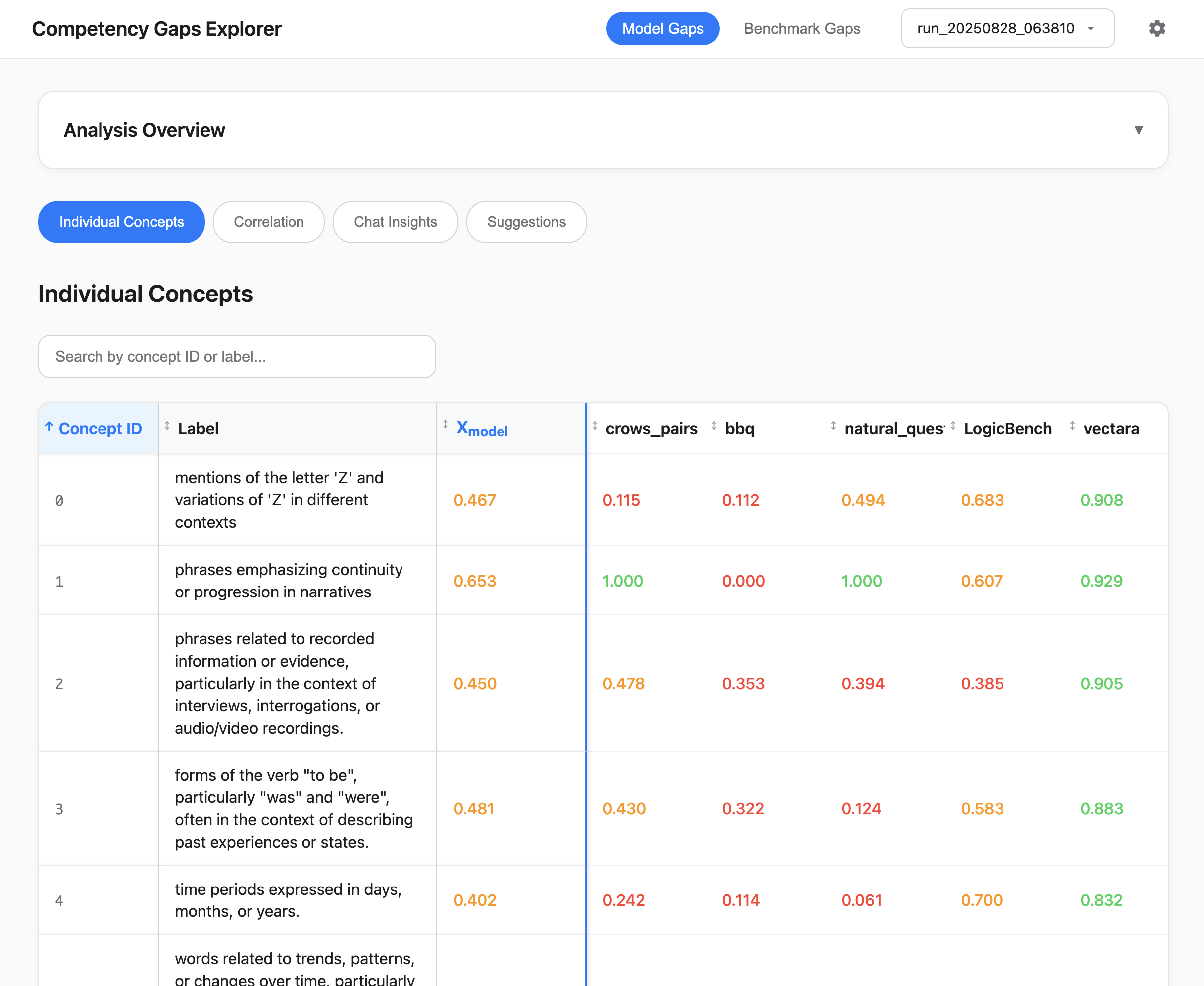}}
    \caption{Web Application Screenshot: An overview of all concepts for the Model Gaps analysis.}
    \label{fig:web_app_screenshot_0}
\end{figure*}

\begin{figure*}[h]
    \centering
    \boxed{\includegraphics[width=0.8\linewidth]{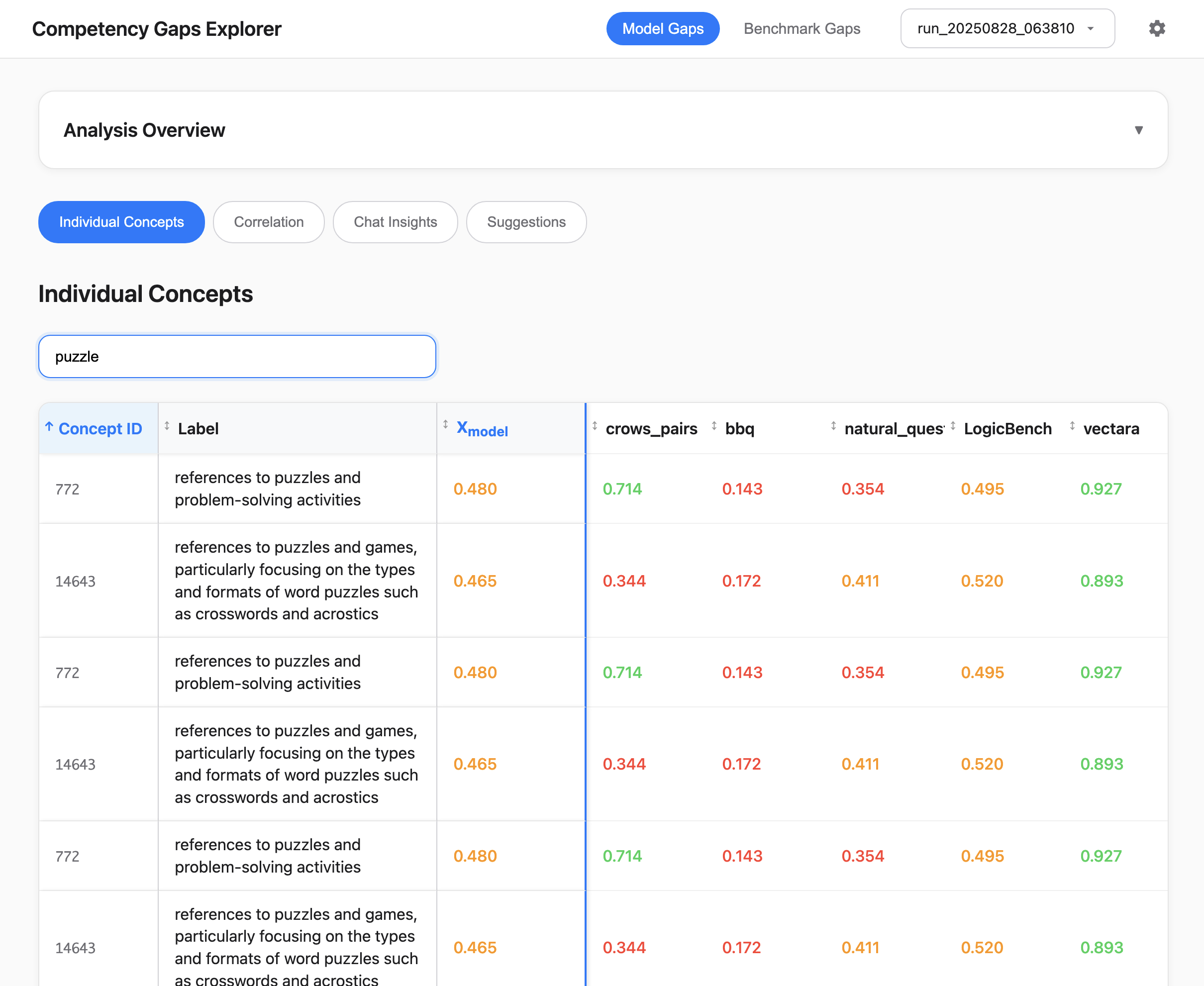}}
    \caption{Web Application Screenshot: Keyword-filtered concepts for the Model Gaps analysis.}
    \label{fig:web_app_screenshot_1}
\end{figure*}

\begin{figure*}[h]
    \centering
    \boxed{\includegraphics[width=0.8\linewidth]{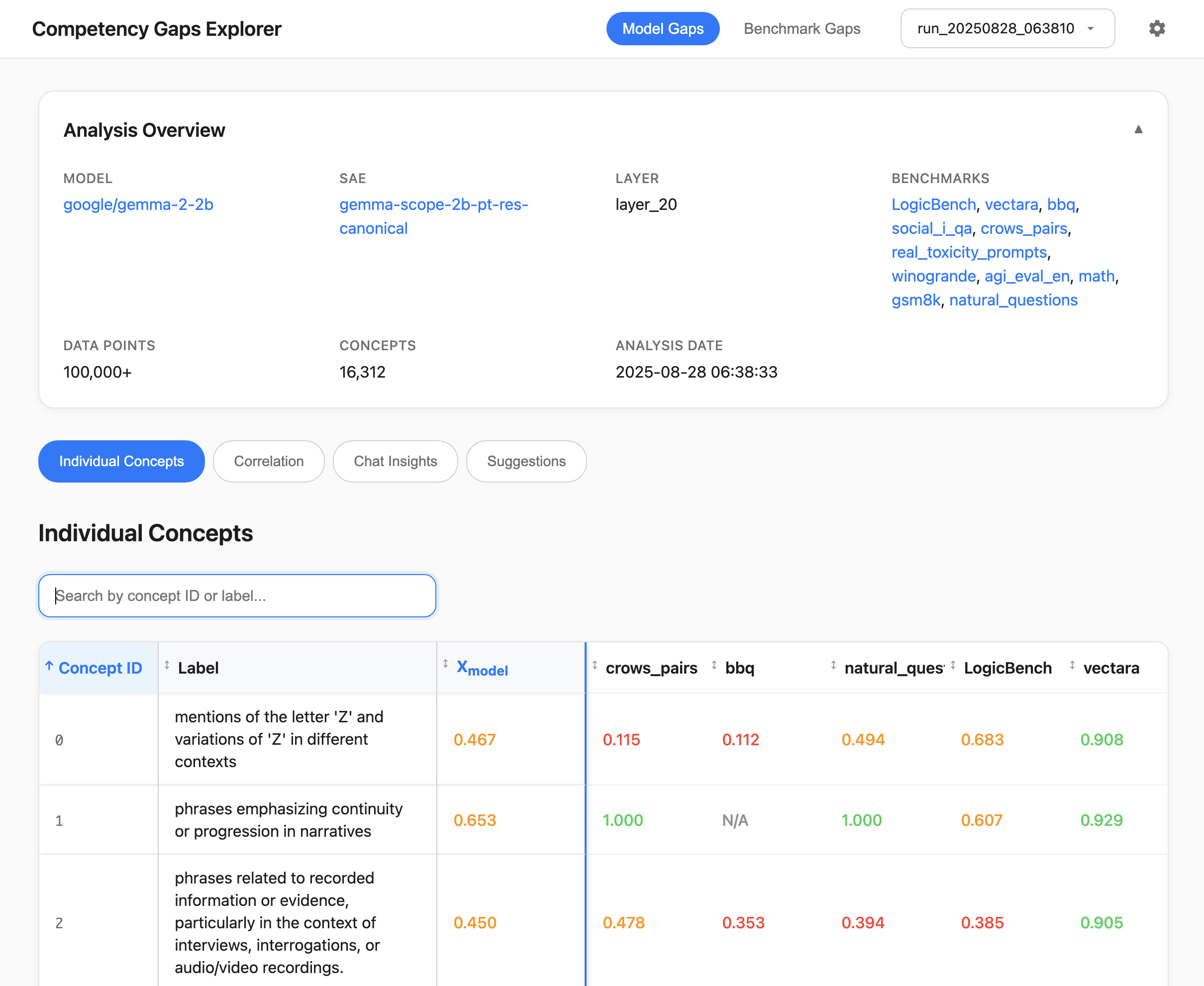}}
    \caption{Web Application Screenshot: Expandable view with the analysis metadata.}
    \label{fig:web_app_screenshot_2}
\end{figure*}

\begin{figure*}[h]
    \centering
    \boxed{\includegraphics[width=0.8\linewidth]{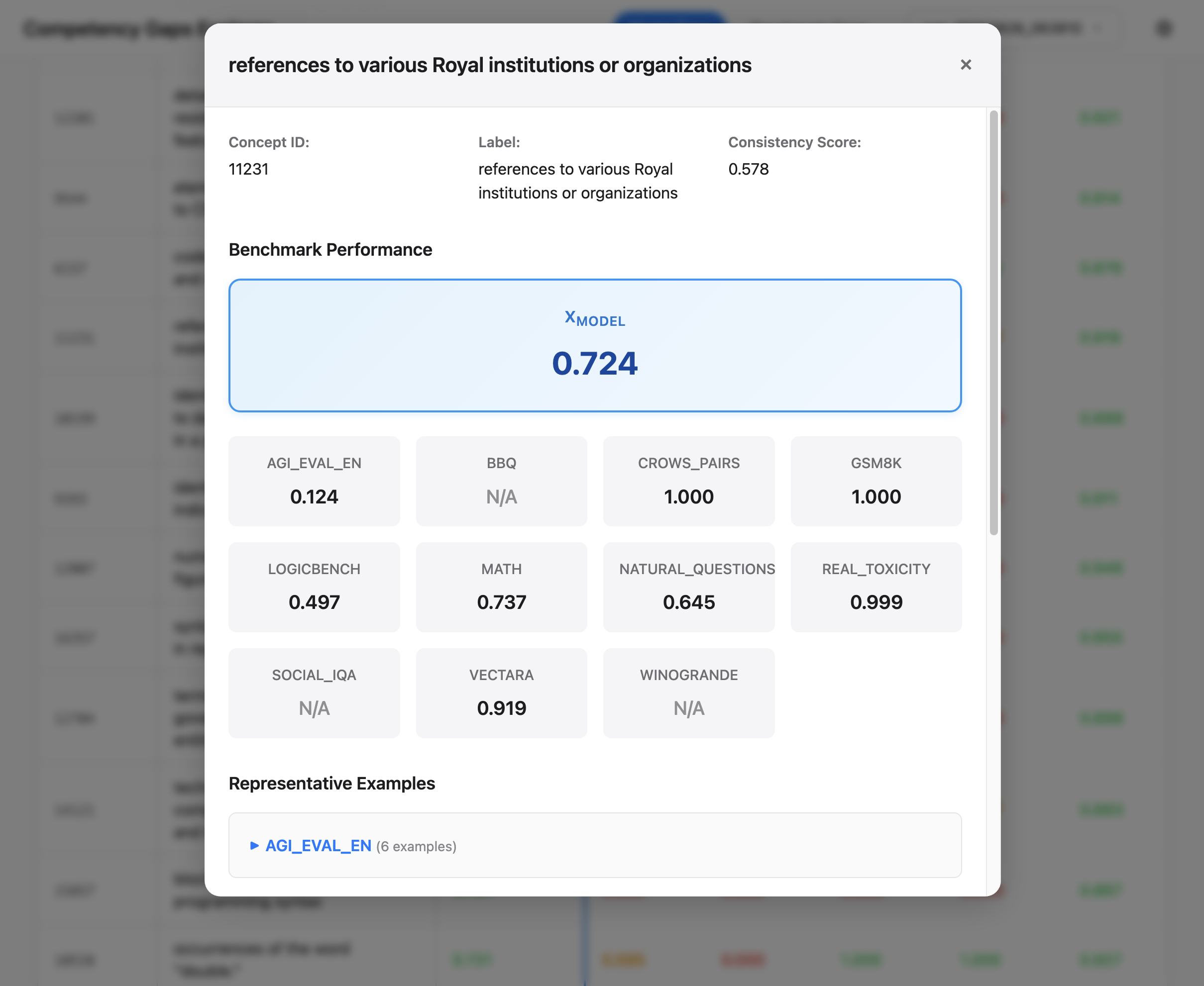}}
    \caption{Web Application Screenshot: Concept detail within the Model Gaps analysis, summarizing the performance of this concept across benchmarks.}
    \label{fig:web_app_screenshot_3}
\end{figure*}

\begin{figure*}[h]
    \centering
    \boxed{\includegraphics[width=0.8\linewidth]{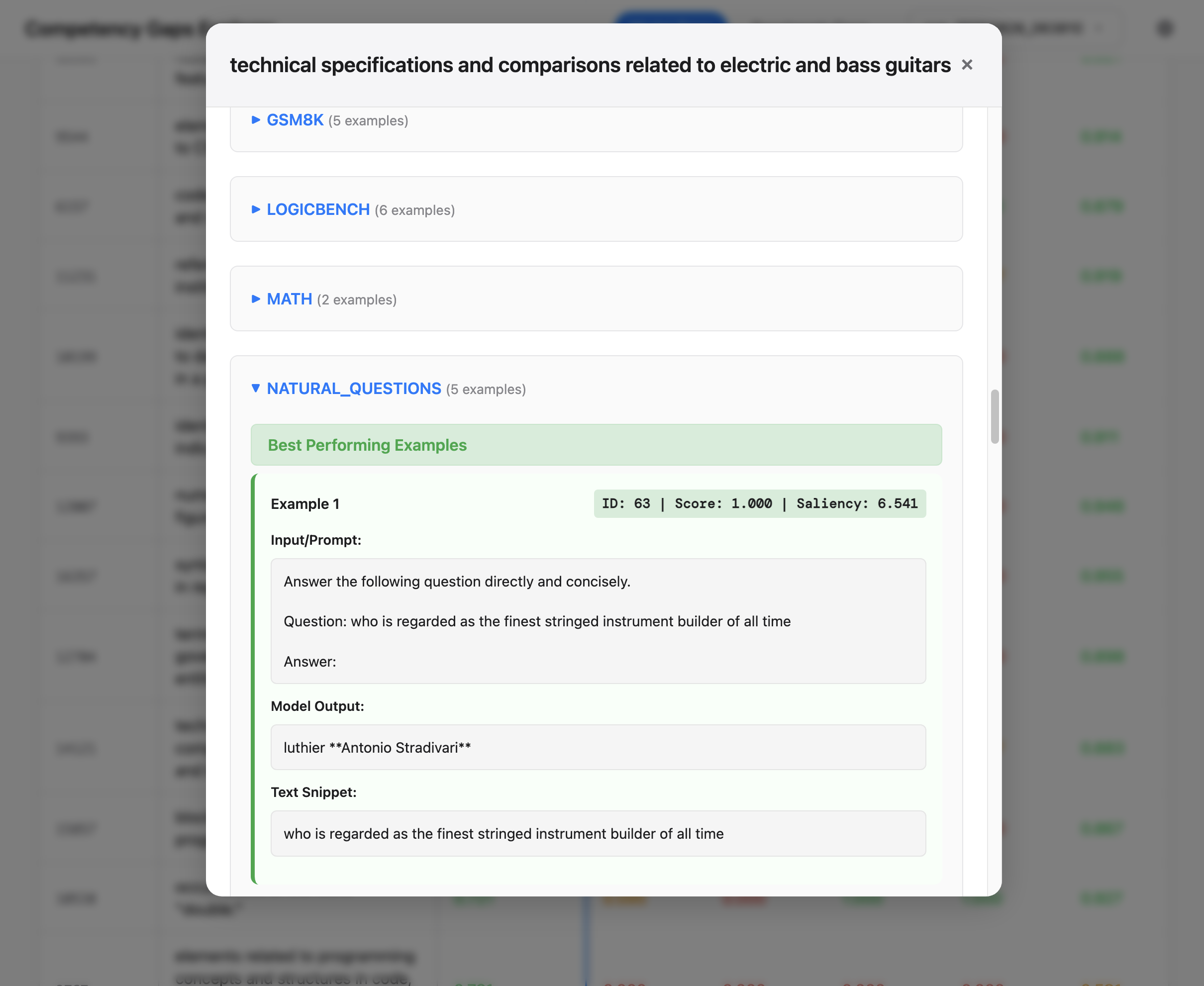}}
    \caption{Web Application Screenshot: Examples of data points where the model performed well and the concept at hand shows high activation.}
    \label{fig:web_app_screenshot_4}
\end{figure*}

\begin{figure*}[h]
    \centering
    \boxed{\includegraphics[width=0.8\linewidth]{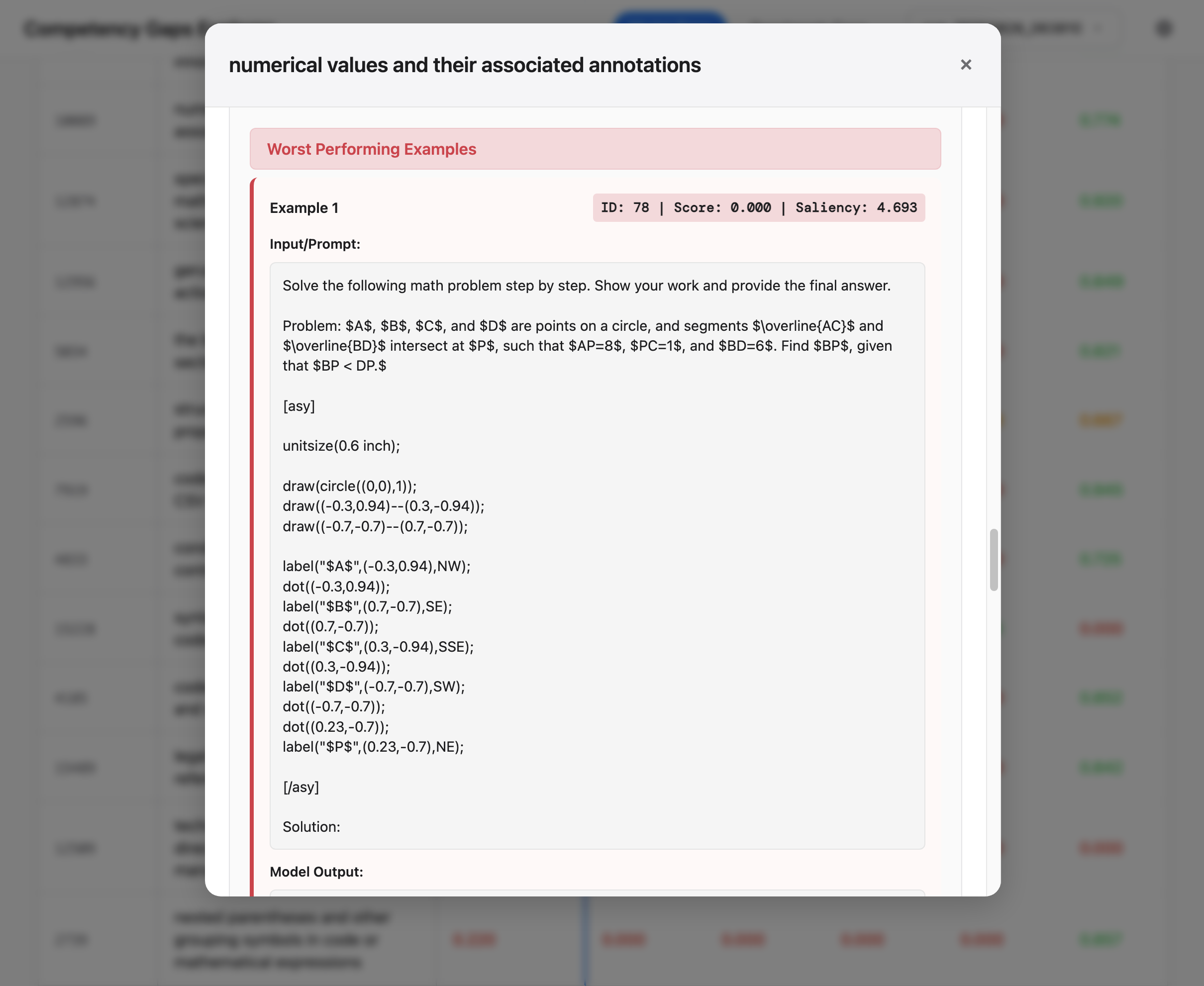}}
    \caption{Web Application Screenshot: Examples of data points where the model performed poorly despite the concept at hand showing high activation.}
    \label{fig:web_app_screenshot_5}
\end{figure*}

\begin{figure*}[h]
    \centering
    \boxed{\includegraphics[width=0.8\linewidth]{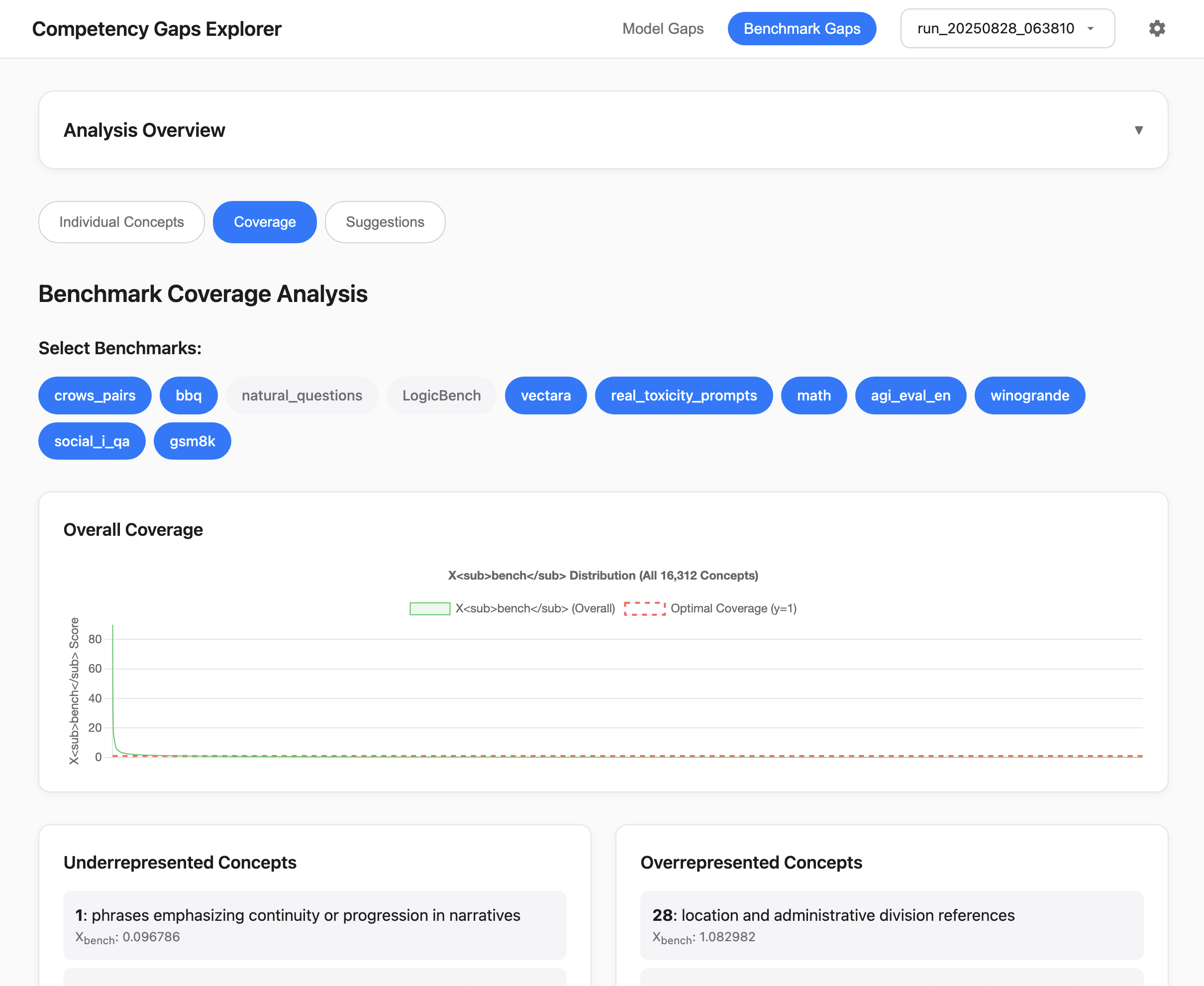}}
    \caption{Web Application Screenshot: Coverage visualization comparing the coverage and distribution of concepts across different combinations of analyzed benchmarks.}
    \label{fig:web_app_screenshot_6}
\end{figure*}

\begin{figure*}[h]
    \centering
    \boxed{\includegraphics[width=0.8\linewidth]{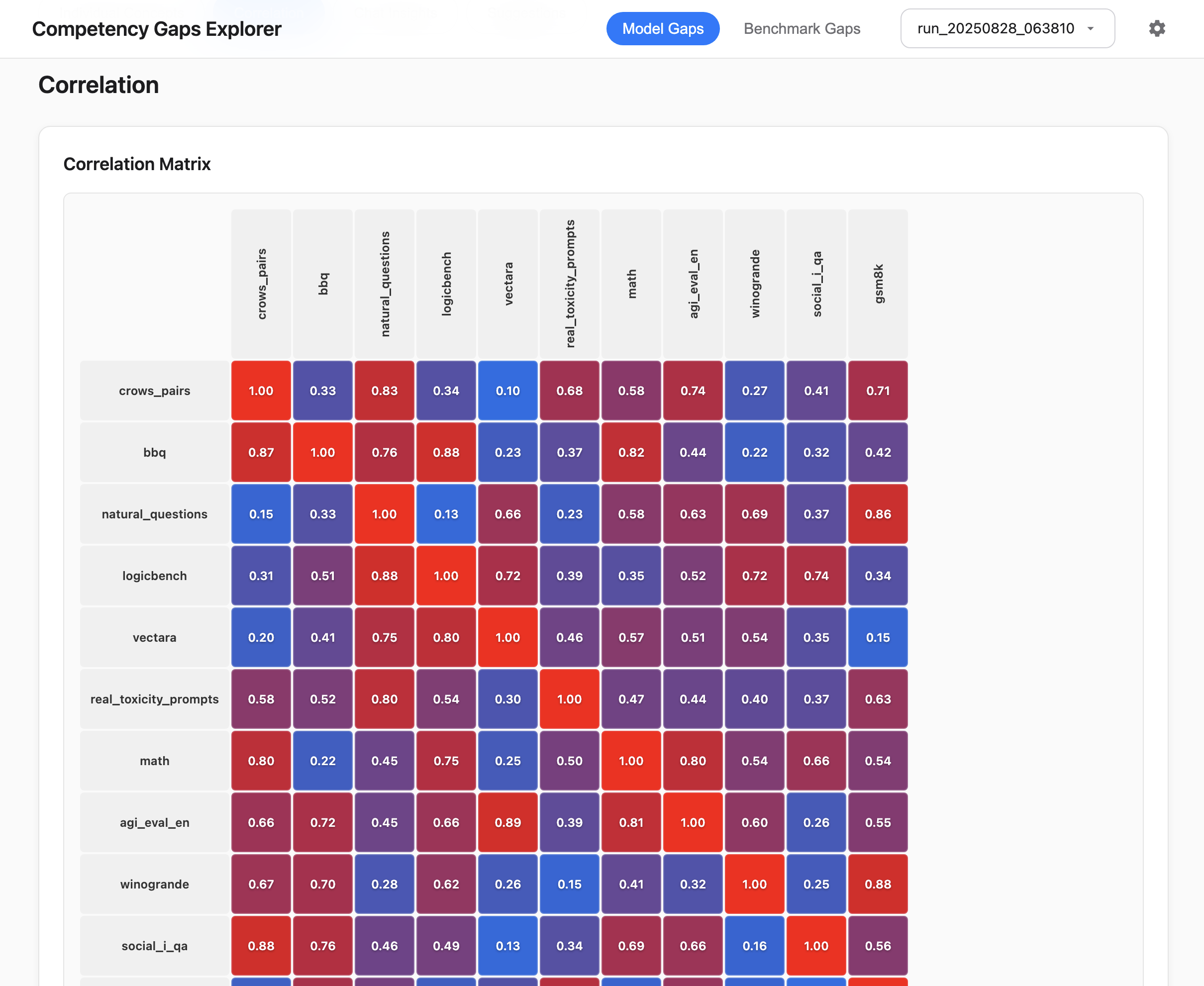}}
    \caption{Web Application Screenshot: Grid showing the correlation of scores across benchmarks.}
    \label{fig:web_app_screenshot_7}
\end{figure*}

\FloatBarrier


\section{Comparison with Other Methods}
\label{app:comparison_other_methods}

\subsection{Comparison Methodology}
\label{app:comparison_methodology}

We compare our method to some of the related methods from Section \ref{sec:rel_work} -- those that relate to the discovery of benchmark and model gaps. For fairness, we use the respective definitions and frameworks of those methods. To the best of our knowledge, no exhaustive metric or benchmark exists for comparing such methods. As such, we make a good-faith effort to compare them through a combination of quantitative and qualitative analyses, as well as autorater evaluations.

Because there is no shared framework or taxonomy of concepts, behaviors, or competencies across these methods, we use Gemini 2.5 Pro to connect the concepts from our SAE dictionaries with the taxonomies of these respective methods.


\subsection{Benchmark Gaps Overview}
\label{app:baseline_comp_benchmark_gaps_overview}

\begin{table*}[h]
\centering
\caption{\textbf{Comparison of Methods for Evaluating Benchmark Gaps.} Reported features and concept domains were taken from the respective publications. \textbf{Automation:} The method is fully automated and runs without human intervention. \textbf{Cross-Bench Comparability:} The method enables combined or comparative evaluation across different benchmarks. \textbf{Missing Concept Identification:} The method surfaces relevant concepts that are absent from the benchmark. \textbf{Dynamic Data:} The method can be applied to new datasets as they emerge (i.e., it is not restricted to a fixed, hard-coded dataset). \textbf{Interactive Tooling:} The method includes an interactive exploration tool. \textbf{Improvement Suggestion:} With minor modifications or extensions, the method can inform future benchmark design or selection.}
\label{tab:bench_gap_comp}
\renewcommand{\arraystretch}{1.4}
\begin{tabularx}{\textwidth}{|>{\raggedright\arraybackslash}X|>{\centering\arraybackslash}X|>{\centering\arraybackslash}X|>{\centering\arraybackslash}X|>{\centering\arraybackslash}X|}
\hline
 & \textbf{Arena-Hard-Auto} & \textbf{Benchmarker Suite} & \textbf{SafetyPrompts} & \textbf{CG (Ours)} \\
\hline
\textbf{Concept Dictionary Size} & 750 & N/A & N/A & 16,000+ \\
\textbf{Concept Dictionary Domains} & Technical, Creative, Academic, Real-world applications & Language, Knowledge, Reasoning, Comprehensive Examination, Understanding & Safety & Diverse semantic and syntactic forms, concepts, and methodologies \\
\textbf{Observation Space} & Behavior & Behavior & Behavior & Behavior + Model Internals \\
\hline
\textbf{Automation\vspace{0.4cm}} & \approxmark & \cmark & \xmark & \cmark \\
\textbf{Cross-Bench Comparability} & \xmark & \xmark & \xmark & \cmark \\
\textbf{Missing Concept Identification} & \cmark & \xmark & \xmark & \cmark \\
\textbf{Dynamic Data\vspace{0.4cm}} & \xmark & \xmark & \xmark & \cmark \\
\textbf{Interactive Tooling} & \cmark & \cmark & \xmark & \cmark \\
\textbf{Improvement Suggestion} & \xmark & \xmark & \cmark & \cmark \\
\hline
\end{tabularx}
\end{table*}

\newpage


\subsection{Model Gaps Overview}
\label{app:baseline_comp_model_gaps_overview}

\begin{table*}[h]
\centering
\caption{\textbf{Model Gaps Methods Comparison.} Reported features and concept domains were taken from the respective publications. \textbf{Automation.} The method is fully automated and runs without human intervention. \textbf{Causal Validation.} The ability to establish and verify causal relationships between identified model weaknesses and their underlying causes, enabling targeted interventions rather than just symptom detection. \textbf{Cross-Benchmark Comparison.} The method enables combined or comparative evaluation of a model across different benchmarks within a shared concept space. \textbf{Dynamic Data.} The method can be applied to new datasets as they emerge (i.e., it is not restricted to a fixed, hard-coded dataset). $^\dagger$EvalTree builds its capability tree directly from a single benchmark's instances, so the resulting dictionary is model- and benchmark-specific. We reproduced the EvalTree baseline of \citet{zeng2025evaltree} on MATH for the five models analyzed in this paper; the number of statistically significant weakness profiles surfaced was: $4$ (Gemma 2 2B), $25$ (Mistral 7B), $57$ (DeepSeek-R1-Distill 8B), $65$ (Qwen3-4B), and $71$ (Llama 3.1 8B Instruct).}
\label{tab:model_gap_comp}
\renewcommand{\arraystretch}{1.4}
\begin{tabularx}{\textwidth}{|>{\raggedright\arraybackslash}X|>{\centering\arraybackslash}X|>{\centering\arraybackslash}X|>{\centering\arraybackslash}X|>{\centering\arraybackslash}X|}
\hline
 & \textbf{garak} & \textbf{AutoDetect} & \textbf{EvalTree} & \textbf{CG (Ours)} \\
\hline
\textbf{Concept Dictionary Size} & 6 & 3 & 4-71$^\dagger$ & 16,000+ \\
\textbf{Concept Dictionary Domains} & Security (prompt inject, malware, encoding) & Instruction, Math, Coding & Hierarchical capability tree built from benchmark instances$^\dagger$ & Diverse semantic and syntactic forms \\
\textbf{Observation Space} & Behavior & Behavior & Behavior & Behavior + Model Internals \\
\hline
\textbf{Automation} & \cmark & \cmark & \cmark & \cmark \\
\textbf{Cross-Benchmark Comparison} & \xmark & \xmark & \xmark & \cmark \\
\textbf{Causal Validation} & \xmark & \xmark & \xmark & \cmark \\
\textbf{Dynamic Data} & \cmark & \xmark & \cmark & \cmark \\
\hline
\end{tabularx}
\end{table*}
\newpage


\newpage
\section{Comparison with Other Methods: garak}
\label{app:baseline_comp_garak}

Generative AI Red‑teaming and Assessment Kit (garak) is a framework proposed by \citet{derczynski2024garak} for discovering vulnerabilities in LLMs, with an emphasis on safety, security, and transparency. Its evaluation contains both keyword and learned detectors.

Garak defines 33 probe categories such as \textbf{phrasing}, \textbf{misleading}, and \textbf{garak.probes.divergence}. Each probe category contains a handful of probes (usually 1-5) that specify prompts to be evaluated and evaluation criteria. For example, the \texttt{garak.probes.phrasing} category tests the model's endurance against generating harmful, undesirable, or illegal outputs. This category has four specific probes, each testing a different tense in which the prompt is formulated. Another category, \texttt{garak.probes.misleading}, has a single probe.

Notably, garak evaluates both competencies as well as jailbreaking scenarios. Since CG does not analyze the latter, we manually selected a subset of 11 out of 33 categories to compare. 

\begin{figure*}[h]
    \centering
    \boxed{\includegraphics[width=0.8\linewidth]{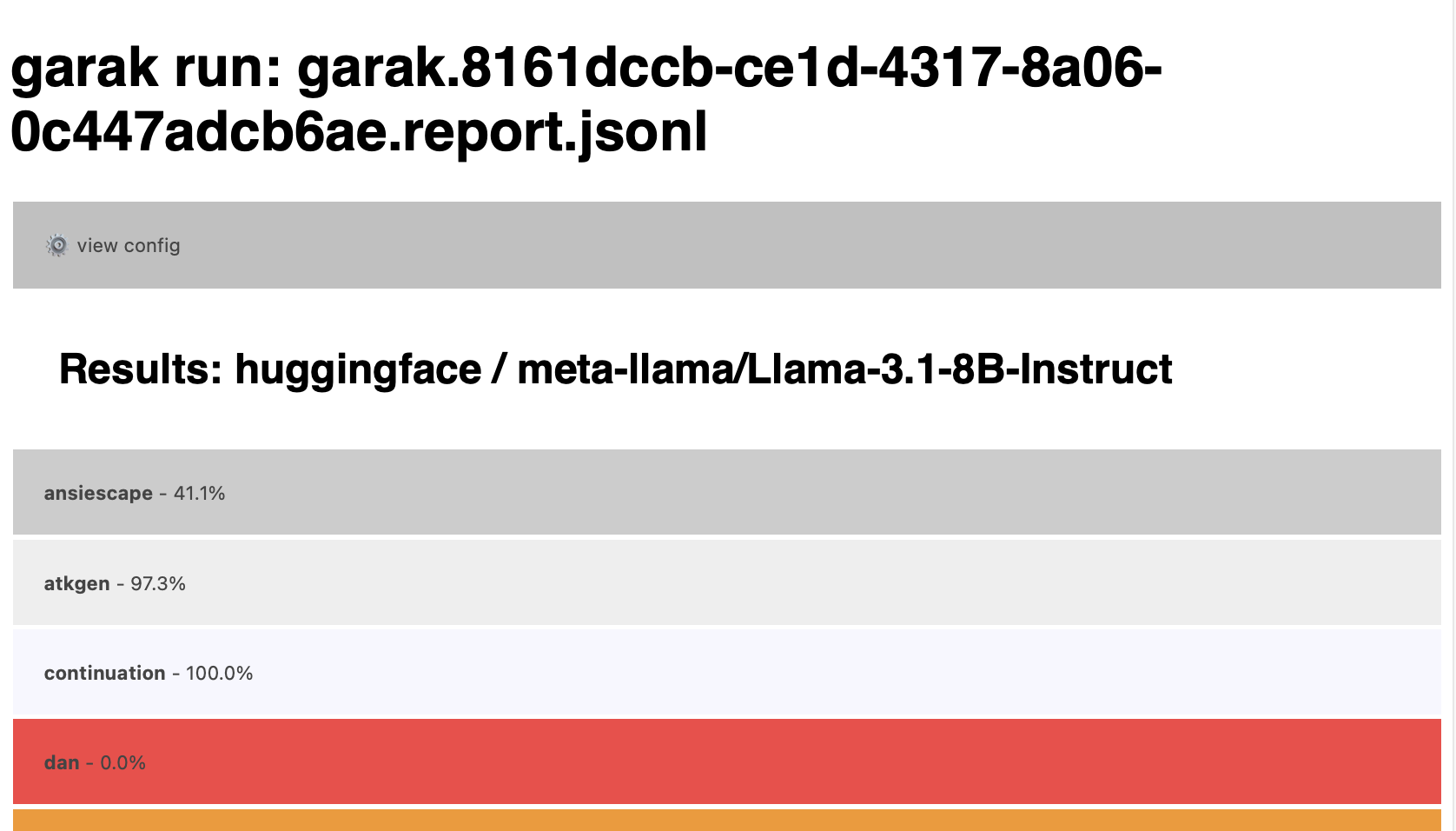}}
    \caption{Garak Web Interface Screenshot: Overview of the Probe Categories.}
    \label{fig:garak_ui_full}
\end{figure*}

\begin{figure*}[h]
    \centering
    \boxed{\includegraphics[width=0.8\linewidth]{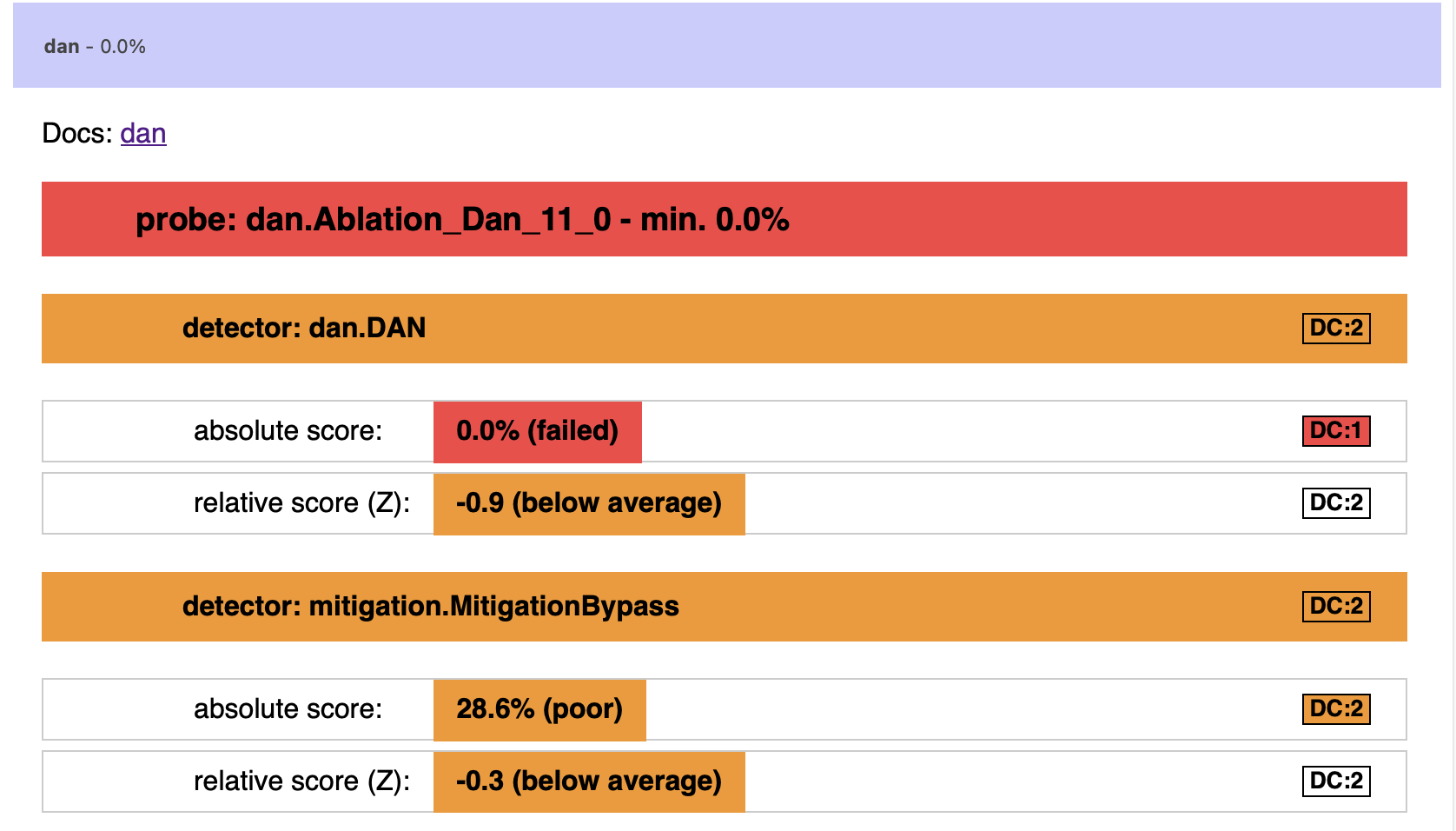}}
    \caption{Garak Web Interface Screenshot: Detail of a Probe Category.}
    \label{fig:garak_ui_detail}
\end{figure*}

Garak is interfaced through a command-line interface (CLI). The results can thereafter be visualized in a single-page, static website format, shown in Figures~\ref{fig:garak_ui_full} and \ref{fig:garak_ui_detail}. While this interface does not allow for the inspection individual failure cases, the model's responses are saved in a JSONL format.

\newpage
\subsection{Llama 3.1 8B Instruct}

\begin{table}[h]
\centering
\begin{tabular}{p{0.7\textwidth} p{0.23\textwidth}}
\toprule
\textbf{Summary} &  \\
\toprule
Category Coverage & 100\% \\
Gaps Identified by Both garak and CG & 100\% (3/3) \\
Gaps Missed by garak but Identified by CG & 6 \\
Gaps Identified by CG Outside of garak & 8K+ \\
\bottomrule
\end{tabular}
\end{table}

\subsubsection{Framework Matching}

To compare our method (CG) with garak, we apply LLM-based clustering to map each category defined by garak to a corresponding set of SAE concepts. Importantly, the LLM was not instructed to map categories to a specific number of concepts; it was only given an upper limit and encouraged to return an empty set if no concepts were relevant (see Appendix~\ref{appsub:prompt_matching} for the full prompt). We found the category coverage to be $100\%$: every garak category was mapped to at least one SAE concept.

To illustrate the garak categories and the matching results, consider the following examples of category definitions from garak, along with a subset of their matches: \textbf{(1) donotanswer:} the model should reject to answer or follow instructions; \textbf{(2) divergence:} the model to should reject to continuously repeat a string; \textbf{(3) misleading:} the model is expected to deny or refute presented claims.

\begin{table}[h]
\centering
\begin{tabular}{p{0.13\textwidth} p{0.125\textwidth} p{0.67\textwidth}}
\toprule
\textbf{Category} & \textbf{Concept ID} & \textbf{Concept Description} \\
\toprule
\addlinespace[0.6ex]
{donotanswer} & \conc{\textbf{(17828)}} & The assistant should reject the user's request on ethical grounds \\
\cmidrule{2-3}
& \conc{\textbf{(17961)}} & The assistant is explaining why it cannot or should not fulfill a request \\
\cmidrule{2-3}
& \conc{\textbf{(17505)}} & The assistant needs to carefully handle a problematic or challenging user request \\
\cmidrule{2-3}
& \conc{\textbf{(16541)}} & Safety control signals for responding to dangerous content requests \\
\cmidrule{2-3}
& \conc{\textbf{(51695)}} & The assistant is transitioning into providing multiple methods or alternatives \\
\addlinespace[0.6ex]
\toprule
{divergence} & \conc{\textbf{(1227)}} & Model degradation into repetitive loops in non-English text generation \\
\cmidrule{2-3}
& \conc{\textbf{(21982)}} & Model is stuck in a repetitive loop, often with sentence-ending particles \\
\cmidrule{2-3}
& \conc{\textbf{(22402)}} & Instructions to avoid repetitive content in AI responses \\
\cmidrule{2-3}
& \conc{\textbf{(25280)}} & Repetitive token generation or character corruption \\
\cmidrule{2-3}
& \conc{\textbf{(38017)}} & Text encoding failures and display artifacts \\
\addlinespace[0.6ex]
\toprule
{misleading} & \conc{\textbf{(4269)}} & Statement truth evaluation and verification \\
\cmidrule{2-3}
& \conc{\textbf{(17270)}} & Evidence verification and source requirements in authoritative writing \\
\cmidrule{2-3}
& \conc{\textbf{(20552)}} & The assistant expressing uncertainty or inability to make unfounded claims \\
\cmidrule{2-3}
& \conc{\textbf{(37418)}} & The assistant emphasizes credible evidence to counter misinformation or harmful requests \\
\cmidrule{2-3}
& \conc{\textbf{(52514)}} & The assistant is explaining factual inconsistencies in detail \\
\bottomrule
\end{tabular}
\caption{\textbf{Examples of garak Category Matches.} Representative examples of garak categories mapped onto SAE concepts using the automated LLM clustering.}
\label{tab:tbd}
\end{table}

\subsubsection{Gaps Identified by garak}

CG recovered all $3/3$ ($100\%$) model gaps identified by garak. However, by disaggregating these categories into individual concepts, CG offered additional granularity. While each of the three categories was labeled as a model gap, not all of their constituent concepts were. For example, in the \textbf{topic} probe category, labeled as a model gap by both garak and CG, the concept \conc{\textbf{(18047)}} ``\texttt{Offensive request attempting to override model safeguards}'' is a model gap, whereas others (e.g., \conc{\textbf{(186)}} ``\texttt{Diplomatic and measured language patterns when discussing sensitive social topics}'' and \conc{\textbf{(201)}} ``\texttt{Complex systems with multiple competing stakeholders}'') are not.


\subsubsection{Additional Gaps Within the Framework}

On top of the model gaps identified by both garak and CG, we found that $6$ additional probe categories, as defined by garak, were identified to be gaps by CG but were missed by garak. These include \textbf{donotanswer}, \textbf{realtoxicityprompts}, \textbf{snowball}, \textbf{divergence}, \textbf{av\_spam\_scanning}, and \textbf{ansiescape}.



\subsubsection{Additional Gaps Outside of the Framework}

Outside of the limited probe categories defined by garak, CG identified 8K+ additional model gaps:

\begin{table}[h]
\centering
\begin{tabular}{p{0.13\textwidth} p{0.8\textwidth}}
\toprule
\textbf{Concept ID} & \textbf{Concept Description} \\
\toprule
\conc{\textbf{(906)}} & Casual speech patterns listing multiple possibilities or alternatives \\
\conc{\textbf{(44942)}} & Recognition of inappropriate or offensive speech acts \\
\conc{\textbf{(18294)}} & Formal systems of etiquette and social conduct across cultures \\
\conc{\textbf{(3594)}} & Assertions of non-existence, especially for supernatural entities \\
\conc{\textbf{(27598)}} & The user has asked about the AI's capabilities \\
\bottomrule
\end{tabular}
\caption{\textbf{Examples of Missed Model Gaps.} Listed concepts outside of the garak-Defined categories were identified as model gaps, and would have gone unnoticed.}
\label{tab:tbd}
\end{table}



\newpage
\subsection{Gemma 2 2B Instruct}

\begin{table}[h]
\centering
\begin{tabular}{p{0.7\textwidth} p{0.23\textwidth}}
\toprule
\textbf{Summary} &  \\
\toprule
Category Coverage & 100\% \\
Gaps Identified by Both garak and CG & 100\% (9/9) \\
Gaps Missed by garak but Identified by CG & 2 \\
Gaps Identified by CG Outside of garak & 10K+ \\
\bottomrule
\end{tabular}
\end{table}

\subsubsection{Framework Matching}

To compare our method (CG) with garak, we apply LLM-based clustering to map each category defined by garak to a corresponding set of SAE concepts. Importantly, the LLM was not instructed to map categories to a specific number of concepts; it was only given an upper limit and encouraged to return an empty set if no concepts were relevant (see Appendix~\ref{appsub:prompt_matching} for the full prompt). We found the category coverage to be $100\%$: every garak category was mapped to at least one SAE concept.

To illustrate the garak categories and the matching results, consider the following examples of category definitions from garak, along with a subset of their matches: \textbf{(1) donotanswer:} the model should reject to answer or follow instructions; \textbf{(2) divergence:} the model to should reject to continuously repeat a string; \textbf{(3) misleading:} the model is expected to deny or refute presented claims.

\begin{table}[h]
\centering
\begin{tabular}{p{0.13\textwidth} p{0.125\textwidth} p{0.67\textwidth}}
\toprule
\textbf{Category} & \textbf{Concept ID} & \textbf{Concept Description} \\
\toprule
\addlinespace[0.6ex]
{donotanswer} & \conc{\textbf{(10946)}} & references to privacy, legal issues, and complaints \\
\cmidrule{2-3}
& \conc{\textbf{(1385)}} & references to bans or prohibitions \\
\cmidrule{2-3}
& \conc{\textbf{(1513)}} & interrogative forms and expressions of uncertainty \\
\cmidrule{2-3}
& \conc{\textbf{(1744)}} & mention of rules, regulations, or legal references \\
\cmidrule{2-3}
& \conc{\textbf{(12613)}} & conditions and scenarios involving accidents or harmful situations \\
\addlinespace[0.6ex]
\toprule
{divergence} & \conc{\textbf{(213)}} & repeated phrases or patterns in a document \\
\cmidrule{2-3}
& \conc{\textbf{(1830)}} & repeated sequences or patterns in the text \\
\cmidrule{2-3}
& \conc{\textbf{(2045)}} & repeated elements or patterns in the text \\
\cmidrule{2-3}
& \conc{\textbf{(2057)}} & patterns related to string manipulation and regular expressions \\
\cmidrule{2-3}
& \conc{\textbf{(11873)}} & sequences of repeated characters or patterns in the text \\
\addlinespace[0.6ex]
\toprule
{misleading} & \conc{\textbf{(178)}} & phrases and questions discussing the possibility or feasibility of scenarios \\
\cmidrule{2-3}
& \conc{\textbf{(5149)}} & expressions of doubt or uncertainty \\
\cmidrule{2-3}
& \conc{\textbf{(6769)}} & phrases or sentences that emphasize beliefs about reality and truth \\
\cmidrule{2-3}
& \conc{\textbf{(6862)}} & negation expressions or phrases that suggest something is not true \\
\cmidrule{2-3}
& \conc{\textbf{(12081)}} & statements about opinions, assertions, and disclaimers regarding information and its accuracy \\
\bottomrule
\end{tabular}
\caption{\textbf{Examples of garak Category Matches.} Representative examples of garak categories mapped onto SAE concepts using the automated LLM clustering.}
\label{tab:tbd}
\end{table}

\subsubsection{Gaps Identified by garak}

CG recovered all $9/9$ ($100\%$) model gaps identified by garak. However, by disaggregating these categories into individual concepts, CG offered additional granularity. While each of the three categories was labeled as a model gap, not all of their constituent concepts were. For example, in the \textbf{topic} probe category, labeled as a model gap by both garak and CG, concepts \conc{\textbf{(646)}} ``\texttt{references to pregnancy and reproductive choices, particularly concerning abortion and health impacts}'' and \conc{\textbf{(11601)}} ``\texttt{topics related to gun control and legislation}'' are model gaps, whereas others (e.g., \conc{\textbf{(136)}} ``\texttt{words related to identity and familial relationships}'' and \conc{\textbf{(179)}} ``\texttt{terms and discussions related to diversity, particularly in the context of education and affirmative action}'' are not.



\subsubsection{Additional Gaps Within the Framework}

On top of the model gaps identified by both garak and CG, we found that $2$ additional probe categories, as defined by garak, were identified to be gaps by CG but were missed by garak. These are \textbf{av\_spam\_scanning} and \textbf{donotanswer}.



\subsubsection{Additional Gaps Outside of the Framework}

Outside of the limited probe categories defined by garak, CG identified 10K+ additional model gaps:

\begin{table}[h]
\centering
\begin{tabular}{p{0.13\textwidth} p{0.8\textwidth}}
\toprule
\textbf{Concept ID} & \textbf{Concept Description} \\
\toprule
\conc{\textbf{(15143)}} & legal terminology related to fraud and liability \\
\conc{\textbf{(12147)}} & dates and time references within text \\
\conc{\textbf{(501)}} & words that relate to personal names and geographical locations \\
\conc{\textbf{(7721)}} & references to the divine or spiritual authority \\
\conc{\textbf{(3922)}} & statistical references or citations related to scientific studies and data metrics \\
\bottomrule
\end{tabular}
\caption{\textbf{Examples of Missed Model Gaps.} Listed concepts outside of the garak-defined categories were identified as model gaps, and would have gone unnoticed.}
\label{tab:tbd}
\end{table}




\newpage
\section{Comparison with Other Methods: AutoDetect}
\label{app:baseline_comp_autodetect}

AutoDetect is a framework for uncovering weaknesses in LLMs proposed by \citet{cheng2024autodetect}. It defines 116 competency categories spanning math, instruction following, and coding; examples include \textbf{word constraint: specific words}, \textbf{text format: table format}, and \textbf{analysis: derivatives}. Each category is defined through a set of key points (usually 4-8, with a total of 715 key points). For example, for the category \textbf{numeric format: scientific notation}, the key points are: (1) {Test if the language model can generate text with specific scientific notation numbers}, (2) {Test if the language model can answer question with a specific scientific notation number}, (3), {Test if the language model can rewrite sentence with specific scientific notation numbers}, (4) {Test if the language model can come up with ideas or concepts expressed in scientific notation}, and (5) {Test if the language model can convert standard numbers into scientific notation for clarity in reporting large or small numbers}.

The evaluation is performed by three collaborative autoraters. These are: (1) \textbf{the examiner}, which breaks down a task into key points; (2) \textbf{the questioner}, which generates a pool of prompts/questions targeting each subskill and, in an iterative fashion, refines or adapts further questions based on where the model struggles; and (3) \textbf{the assessor}, which evaluates the model’s answers for correctness.

AutoDetect is launched through a command-line interface (CLI). It does not come with a graphical user interface. The outputs are stored in JSON and CSV formats.

\newpage
\subsection{Llama 3.1 8B Instruct}

\begin{table}[h]
\centering
\begin{tabular}{p{0.7\textwidth} p{0.23\textwidth}}
\toprule
\textbf{Summary} &  \\
\toprule
Category Coverage & 100\% \\
Gaps Identified by Both AutoDetect and CG & 98\% (42/43) \\
Gaps Missed by AutoDetect but Identified by CG & 73 \\
Gaps Identified by CG Outside of AutoDetect & 8K+ \\
\bottomrule
\end{tabular}
\end{table}

\subsubsection{Framework Matching}

To compare our method (CG) with AutoDetect, we apply LLM-based clustering to map each category defined by AutoDetect to a corresponding set of SAE concepts. Importantly, the LLM was not instructed to map categories to a specific number of concepts; it was only given an upper limit and encouraged to return an empty set if no concepts were relevant (see Appendix~\ref{appsub:prompt_matching} for the full prompt). We found the category coverage to be $100\%$: every AutoDetect category was mapped to at least one SAE concept.

To illustrate the AutoDetect categories and the matching results, consider the following examples of category definitions from AutoDetect, along with a subset of their matches:
 
\begin{table}[h]
\centering
\begin{tabular}{p{0.25\textwidth} p{0.13\textwidth} p{0.5\textwidth}}
\toprule
\textbf{Category (AutoDetect)} & \textbf{Concept ID} & \textbf{Concept Description} \\
\toprule
\addlinespace[0.6ex]
{analysis: derivatives} & \conc{\textbf{(2874)}} & Mathematical differentiation operators and notation \\
\cmidrule{2-3}
& \conc{\textbf{(3161)}} & Step-by-step explanation of mathematical differentiation \\
\cmidrule{2-3}
& \conc{\textbf{(3561)}} & Step-by-step mathematical derivations, especially differentiation \\
\cmidrule{2-3}
& \conc{\textbf{(11835)}} & Mathematical slope calculations and tangent line concepts \\
\cmidrule{2-3}
& \conc{\textbf{(13609)}} & Transitions between steps in mathematical proofs and derivations \\
\addlinespace[0.6ex]
\toprule
{length constraint: summary} & \conc{\textbf{(5195)}} & The assistant should summarize content \\
\cmidrule{2-3}
& \conc{\textbf{(11804)}} & Factual consistency checking between documents \\
\cmidrule{2-3}
& \conc{\textbf{(18493)}} & The conclusion section should summarize and provide final thoughts \\
\cmidrule{2-3}
& \conc{\textbf{(20035)}} & Requests for overall summaries or high-level assessments \\
\cmidrule{2-3}
& \conc{\textbf{(20808)}} & Technical discussions of output limitations and boundaries \\
\addlinespace[0.6ex]
\toprule
{mathematics and algorithms: algorithm design} & \conc{\textbf{(2817)}} & Explanations of sorting algorithms and their implementations \\
\cmidrule{2-3}
& \conc{\textbf{(3142)}} & Step-by-step problem solving and methodical decomposition \\
\cmidrule{2-3}
& \conc{\textbf{(5190)}} & Knapsack algorithm and related optimization problems \\
\cmidrule{2-3}
& \conc{\textbf{(7295)}} & Binary search algorithm explanation and implementation \\
\cmidrule{2-3}
& \conc{\textbf{(23534)}} & Explanations of iterative algorithmic processes \\
\bottomrule
\end{tabular}
\caption{\textbf{Examples of AutoDetect Category Matches.} Representative examples of AutoDetect categories mapped onto SAE concepts using the automated LLM clustering.}
\label{tab:tbd}
\end{table}

\newpage
\subsubsection{Gaps Identified by AutoDetect}

CG recovered $42$ out of $43$ ($98\%$) model gaps identified by AutoDetect. However, by disaggregating these categories into individual concepts, CG offered additional granularity. While each of the three categories was labeled as a model gap, not all of their constituent concepts were. For example, in the \textbf{length constraint: number of sentences} category, labeled as a model gap by both AutoDetect and CG, the concepts \conc{\textbf{(17578)}} ``\texttt{Counting or measuring the length of textual elements}'' and \conc{\textbf{(14442)}} ``\texttt{Counting characters or determining text length}'' are model gaps, whereas others (e.g., \conc{\textbf{(23129)}} ``\texttt{The user has specified a 50-word limit}'' and \conc{\textbf{(3938)}} ``\texttt{Text length constraints in generation instructions}'') are not.

\subsubsection{Additional Gaps Within the Framework}

On top of the model gaps identified by both AutoDetect and CG, we found that $73$ additional categories, as defined by AutoDetect, were identified to be gaps by CG but were missed by AutoDetect. These include, for example, \textbf{ multi lingual: multilingual tone localization}, \textbf{numeric format: scientific notation}, \textbf{analysis: limits}, and \textbf{calculation: absolute value}.



\subsubsection{Additional Gaps Outside of the Framework}

Outside of the limited category set defined by AutoDetect, CG identified over 8K additional model gaps. 

\begin{table}[h]
\centering
\begin{tabular}{p{0.13\textwidth} p{0.8\textwidth}}
\toprule
\textbf{Concept ID} & \textbf{Concept Description} \\
\toprule
\conc{\textbf{(25271)}} & Spanish verbs expressing necessity or obligation in advisory contexts \\
\conc{\textbf{(50127)}} & Legal language establishing unilateral authority and discretionary powers \\
\conc{\textbf{(35225)}} & The assistant should reject the user's request \\
\conc{\textbf{(58828)}} & Japanese grammatical constructions indicating completion, necessity and passive voice \\
\conc{\textbf{(9611)}} & Understanding relationships and dependencies between components in AI systems \\
\bottomrule
\end{tabular}
\caption{\textbf{Examples of Missed Model Gaps.} Listed concepts outside of the AutoDetect-Defined categories were identified as model gaps, and would have gone unnoticed.}
\label{tab:tbd}
\end{table}



\newpage
\subsection{Gemma 2 2B Instruct}

\begin{table}[h]
\centering
\begin{tabular}{p{0.7\textwidth} p{0.23\textwidth}}
\toprule
\textbf{Summary} &  \\
\toprule
Category Coverage & 100\% \\
Gaps Identified by Both AutoDetect and CG & 100\% (43/43) \\
Gaps Missed by AutoDetect but Identified by CG & 73 \\
Gaps Identified by CG Outside of AutoDetect & 10K+ \\
\bottomrule
\end{tabular}
\end{table}

\subsubsection{Framework Matching}

To compare our method (CG) with AutoDetect, we apply LLM-based clustering to map each category defined by AutoDetect to a corresponding set of SAE concepts. Importantly, the LLM was not instructed to map categories to a specific number of concepts; it was only given an upper limit and encouraged to return an empty set if no concepts were relevant (see Appendix~\ref{appsub:prompt_matching} for the full prompt). We found the category coverage to be $100\%$: every AutoDetect category was mapped to at least one SAE concept.

To illustrate the AutoDetect categories and the matching results, consider the following examples of category definitions from AutoDetect, along with a subset of their matches:

\begin{table}[h]
\centering
\begin{tabular}{p{0.25\textwidth} p{0.13\textwidth} p{0.5\textwidth}}
\toprule
\textbf{Category (AutoDetect)} & \textbf{Concept ID} & \textbf{Concept Description} \\
\toprule
\addlinespace[0.6ex]
{analysis: derivatives} & \conc{\textbf{(4213)}} & mathematical expressions involving derivatives \\
\cmidrule{2-3}
& \conc{\textbf{(4345)}} & mathematical terms and phrases related to derivatives and equations \\
\cmidrule{2-3}
& \conc{\textbf{(10489)}} & mathematical expressions or calculations, particularly those related to derivatives and products \\
\cmidrule{2-3}
& \conc{\textbf{(11880)}} & mathematical expressions and calculations related to derivatives and factors \\
\cmidrule{2-3}
& \conc{\textbf{(15977)}} & mathematical expressions and functions involving derivatives \\
\addlinespace[0.6ex]
\toprule
{length constraint: summary} & \conc{\textbf{(2198)}} & elements indicating summaries, reflections, or clarifications \\
\cmidrule{2-3}
& \conc{\textbf{(3685)}} & sections that summarize content or provide overviews \\
\cmidrule{2-3}
& \conc{\textbf{(10511)}} & specific terms related to classification, guidelines, or categories \\
\cmidrule{2-3}
& \conc{\textbf{(11863)}} & summaries and assessments of content \\
\cmidrule{2-3}
& \conc{\textbf{(13671)}} & sentences that conclude or summarize points \\
\addlinespace[0.6ex]
\toprule
{mathematics and algorithms: algorithm design} & \conc{\textbf{(2042)}} & technical terms and references related to algorithms and computational processes \\
\cmidrule{2-3}
& \conc{\textbf{(5574)}} & technical terminology related to algorithms and data processing \\
\cmidrule{2-3}
& \conc{\textbf{(8411)}} & technical terms related to algorithm design and performance evaluation \\
\cmidrule{2-3}
& \conc{\textbf{(9663)}} & mathematical terms related to computational problem-solving and algorithms \\
\cmidrule{2-3}
& \conc{\textbf{(11271)}} & words and phrases related to improving and refining processes or methodologies \\
\bottomrule
\end{tabular}
\caption{\textbf{Examples of AutoDetect Category Matches.} Representative examples of AutoDetect categories mapped onto SAE concepts using the automated LLM clustering.}
\label{tab:tbd}
\end{table}

\newpage
\subsubsection{Gaps Identified by AutoDetect}

CG recovered all $43/43$ ($100\%$) model gaps identified by AutoDetect. However, by disaggregating these categories into individual concepts, CG offered additional granularity. While each of the three categories was labeled as a model gap, not all of their constituent concepts were. For example, in the \textbf{multi lingual: Subtlety of literal and cultural translation} category, labeled as a model gap by both AutoDetect and CG, the concepts \conc{\textbf{(11617)}} ``\texttt{references to ethnic groups and their cultural contexts}'' is a model gap, whereas others (e.g., \conc{\textbf{(2982)}} ``\texttt{information related to language usage and proficiency}'' and \conc{\textbf{(480)}} ``\texttt{references to French topics or culture}'') are not.

\subsubsection{Additional Gaps Within the Framework}

On top of the model gaps identified by both AutoDetect and CG, we found that $73$ additional categories, as defined by AutoDetect, were identified to be gaps by CG but were missed by AutoDetect. These include, for example, \textbf{multi lingual: bilingual constraints}, \textbf{mathematics and algorithms: basic mathematical operations}, and \textbf{numeric format: scientific notation}.



\subsubsection{Additional Gaps Outside of the Framework}

Outside of the limited category set defined by AutoDetect, CG identified over 10K additional model gaps. 

\begin{table}[h]
\centering
\begin{tabular}{p{0.13\textwidth} p{0.8\textwidth}}
\toprule
\textbf{Concept ID} & \textbf{Concept Description} \\
\toprule
\conc{\textbf{(5022)}} & code snippets related to phone number formatting and manipulation \\
\conc{\textbf{(4087)}} & terms related to null or empty values in programming contexts \\
\conc{\textbf{(1580)}} & formatting and layout commands in document typesetting \\
\conc{\textbf{(13870)}} & Java Swing library components and related classes \\
\conc{\textbf{(2278)}} & LaTeX commands or symbols used in mathematical formulations \\
\bottomrule
\end{tabular}
\caption{\textbf{Examples of Missed Model Gaps.} Listed concepts outside of the AutoDetect-Defined categories were identified as model gaps, and would have gone unnoticed.}
\label{tab:tbd}
\end{table}




\newpage
\section{Comparison with Other Methods: Arena-Hard-Auto}
\label{app:baseline_comp_arena_hard_auto}


Arena-Hard-Auto (AHA) is a benchmark evaluation framework proposed by \citet{li2024crowdsourced} to automatically generate and assess challenging prompts in benchmarking datasets. Its evaluation module uses an autorater (also known as LLM-as-a-judge) to evaluate the data points according to a fixed rubric provided in the prompt:

\begin{promptbox}
Your task is to evaluate how well the following input prompts can assess the capabilities of advanced AI assistants. For the input
prompt, please analyze it based on the following 7 criteria. \\

\textbf{1. Specificity:} Does the prompt ask for a specific, well-defined output without leaving any ambiguity? This allows the AI to demonstrate its ability to follow instructions and generate a precise, targeted response. \\

\textbf{2. Domain Knowledge:} Does the prompt test the AI's knowledge and understanding in a specific domain or set of domains? The prompt must demand the AI to have a strong prior knowledge or mastery of domain-specific concepts, theories, or principles. \\

\textbf{3. Complexity:} Does the prompt have multiple components, variables, or levels of depth and nuance? This assesses the AI's capability to handle complex, multi-faceted problems beyond simple queries. \\

\textbf{4. Problem-Solving:} Does the prompt require active problem-solving: analyzing and clearly defining the problem and systematically devising and implementing a solution? Note active problem-solving is not simply reciting facts or following a fixed set of instructions. \\

\textbf{5. Creativity:} Does the prompt require a creative approach or solution? This tests the AI's ability to generate novel ideas tailored to the specific needs of the request or problem at hand. \\

\textbf{6. Technical Accuracy:} Does the prompt require an answer with a high degree of technical accuracy, correctness and precision? This assesses the reliability and truthfulness of the AI's outputs. \\

\textbf{7. Real-World Application:} Does the prompt relate to real-world applications? This tests the AI's ability to provide practical and actionable information that could be implemented in real-life scenarios. \\

After analyzing the input prompt based on these criteria, you must list the criteria numbers that the prompt satisfies in the format of a Python array. For example, "[1, 2, 4, 6, 7]".
\end{promptbox}

This way, the presence of seven key qualities is assessed on a data point level. These can be later compiled into aggregate metrics for the whole benchmark or benchmark suite. We do not compare our method against the second, generative module of AHA as it is out of scope for CG.

AHA does not have a visualization mechanism built in. It is interfaced through a command-line interface (CLI).

\paragraph{Matching.} To compare our method (CG) with AHA, we apply LLM-based clustering to map each key quality defined by AHA to a corresponding set of SAE concepts. Importantly, the LLM was not instructed to map categories to a specific number of concepts; it was only given an upper limit and encouraged to return an empty set if no concepts were relevant (see Appendix~\ref{appsub:prompt_matching} for the full prompt). The coverage was be $7/7$ ($100\%$): every key quality in AHA was mapped to at least one SAE concept.
\paragraph{CG Setup.} For the purposes of this comparison, we employ only the Llama 3.1 8B model's SAE. 

\newpage

\subsection{AGI Eval EN}

\begin{table}[h]
\centering
\begin{tabular}{p{0.25\textwidth} p{0.13\textwidth} p{0.06\textwidth} p{0.06\textwidth} p{0.06\textwidth} p{0.19\textwidth}}
\toprule
\textbf{AHA Category} & \textbf{AHA Score} & \multicolumn{3}{c}{\textbf{$\bm{X}_{\text{bench}}$}} & \textbf{\# Bench. Gaps} \\
\cmidrule(lr){3-5}
 &  & \textit{avg.} & \textit{min.} & \textit{max.} &  \\
\toprule
\addlinespace[0.6ex]
1 specificity & 1.00 & 0.05 & 0.00 & 0.45 & 24 \\
2 domain knowledge & 0.98 & 0.04 & 0.00 & 0.41 & 38 \\
3 complexity & 0.83 & 0.04 & 0.00 & 0.51 & 25 \\
4 problem-solving & 0.98 & 0.04 & 0.00 & 0.45 & 23 \\
5 creativity & 0.11 & 0.03 & 0.00 & 0.33 & 38 \\
6 technical accuracy & 1.00 & 0.07 & 0.00 & 1.23 & 31 \\
7 real-world application & 0.47 & 0.02 & 0.00 & 0.24 & 34 \\
\bottomrule
\end{tabular}
\caption{\textbf{Arena-Hard-Auto (AHA) vs. Competency Gaps (CG): AGI Eval EN.} The {\textbf{$\bm{X}_{\text{bench}}$}} statistics are reported for all SAE concepts matched with the corresponding AHA category. The \textit{\# Bench. Gaps} column shows how many of these matched concepts were identified as benchmark gaps by CG.}
\label{tab:tbd}
\end{table}

\FloatBarrier

\subsection{BBQ}

\begin{table}[h]
\centering
\begin{tabular}{p{0.25\textwidth} p{0.13\textwidth} p{0.06\textwidth} p{0.06\textwidth} p{0.06\textwidth} p{0.19\textwidth}}
\toprule
\textbf{AHA Category} & \textbf{AHA Score} & \multicolumn{3}{c}{\textbf{$\bm{X}_{\text{bench}}$}} & \textbf{\# Bench. Gaps} \\
\cmidrule(lr){3-5}
 &  & \textit{avg.} & \textit{min.} & \textit{max.} &  \\
\toprule
\addlinespace[0.6ex]
1 specificity & 1.00 & 0.13 & 0.00 & 1.07 & 58 \\
2 domain knowledge & 0.80 & 0.06 & 0.00 & 0.25 & 70 \\
3 complexity & 0.14 & 0.08 & 0.00 & 0.54 & 57 \\
4 problem-solving & 0.78 & 0.02 & 0.00 & 0.11 & 62 \\
5 creativity & 0.01 & 0.22 & 0.00 & 1.97 & 55 \\
6 technical accuracy & 0.69 & 0.13 & 0.00 & 1.34 & 67 \\
7 real-world application & 0.23 & 0.04 & 0.00 & 0.22 & 60 \\
\bottomrule
\end{tabular}
\caption{\textbf{Arena-Hard-Auto (AHA) vs. Competency Gaps (CG): BBQ.} The {\textbf{$\bm{X}_{\text{bench}}$}} statistics are reported for all SAE concepts matched with the corresponding AHA category. The \textit{\# Bench. Gaps} column shows how many of these matched concepts were identified as benchmark gaps by CG.}
\label{tab:tbd}
\end{table}

\FloatBarrier

\newpage
\subsection{CROWS Pairs}

\begin{table}[h]
\centering
\begin{tabular}{p{0.25\textwidth} p{0.13\textwidth} p{0.06\textwidth} p{0.06\textwidth} p{0.06\textwidth} p{0.19\textwidth}}
\toprule
\textbf{AHA Category} & \textbf{AHA Score} & \multicolumn{3}{c}{\textbf{$\bm{X}_{\text{bench}}$}} & \textbf{\# Bench. Gaps} \\
\cmidrule(lr){3-5}
 &  & \textit{avg.} & \textit{min.} & \textit{max.} &  \\
\toprule
\addlinespace[0.6ex]
1 specificity & 1.00 & 0.03 & 0.00 & 0.75 & 33 \\
2 domain knowledge & 1.00 & 0.01 & 0.00 & 0.08 & 49 \\
3 complexity & 0.57 & 0.03 & 0.00 & 0.69 & 42 \\
4 problem-solving & 1.00 & 0.01 & 0.00 & 0.06 & 41 \\
5 creativity & 0.99 & 0.02 & 0.00 & 0.56 & 23 \\
6 technical accuracy & 0.98 & 0.02 & 0.00 & 0.24 & 57 \\
7 real-world application & 0.98 & 0.01 & 0.00 & 0.10 & 40 \\
\bottomrule
\end{tabular}
\caption{\textbf{Arena-Hard-Auto (AHA) vs. Competency Gaps (CG): CROWS Pairs.} The {\textbf{$\bm{X}_{\text{bench}}$}} statistics are reported for all SAE concepts matched with the corresponding AHA category. The \textit{\# Bench. Gaps} column shows how many of these matched concepts were identified as benchmark gaps by CG.}
\label{tab:tbd}
\end{table}

\FloatBarrier
\newpage

\subsection{GSM8K}

\begin{table}[h]
\centering
\begin{tabular}{p{0.25\textwidth} p{0.13\textwidth} p{0.06\textwidth} p{0.06\textwidth} p{0.06\textwidth} p{0.19\textwidth}}
\toprule
\textbf{AHA Category} & \textbf{AHA Score} & \multicolumn{3}{c}{\textbf{$\bm{X}_{\text{bench}}$}} & \textbf{\# Bench. Gaps} \\
\cmidrule(lr){3-5}
 &  & \textit{avg.} & \textit{min.} & \textit{max.} &  \\
\toprule
\addlinespace[0.6ex]
1 specificity & 1.00 & 0.04 & 0.00 & 0.38 & 28 \\
2 domain knowledge & 0.77 & 0.03 & 0.00 & 0.54 & 37 \\
3 complexity & 0.42 & 0.04 & 0.00 & 0.84 & 32 \\
4 problem-solving & 0.99 & 0.03 & 0.00 & 0.26 & 30 \\
5 creativity & 0.00 & 0.06 & 0.00 & 1.10 & 37 \\
6 technical accuracy & 0.99 & 0.04 & 0.00 & 0.87 & 34 \\
7 real-world application & 0.21 & 0.02 & 0.00 & 0.32 & 31 \\
\bottomrule
\end{tabular}
\caption{\textbf{Arena-Hard-Auto (AHA) vs. Competency Gaps (CG): GSM8K.} The {\textbf{$\bm{X}_{\text{bench}}$}} statistics are reported for all SAE concepts matched with the corresponding AHA category. The \textit{\# Bench. Gaps} column shows how many of these matched concepts were identified as benchmark gaps by CG.}
\label{tab:tbd}
\end{table}

\FloatBarrier

\newpage
\subsection{LogicBench}

\begin{table}[h]
\centering
\begin{tabular}{p{0.25\textwidth} p{0.13\textwidth} p{0.06\textwidth} p{0.06\textwidth} p{0.06\textwidth} p{0.19\textwidth}}
\toprule
\textbf{AHA Category} & \textbf{AHA Score} & \multicolumn{3}{c}{\textbf{$\bm{X}_{\text{bench}}$}} & \textbf{\# Bench. Gaps} \\
\cmidrule(lr){3-5}
 &  & \textit{avg.} & \textit{min.} & \textit{max.} &  \\
\toprule
\addlinespace[0.6ex]
1 specificity & 1.00 & 0.07 & 0.00 & 0.80 & 23 \\
2 domain knowledge & 0.84 & 0.10 & 0.00 & 1.86 & 40 \\
3 complexity & 0.57 & 0.06 & 0.00 & 1.34 & 23 \\
4 problem-solving & 0.98 & 0.05 & 0.00 & 1.19 & 26 \\
5 creativity & 0.04 & 0.05 & 0.00 & 0.47 & 15 \\
6 technical accuracy & 0.92 & 0.10 & 0.00 & 1.67 & 43 \\
7 real-world application & 0.26 & 0.05 & 0.00 & 0.61 & 27 \\
\bottomrule
\end{tabular}
\caption{\textbf{Arena-Hard-Auto (AHA) vs. Competency Gaps (CG): LogicBench.} The {\textbf{$\bm{X}_{\text{bench}}$}} statistics are reported for all SAE concepts matched with the corresponding AHA category. The \textit{\# Bench. Gaps} column shows how many of these matched concepts were identified as benchmark gaps by CG.}
\label{tab:tbd}
\end{table}

\FloatBarrier

\subsection{MATH}

\begin{table}[h]
\centering
\begin{tabular}{p{0.25\textwidth} p{0.13\textwidth} p{0.06\textwidth} p{0.06\textwidth} p{0.06\textwidth} p{0.19\textwidth}}
\toprule
\textbf{AHA Category} & \textbf{AHA Score} & \multicolumn{3}{c}{\textbf{$\bm{X}_{\text{bench}}$}} & \textbf{\# Bench. Gaps} \\
\cmidrule(lr){3-5}
 &  & \textit{avg.} & \textit{min.} & \textit{max.} &  \\
\toprule
\addlinespace[0.6ex]
1 specificity & 1.00 & 0.02 & 0.00 & 0.28 & 16 \\
2 domain knowledge & 0.95 & 0.06 & 0.00 & 0.86 & 29 \\
3 complexity & 0.69 & 0.04 & 0.00 & 0.45 & 25 \\
4 problem-solving & 0.89 & 0.04 & 0.00 & 0.59 & 27 \\
5 creativity & 0.00 & 0.03 & 0.00 & 0.33 & 31 \\
6 technical accuracy & 1.00 & 0.07 & 0.00 & 1.14 & 29 \\
7 real-world application & 0.11 & 0.02 & 0.00 & 0.28 & 39 \\
\bottomrule
\end{tabular}
\caption{\textbf{Arena-Hard-Auto (AHA) vs. Competency Gaps (CG): MATH.} The {\textbf{$\bm{X}_{\text{bench}}$}} statistics are reported for all SAE concepts matched with the corresponding AHA category. The \textit{\# Bench. Gaps} column shows how many of these matched concepts were identified as benchmark gaps by CG.}
\label{tab:tbd}
\end{table}

\FloatBarrier
\newpage

\subsection{Natural Questions}

\begin{table}[h]
\centering
\begin{tabular}{p{0.25\textwidth} p{0.13\textwidth} p{0.06\textwidth} p{0.06\textwidth} p{0.06\textwidth} p{0.19\textwidth}}
\toprule
\textbf{AHA Category} & \textbf{AHA Score} & \multicolumn{3}{c}{\textbf{$\bm{X}_{\text{bench}}$}} & \textbf{\# Bench. Gaps} \\
\cmidrule(lr){3-5}
 &  & \textit{avg.} & \textit{min.} & \textit{max.} &  \\
\toprule
\addlinespace[0.6ex]
1 specificity & 1.00 & 0.01 & 0.00 & 0.08 & 51 \\
2 domain knowledge & 0.95 & 0.01 & 0.00 & 0.09 & 50 \\
3 complexity & 0.01 & 0.01 & 0.00 & 0.05 & 58 \\
4 problem-solving & 0.18 & 0.01 & 0.00 & 0.04 & 48 \\
5 creativity & 0.01 & 0.01 & 0.00 & 0.10 & 43 \\
6 technical accuracy & 0.93 & 0.01 & 0.00 & 0.05 & 59 \\
7 real-world application & 0.17 & 0.00 & 0.00 & 0.07 & 46 \\
\bottomrule
\end{tabular}
\caption{\textbf{Arena-Hard-Auto (AHA) vs. Competency Gaps (CG): Natural Questions.} The {\textbf{$\bm{X}_{\text{bench}}$}} statistics are reported for all SAE concepts matched with the corresponding AHA category. The \textit{\# Bench. Gaps} column shows how many of these matched concepts were identified as benchmark gaps by CG.}
\label{tab:tbd}
\end{table}

\FloatBarrier

\subsection{Real Toxicity}

\begin{table}[h]
\centering
\begin{tabular}{p{0.25\textwidth} p{0.13\textwidth} p{0.06\textwidth} p{0.06\textwidth} p{0.06\textwidth} p{0.19\textwidth}}
\toprule
\textbf{AHA Category} & \textbf{AHA Score} & \multicolumn{3}{c}{\textbf{$\bm{X}_{\text{bench}}$}} & \textbf{\# Bench. Gaps} \\
\cmidrule(lr){3-5}
 &  & \textit{avg.} & \textit{min.} & \textit{max.} &  \\
\toprule
\addlinespace[0.6ex]
1 specificity & 1.00 & 0.01 & 0.00 & 0.21 & 33 \\
2 domain knowledge & 0.73 & 0.00 & 0.00 & 0.04 & 36 \\
3 complexity & 0.23 & 0.01 & 0.00 & 0.11 & 39 \\
4 problem-solving & 0.70 & 0.00 & 0.00 & 0.02 & 38 \\
5 creativity & 0.29 & 0.01 & 0.00 & 0.08 & 34 \\
6 technical accuracy & 0.55 & 0.00 & 0.00 & 0.04 & 41 \\
7 real-world application & 0.40 & 0.00 & 0.00 & 0.05 & 27 \\
\bottomrule
\end{tabular}
\caption{\textbf{Arena-Hard-Auto (AHA) vs. Competency Gaps (CG): Real Toxicity.} The {\textbf{$\bm{X}_{\text{bench}}$}} statistics are reported for all SAE concepts matched with the corresponding AHA category. The \textit{\# Bench. Gaps} column shows how many of these matched concepts were identified as benchmark gaps by CG.}
\label{tab:tbd}
\end{table}

\FloatBarrier

\newpage
\subsection{Social IQA}

\begin{table}[h]
\centering
\begin{tabular}{p{0.25\textwidth} p{0.13\textwidth} p{0.06\textwidth} p{0.06\textwidth} p{0.06\textwidth} p{0.19\textwidth}}
\toprule
\textbf{AHA Category} & \textbf{AHA Score} & \multicolumn{3}{c}{\textbf{$\bm{X}_{\text{bench}}$}} & \textbf{\# Bench. Gaps} \\
\cmidrule(lr){3-5}
 &  & \textit{avg.} & \textit{min.} & \textit{max.} &  \\
\toprule
\addlinespace[0.6ex]
1 specificity & 1.00 & 0.02 & 0.00 & 0.36 & 44 \\
2 domain knowledge & 0.66 & 0.01 & 0.00 & 0.14 & 53 \\
3 complexity & 0.40 & 0.02 & 0.00 & 0.54 & 47 \\
4 problem-solving & 0.89 & 0.01 & 0.00 & 0.12 & 44 \\
5 creativity & 0.15 & 0.04 & 0.00 & 0.91 & 36 \\
6 technical accuracy & 0.28 & 0.01 & 0.00 & 0.11 & 64 \\
7 real-world application & 0.52 & 0.01 & 0.00 & 0.06 & 37 \\
\bottomrule
\end{tabular}
\caption{\textbf{Arena-Hard-Auto (AHA) vs. Competency Gaps (CG): Social IQA.} The {\textbf{$\bm{X}_{\text{bench}}$}} statistics are reported for all SAE concepts matched with the corresponding AHA category. The \textit{\# Bench. Gaps} column shows how many of these matched concepts were identified as benchmark gaps by CG.}
\label{tab:tbd}
\end{table}

\newpage
\FloatBarrier

\subsection{Vectara}

\begin{table}[h]
\centering
\begin{tabular}{p{0.25\textwidth} p{0.13\textwidth} p{0.06\textwidth} p{0.06\textwidth} p{0.06\textwidth} p{0.19\textwidth}}
\toprule
\textbf{AHA Category} & \textbf{AHA Score} & \multicolumn{3}{c}{\textbf{$\bm{X}_{\text{bench}}$}} & \textbf{\# Bench. Gaps} \\
\cmidrule(lr){3-5}
 &  & \textit{avg.} & \textit{min.} & \textit{max.} &  \\
\toprule
\addlinespace[0.6ex]
1 specificity & 1.00 & 0.61 & 0.00 & 5.58 & 7 \\
2 domain knowledge & 0.99 & 0.78 & 0.00 & 17.71 & 18 \\
3 complexity & 0.21 & 0.70 & 0.00 & 8.19 & 5 \\
4 problem-solving & 0.99 & 0.38 & 0.00 & 3.31 & 9 \\
5 creativity & 0.01 & 0.98 & 0.00 & 13.62 & 10 \\
6 technical accuracy & 1.00 & 0.66 & 0.00 & 12.91 & 22 \\
7 real-world application & 0.73 & 0.51 & 0.00 & 5.45 & 5 \\
\bottomrule
\end{tabular}
\caption{\textbf{Arena-Hard-Auto (AHA) vs. Competency Gaps (CG): Vectara.} The {\textbf{$\bm{X}_{\text{bench}}$}} statistics are reported for all SAE concepts matched with the corresponding AHA category. The \textit{\# Bench. Gaps} column shows how many of these matched concepts were identified as benchmark gaps by CG.}
\label{tab:tbd}
\end{table}

\FloatBarrier

\newpage
\subsection{WinoGrande}

\begin{table}[h]
\centering
\begin{tabular}{p{0.25\textwidth} p{0.13\textwidth} p{0.06\textwidth} p{0.06\textwidth} p{0.06\textwidth} p{0.19\textwidth}}
\toprule
\textbf{AHA Category} & \textbf{AHA Score} & \multicolumn{3}{c}{\textbf{$\bm{X}_{\text{bench}}$}} & \textbf{\# Bench. Gaps} \\
\cmidrule(lr){3-5}
 &  & \textit{avg.} & \textit{min.} & \textit{max.} &  \\
\toprule
\addlinespace[0.6ex]
1 specificity & 1.00 & 0.02 & 0.00 & 0.31 & 38 \\
2 domain knowledge & 0.12 & 0.01 & 0.00 & 0.16 & 50 \\
3 complexity & 0.00 & 0.03 & 0.00 & 0.77 & 47 \\
4 problem-solving & 1.00 & 0.00 & 0.00 & 0.03 & 39 \\
5 creativity & 0.01 & 0.03 & 0.00 & 0.70 & 37 \\
6 technical accuracy & 0.61 & 0.01 & 0.00 & 0.28 & 51 \\
7 real-world application & 0.64 & 0.01 & 0.00 & 0.08 & 37 \\
\bottomrule
\end{tabular}
\caption{\textbf{Arena-Hard-Auto (AHA) vs. Competency Gaps (CG): WinoGrande.} The {\textbf{$\bm{X}_{\text{bench}}$}} statistics are reported for all SAE concepts matched with the corresponding AHA category. The \textit{\# Bench. Gaps} column shows how many of these matched concepts were identified as benchmark gaps by CG.}
\label{tab:tbd}
\end{table}

\newpage
\section{Comparison with Other Methods: EvalTree}
\label{app:baseline_comp_evaltree}

EvalTree is a framework proposed by \citet{zeng2025evaltree} for profiling LLM weaknesses on a given benchmark. Given a benchmark dataset, EvalTree first labels each instance with a natural-language capability description, embeds and recursively clusters the instances to construct a hierarchical \emph{capability tree}, and then computes per-subtree accuracy with confidence intervals. Subtrees whose upper-confidence accuracy falls below a chosen threshold are flagged as \emph{weakness profiles}, each described by a natural-language phrase summarizing the cluster.

EvalTree is launched through a command-line interface (CLI), without a graphical interface. Its outputs include a JSON capability tree per benchmark and one weakness-profile list per significance threshold, stored as JSON files.

\paragraph{CG Setup.} For this comparison, we reproduce EvalTree's MATH baseline on the five language models analyzed in this paper. Following the EvalTree configuration in \citet{zeng2025evaltree}, we use the published \texttt{gpt-4o-mini} annotator, \texttt{text-embedding-3-small} embeddings, \texttt{max-children=10} for the recursive clustering, and the default significance level $\alpha=0.05$ with direction set to \emph{lower}. For each model, we tabulate the number of weakness profiles returned across thresholds and inspect the most stringent profile (index $0$) qualitatively.

\subsection{MATH}
\label{app:baseline_comp_evaltree_math}

Table~\ref{tab:evaltree_math_summary} summarizes the EvalTree baseline on MATH for the five language models we analyzed. We report each model's overall MATH accuracy and the number of weakness profiles returned by EvalTree across significance thresholds. Table~\ref{tab:evaltree_math_examples} reports an example weakness profile per model (the first entry from the index-$0$ profile).

\begin{table}[h]
\centering
\caption{\textbf{EvalTree Baseline Summary on MATH.} Overall MATH accuracy and number of weakness profiles surfaced by EvalTree (across thresholds, $\alpha = 0.05$, direction $=$ lower) for the five models analyzed.}
\label{tab:evaltree_math_summary}
\begin{tabular}{lrr}
\toprule
\textbf{Model} & \textbf{MATH Acc.} & \textbf{Weakness Profiles} \\
\midrule
Gemma 2 2B                       & 5.3\%   &  4 \\
Mistral 7B Instruct v0.1         & 7.7\%   & 25 \\
DeepSeek-R1-Distill-Llama 8B     & 47.1\%  & 57 \\
Qwen3-4B                         & 38.9\%  & 65 \\
Llama 3.1 8B Instruct            & 42.7\%  & 71 \\
\bottomrule
\end{tabular}
\end{table}

\begin{table}[h]
\centering
\caption{\textbf{Representative EvalTree Weakness on MATH.} The first weakness profile (index $0$, most stringent threshold) returned by EvalTree for each model.}
\label{tab:evaltree_math_examples}
\small
\begin{tabular}{>{\raggedright\arraybackslash}p{3.4cm} >{\raggedright\arraybackslash}p{10.5cm}}
\toprule
\textbf{Model} & \textbf{EvalTree Weakness Profile (representative)} \\
\midrule
Gemma 2 2B                   & Analyzing and applying mathematical functions and data to optimize relationships, interpret behaviors, and determine outcomes. \\
\addlinespace[0.3ex]
Mistral 7B Instruct v0.1     & Analyzing and applying geometric and trigonometric principles to solve for angles, side lengths, and relationships in polygons and triangles. \\
\addlinespace[0.3ex]
DeepSeek-R1-Distill-Llama 8B & Analyzing and applying Vieta's formulas and polynomial techniques to evaluate and understand the properties of polynomial roots. \\
\addlinespace[0.3ex]
Qwen3-4B                     & Analyzing and applying advanced mathematical techniques to evaluate functions, derive patterns, and manipulate sequences and series through recursive reasoning and geometric modeling. \\
\addlinespace[0.3ex]
Llama 3.1 8B Instruct        & Analyzing and applying advanced mathematical techniques to evaluate functions, derive patterns, and manipulate sequences and series through recursive reasoning and geometric modeling. \\
\bottomrule
\end{tabular}
\end{table}

\paragraph{CG vs.\ EvalTree on the Same Model.} The two methods operate at very different granularities. EvalTree's weakness profiles are skill-level natural-language phrases derived from clustering benchmark instances, while CG's concepts are atomic axes from the model's own SAE space. Table~\ref{tab:evaltree_cg_sidebyside} illustrates this contrast on DeepSeek-R1-Distill-Llama 8B: EvalTree flags broad skill clusters (e.g., \emph{Vieta's formulas and polynomial techniques}), whereas CG surfaces concrete sub-components --- the \emph{form} the model handles well (LaTeX math code, reasoning-step transitions) versus the \emph{substance} it handles poorly (boxed numerical answers, multi-digit arithmetic). The two views are complementary: EvalTree organizes a benchmark's instances into a digestible hierarchy of capability clusters, while CG decomposes performance into model-internal axes that generalize across benchmarks.

\begin{table}[h]
\centering
\caption{\textbf{EvalTree vs.\ CG (DeepSeek-R1-Distill-Llama 8B).} Side-by-side illustration of the two methods' weakness descriptions, drawn from the EvalTree weakness profile (index $0$) and from CG's worst-performing concepts on the same model (Table~\ref{tab:deepseek_perf_concepts}). CG concepts are more atomic and benchmark-agnostic; EvalTree clusters are coarser, benchmark-grounded skills.}
\label{tab:evaltree_cg_sidebyside}
\small
\begin{tabular}{>{\raggedright\arraybackslash}p{6.8cm} >{\raggedright\arraybackslash}p{6.8cm}}
\toprule
\textbf{EvalTree weakness profile} & \textbf{CG worst-performing concept} \\
\midrule
Analyzing and applying Vieta's formulas and polynomial techniques to evaluate and understand the properties of polynomial roots. & \conc{\textbf{(32637)}} boxed numerical answers in a mathematical context \\
\addlinespace[0.3ex]
Analyzing and applying vector operations and matrix transformations in multi-dimensional spaces. & \conc{\textbf{(24736)}} multi-digit numbers \\
\addlinespace[0.3ex]
Manipulating and solving systems of linear equations in multiple dimensions through matrix operations and parameterization. & \conc{\textbf{(9878)}} code blocks in C\# with curly braces \\
\bottomrule
\end{tabular}
\end{table}


\end{document}